\documentclass[lettersize,journal]{IEEEtran}
\usepackage{amsmath,amsfonts}
\usepackage{algorithmic}
\usepackage{algorithm}
\usepackage{array}
\usepackage[caption=false,font=normalsize,labelfont=sf,textfont=sf]{subfig}
\usepackage{textcomp}
\usepackage{stfloats}
\usepackage{url}
\usepackage{verbatim}
\usepackage{graphicx}
\usepackage{cite}
\usepackage{xcolor}
\usepackage{colortbl}
\usepackage{xurl}
\usepackage{hyperref}
\usepackage{tabularx}
\usepackage{booktabs}
\usepackage{balance}
\usepackage{tikz}
\usepackage{orcidlink}
\usetikzlibrary{arrows.meta}
\usepackage{soul}
\usepackage{multirow}
\usepackage{fancyhdr}
\usepackage{threeparttable}
\usepackage{ragged2e}
\definecolor{trustHigh}{HTML}{DDEED8} % soft green
\definecolor{trustMed}{HTML}{FCE4D6}  % soft orange/amber
\definecolor{trustLow}{HTML}{D9EAF7}  % soft blue-gray

\newcommand{\High}{\cellcolor{trustHigh}\textbf{High}}
\newcommand{\Medium}{\cellcolor{trustMed}\textbf{Medium}}
\newcommand{\Low}{\cellcolor{trustLow}\textbf{Low}}
\newcommand{\LowMed}{\cellcolor{trustMed}\textbf{Low--Medium}}
\newcommand{\MedHigh}{\cellcolor{trustHigh}\textbf{Medium--High}}

\definecolor{trustStrong}{HTML}{2E6F57}   % deep muted green
\definecolor{trustPartial}{HTML}{C07A00}  % distinct professional amber
\definecolor{trustWeak}{HTML}{A33A3A}     % deep muted red

\newcommand{\Full}{\textcolor{trustStrong}{\ensuremath{\checkmark}}}
\newcommand{\Partial}{\textcolor{trustPartial}{\ensuremath{\triangle}}}
\newcommand{\Weak}{\textcolor{trustWeak}{\ensuremath{\times}}}

\newcommand{\todo}[1]{\textcolor{red}{#1}}

\makeatletter
\newcommand{\ieeeruninsubsection}{%
  \@startsection{subsection}{2}{\parindent}%
    {0ex plus 0.1ex minus 0.1ex}%
    {0ex}%
    {\normalfont\normalsize\itshape}%
}
\makeatother

\begin{document}
\bstctlcite{IEEEexample:BSTcontrol}
\title{Engineering Trustworthy Agentic AI for Critical Systems}

% \author{IEEE Publication Technology,~\IEEEmembership{Staff,~IEEE,}
        % <-this % stops a space
% \thanks{This paper was produced by the IEEE Publication Technology Group. They are in Piscataway, NJ.}% <-this % stops a space
% \thanks{Manuscript received April 19, 2021; revised August 16, 2021.}}

% % The paper headers
% \markboth{Journal of \LaTeX\ Class Files,~Vol.~14, No.~8, August~2021}%
% {Shell \MakeLowercase{\textit{et al.}}: A Sample Article Using IEEEtran.cls for IEEE Journals}

% \IEEEpubid{}
% Remember, if you use this you must call \IEEEpubidadjcol in the second
% column for its text to clear the IEEEpubid mark.

\author{Omar~Al-Refai~\orcidlink{0009-0002-8357-0668},~\IEEEmembership{Student Member,~IEEE,}
        Ibrahim~Shahbaz~\orcidlink{0009-0007-2520-3970},~\IEEEmembership{Student Member,~IEEE,}
        Adam~Ali~Husseinat~\orcidlink{0009-0002-0426-0486},~\IEEEmembership{Member,~IEEE,}
        Michael~Mandulak~\orcidlink{0009-0002-6656-6237},
        Jaewon~Kim~\orcidlink{0000-0002-3021-729X},~\IEEEmembership{Member,~IEEE,}
        Eman~Hammad~\orcidlink{0000-0001-6069-1550},~\IEEEmembership{Senior Member,~IEEE}

 \thanks{Corresponding authors: Omar Al-Refai (e-mail: omaralrefai@tamu.edu) and Eman Hammad (e-mail: eman.hammad@tamu.edu).}       
\thanks{O. Al-Refai, I. Shahbaz, A. Ali Husseinat, and E. Hammad are with the Innovations in Systems Trust and Resilience (iSTAR) Laboratory, Texas A\&M University, TX, USA. E.Hammad and M. Mandulak are with the Texas A\&M Institute of Data Science - Security, Privacy and Resilience for Trusted AI (SPARTA) Thematic Lab., Texas A\&M University, TX, USA. J. Kim is with the Texas A\&M Global Cyber Research Institute (GCRI), Texas A\&M University, TX, USA.}

 }

\maketitle
\thispagestyle{fancy}
\pagestyle{fancy}
\fancyhf{}
\renewcommand{\headrulewidth}{0pt}
\renewcommand{\footrulewidth}{0pt}
\fancyfoot[C]{\scriptsize This work has been submitted to the IEEE for possible publication. Copyright may be transferred without notice, after which this version may no longer be accessible.}
\thispagestyle{fancy}
\begin{abstract}
Agentic artificial intelligence systems, capable of autonomous perception, planning, tool use, and multi-step action, are increasingly proposed for critical engineering domains where decisions carry physical, operational, or economic consequences. This survey addresses a gap in current literature by treating trustworthiness, whether agentic behavior can be verified, audited, and trusted under the constraints that engineering practice actually requires, as a first-class engineering property, rather than evaluating agentic AI by task capability alone. The study adopts a trustworthiness model organized around five cross-cutting dimensions: safety and constraint satisfaction; robustness and reliability; transparency and interpretability; accountability and auditability; and privacy and security. This is mapped onto an agentic assurance workflow spanning perception through audit. Building on this foundation, agentic systems architectures, threats, concrete trust mechanisms, and quantitative metrics are surveyed for direct application in agentic systems development and evaluation. These principles are then examined across four constraint-bound engineering domains: power systems, autonomous vehicles/robotics/UAVs, high-performance computing, and communication networks, identifying recurring design patterns, shared failure modes, and domain-specific gaps. Synthesizing across those domains, agentic AI trustworthiness is shown to be a single problem, with a path outlined toward a reusable, cross-domain assurance framework analogous to the graded certification regimes used by mature safety-critical engineering fields.
\end{abstract}

%=================================================================

\section{Introduction}

\IEEEPARstart{A}{gentic} artificial intelligence (Agentic AI) refers to AI systems designed to autonomously perceive their environment, reason over goals and intermediate states, and execute sequences of actions; often iteratively and over extended horizons to achieve objectives within dynamic environments~\cite{fi17090404}. The key difference between agentic AI systems and static predictive single-turn large language models (LLMs) lies in their incorporation of explicit decision-making loops, memory, tool use, and sometimes coordination among heterogeneous agents, including humans.

Recent advances have operationalized these ideas through architectures such as reasoning–acting interleaving (e.g., ReAct)~\cite{yao2023reactsynergizingreasoningacting}, self-reflective agents (e.g., Reflexion)~\cite{shinn2023reflexionlanguageagentsverbal}, deliberative search over intermediate reasoning steps (e.g., Tree-of-Thoughts)~\cite{yao2023treethoughtsdeliberateproblem}, and multi-agent LLM-based systems capable of communication and task decomposition. Collectively, these approaches mark a shift from passive AI components toward autonomous, goal-directed systems that can adapt, plan, and act in complex environments. Although these capabilities have shown promise in various domains, their implementation in critical infrastructure systems such as power grids, autonomous vehicles and robotic platforms, high-performance computing facilities, and communication networks presents fundamental trustworthiness challenges~\cite{11103638}. Failures in these environments can cause cascading outages, physical harm, financial loss, or safety hazards to the public and human operators, in addition to performance degradations~\cite{nithrakashyap2025aiagents, damle2025sleepwalking}.

Trust in agentic AI is an existential requirement rather than an auxiliary concern, requiring autonomous agents to consistently respect operational constraints, maintain resilience against adversity and uncertainty, provide transparency into decision-making processes, facilitate accountability and auditability, and prevent abuse or unintentional escalation. Agentic AI systems, in contrast to traditional automation, can create and execute new action sequences, communicate with external tools or simulators, and adjust their tactics online, thereby enhancing both their potential impact and their failure modes.

This survey addresses the growing gap between the rapid development of agentic AI architectures and the limited understanding of their reliability when used in critical infrastructures. This work offers a structured, trust-centered analysis that specifically targets domains where autonomous decision-making carries significant societal, economic, and safety consequences, rather than providing a broad or capability-driven overview of agentic AI. In addition to cataloging agentic trustworthiness techniques, we synthesize existing knowledge on whether and under what circumstances such systems can be responsibly relied upon in operational environments with strict requirements and low fault tolerance.

We contribute to this effort in four ways. First, we create a unified taxonomy of agentic AI architectures that encompasses prevalent design patterns from the literature, including reasoning-acting interleaving agents, self-reflective and memory-augmented agents, tool-augmented agents based on external environments, deliberative planning agents based on structured search, and multi-agent systems with coordination and role specialization capabilities. The survey's conceptual foundation, grounded in this taxonomy, enables systematic comparisons between otherwise disparate methods.

Second, we organize these architectural classes around five cross-cutting trustworthiness dimensions: safety and constraint satisfaction, robustness and reliability, transparency and interpretability, accountability and auditability, and privacy and security, mapped onto a ten-point agentic workflow spanning perception through audit~\cite{AgenticAIAutonomousSurvey}. Instead of treating trust as a post-hoc or abstract issue, we examine where in this workflow trust mechanisms are present, and where they are absent, exposing recurring patterns of incomplete coverage, implicit assumptions, and unresolved risk that are frequently overlooked in capability-focused discussions of agentic AI.

Third, we translate this workflow-level view into material developers can act on directly: a survey of trust mechanisms already embedded in agentic loops, including guardrails, safe reinforcement learning formulations, and robustness measures against distribution shift and adversarial manipulation; a review of how AI safety evaluation has evolved from representation- and reasoning-level probes to behavior, risk, and trajectory-level assessment of agentic workflows; and a consolidated set of quantitative metrics, spanning tool use, memory coherence, planning quality, policy adherence, and goal drift, that can be reported directly rather than inferred from task success alone.

Fourth, we ground this architectural and trust analysis in four constraint-bound engineering domains: power systems, autonomous vehicles/robotics/UAVs, high-performance computing, and communication networks. We examine how agentic AI has been proposed or implemented across these domains, including the types of simulators, benchmarks, and testbeds used for assessment, and the degree to which trustworthiness is operationalized through quantifiable constraints, failure analyses, or deployment-relevant metrics. By comparing and contrasting agentic AI in these domains, we identify and analyze discrepancies between real-world operational requirements and benchmark-driven validation, as well as domain-specific failure modes that generic agent evaluations do not capture. This led to a major insight that agentic AI trustworthiness is not four separate engineering problems but one problem observed from four vantage points, obscured by domain-specific vocabulary and disconnected evaluation practices. The recurring gaps of 1) the absence of domain-aware safety guarantees under distribution shift and adversarial conditions, 2) the lack of standardized trust-oriented evaluation, 3) limited integration of formal verification and runtime monitoring, and 4) inadequate human-agent oversight appear in each domain under different vocabulary but share a common structure. Building on this insight, the survey outlines a path toward a reusable, cross-domain trustworthiness framework analogous to the graded certification regimes that aviation and automotive safety engineering already use, shifting the field's next phase of progress from architectural novelty toward assurance infrastructure.

In contrast to current surveys on agentic AI that mostly focus on reasoning skills, coordination methods, and performance metrics, this work highlights trustworthiness and the challenges posed by critical infrastructure. Unlike surveys on trustworthy AI that mainly discuss static models or single-decision scenarios, we specifically address the unique risks posed by autonomous, sequential, and tool-integrated agentic systems. This survey fills an important gap where agentic AI, trustworthiness, and high-stakes real-world applications meet. It provides a solid foundation for researchers and practitioners seeking to transition agentic AI from experimental settings to practical applications.

The remainder of this paper is organized as follows. Section~\ref{sec:background} introduces agentic AI architectures and the four target domains. Section~\ref{sec:trust_dimensions} presents our five trustworthiness dimensions and the ten-point assurance workflow. Section~\ref{sec:taxonomy} analyzes representative agentic architectures through this trust lens. Section~\ref{sec:trust_mechanisms} surveys concrete trust mechanisms, Section~\ref{sec:evolution_of_benchmarks} traces the evolution of AI safety evaluation from representation-level probes to agentic, trajectory-level assessment, and Section~\ref{sec:metrics} consolidates quantitative metrics for agentic workflows. Sections~\ref{sec:ps_main}, \ref{sec:av_robotics_uav}, \ref{sec:hpc}, and \ref{sec:comm} examine domain-specific applications, Section~\ref{sec:cross_domain} synthesizes cross-domain insights and gaps, and Section~\ref{sec:open_challenges} concludes with open challenges and future directions. 

%=================================================================
\section{Background: Agentic Architectures and Trustworthy AI}
\label{sec:background}
%TODO: Define what we mean by Agentic Architecture

\subsection{Agentic Architectures with LLMs}

Recent advances in large language models (LLMs) have led to a shift from static, single-turn inference to agentic AI structures. In these systems, models are part of closed-loop setups that can make decisions over time, interact with their surroundings, and adapt their behavior. Unlike traditional LLM usage, where models produce responses based on a set prompt, agentic structures treat LLMs as decision-making cores. These models constantly observe state information, consider goals and outcomes, take actions, and update their internal information based on the feedback they receive. This change has led to a growing number of LLM-based agents.

At a high level, agentic architectures can be categorized based on the number of decision-making entities involved and their interaction patterns. Single-agent architectures consist of a single LLM-driven agent operating within an environment, often enhanced with memory and access to tools. These agents typically handle planning, reasoning, and acting independently to achieve a specific goal. In contrast, multi-agent architectures involve several agents, each potentially powered by an LLM, that coordinate, communicate, or compete to solve tasks together. Multi-agent systems add complexity through communication protocols, role specialization, negotiation, and conflict resolution. They also increase the challenges posed by coordination failures and unexpected behaviors.

A key feature of modern agentic AI systems is their use of tools and interaction with their environment. Instead of working solely in text space, agents increasingly connect with external APIs, simulators, databases, code interpreters, cyber ranges, or physical system models~\cite{toolsapis2026}. Using these tools allows agents to go beyond language generation. They can query real-time information, execute control actions, or manipulate simulated or real-world environments. In critical infrastructure contexts, these environments may include power system simulators, reinforcement learning settings, network emulators, or operational dashboards. This tightens the link between agent decisions and system-level consequences.

\begin{figure}[t]
    \centering
    \includegraphics[
        width=1.0\linewidth,
        trim={40pt 35pt 20pt 35pt},
        clip
    ]{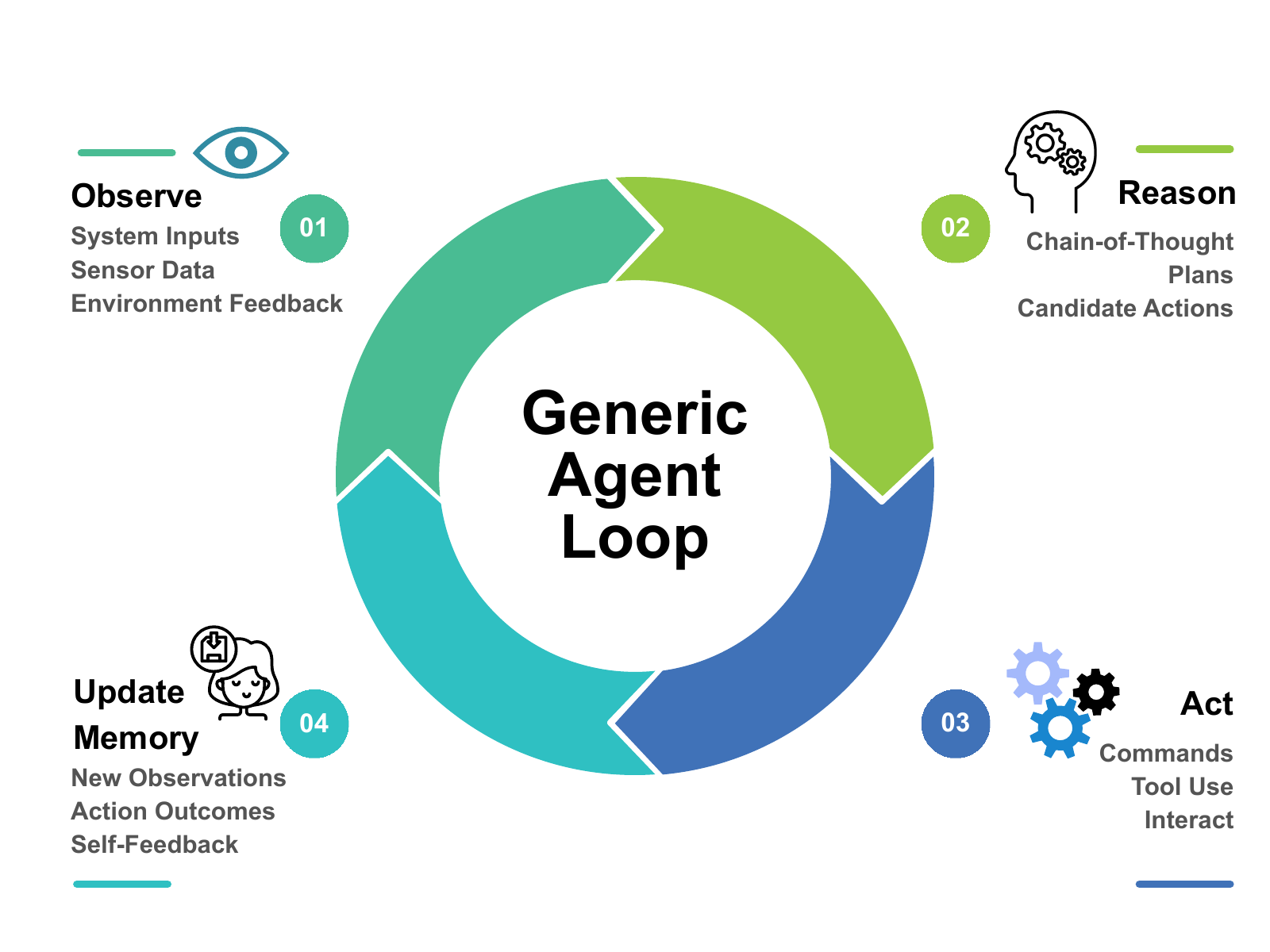}
    \caption{Generic agentic AI observe--reason--act loop.}
    \label{fig:generic_agent_loop}
\end{figure}

Most agentic architectures can be described using a generic agent loop, as shown in Figure~\ref{fig:generic_agent_loop}, which captures the repeating structure behind different implementations. In this loop, the agent first observes the current state of the environment, including system outputs, sensor data, or feedback from previous actions. It then reasons over this information to determine the best course of action, often using intermediate representations such as chain-of-thought reasoning, plans, or evaluations of candidate actions. The agent then acts by issuing commands, using tools, or interacting with other agents or environments. Finally, the agent updates its memory, adding new observations, action outcomes, or self-reflective feedback to guide future decisions. This observe, reason, act, and update cycle continues until a termination condition is met.

While this agent loop offers flexibility, it also raises new trust issues that do not exist with static model inference. Errors can accumulate over iterations, actions may have irreversible effects, and unexpected behaviors can emerge from long-term interactions with complex environments or other agents. Therefore, it is vital to understand how different agentic architectures implement this loop, and where safeguards, constraints, or oversight mechanisms are included. This understanding is crucial for assessing their suitability in high-stakes, safety-critical situations. This foundational insight drives the taxonomy and trust-centered analysis presented in the rest of this survey.

Building on the single-agent loop, multi-agent systems (MAS) spread these processes across a network of specialized entities. In a multi-agent setup, the "environment" often becomes a shared space. Here, the actions of one agent become the observations of another. This creates a coordination loop that controls how individual agents interact. These systems typically depend on a communication layer, often called a "blackboard" or a messaging protocol, to share plans, intermediate reasoning, or tool outputs.

The multi-agent setup introduces three critical components beyond the standard loop:
\begin{itemize}
    \item \textbf{Role Specialization:} Agents are often assigned specific "personas" or subsets of tools (e.g., a "Coder" agent and a "Reviewer" agent), allowing for a divide-and-conquer approach to complex tasks.
    \item \textbf{Communication Protocols:} Agents must decide not just \textit{what} to do, but \textit{what to tell} others. This includes sharing state updates, requesting assistance, or negotiating during conflicts.
    \item \textbf{Global vs. Local Memory:} While individual agents maintain local logs of their reasoning, a multi-agent system often incorporates a shared memory or "Global State" to ensure all participants remain aligned on the ultimate goal.
\end{itemize}

\begin{figure*}[t]
    \centering
    \includegraphics[
        width=\linewidth,
        trim={120pt 120pt 120pt 120pt},
        clip
    ]{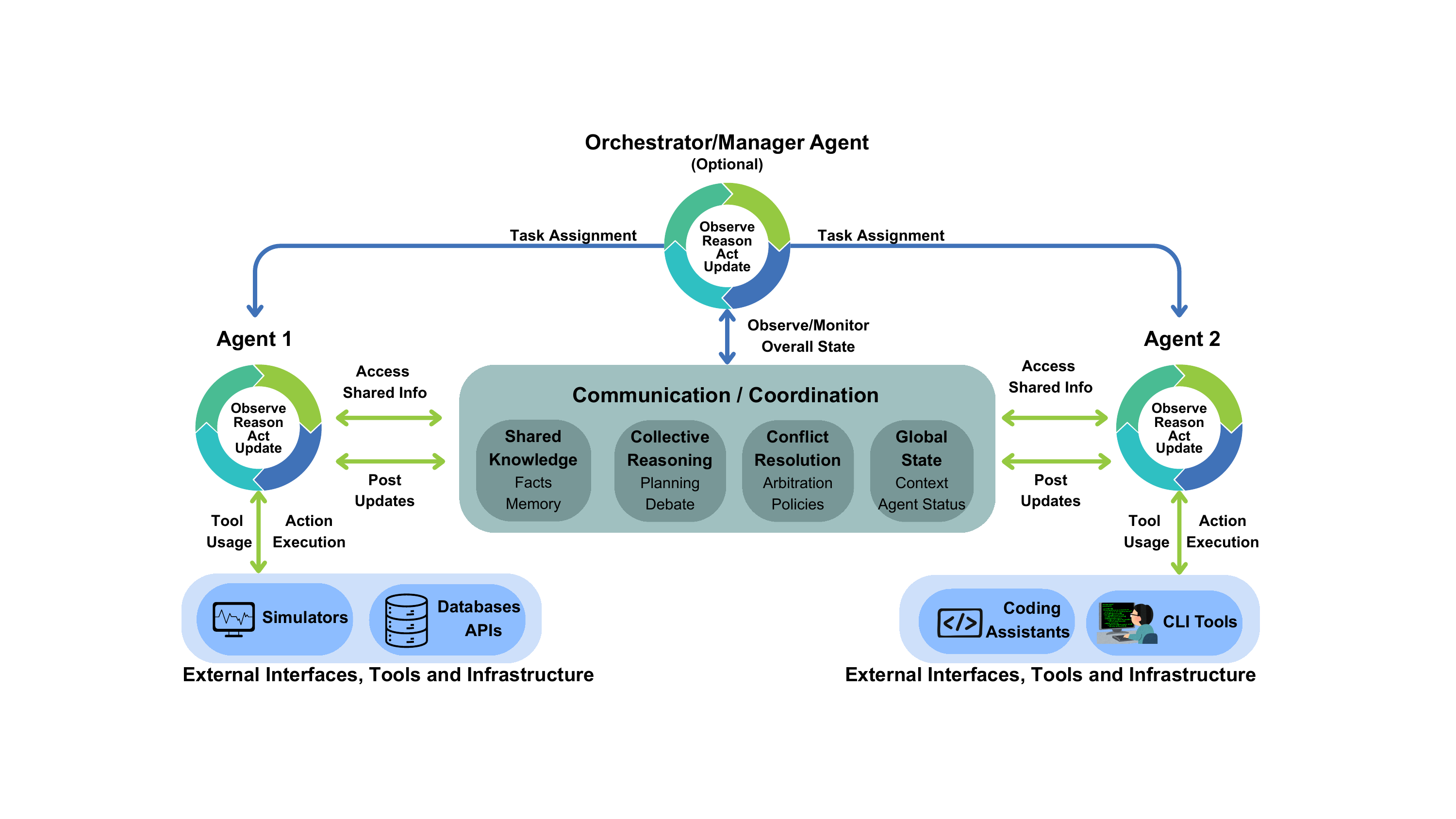}
    \caption{Generic multi-agent architecture showing interconnected loops, shared memory, and collective interaction in a common environment.}
    \label{fig:multi_agent_setup}
\end{figure*}

As in the MAS illustrated in Figure~\ref{fig:multi_agent_setup}, the complexity shifts from individual thinking to group coordination. While this can improve performance on separate tasks, it also brings in "emergent risks." Minor errors in one agent's output propagate as they pass through the network of agent interactions.

This transition from individual reasoning to collective behavior necessitates a dual-layered approach to trustworthiness. We must evaluate intra-agent reliability, ensuring the internal decision-making core of each entity remains robust, alongside inter-agent trust, which governs the integrity of communication, coordination, and delegated authority across the network. Assessing these two dimensions is essential to preventing local errors from escalating into systemic failures within the broader ecosystem.

\subsection{Case Studies}

This survey presents a general case-study perspective on trustworthy agentic AI, while synthesizing the discussion through four representative engineering domains: \textbf{power systems}, \textbf{autonomous vehicles, robotics, and UAVs}, \textbf{high-performance computing}, and \textbf{communication networks}. These domains provide a concrete basis for examining how agentic AI may be introduced into existing monitoring, control, optimization, and decision-making loops to support human operators, augment established automation pipelines, or increasingly assume parts of the operational process. They are selected because they are mature engineering fields with well-defined operational constraints, hard system requirements, and safety-critical deployments. Consequently, trustworthiness cannot be viewed solely as a property of the AI model, but must instead be evaluated against domain-specific requirements, including safety, reliability, timing, resilience, resource constraints, security, accountability, and the consequences of incorrect autonomous actions. While these four domains do not encompass every application of agentic AI, they provide a technically grounded foundation for studying trust in systems with significant real-world consequences.

Together, these domains span a broad range of system scales and architectural characteristics. Power systems represent large-scale cyber-physical infrastructure with tightly coupled physical dynamics; autonomous vehicles, robotics, and UAVs embody mission-level autonomy in dynamic environments; high-performance computing exposes low-level computational and resource-management constraints; and communication networks capture the distributed data flows that interconnect these systems. This bounded yet diverse scope enables meaningful comparisons of trustworthiness challenges across complementary engineering settings while maintaining a focus on domains where agentic AI requires rigorous architectural design and system-level assurance.

%=================================================================

\section{Dimensions of Trustworthiness}
\label{sec:trust_dimensions}

In critical engineering domains, trustworthiness cannot be treated as a static property of an AI model or an isolated software component. Agentic AI systems operate over extended time horizons, interact with tools and environments, and may trigger actions whose effects propagate across physical, cyber, and organizational layers~\cite{AGITrustworthiness}. Accordingly, this survey adopts a resilience-oriented view of trustworthiness: an agentic system is trustworthy to the extent that it can \emph{anticipate}, \emph{withstand}, \emph{recover from}, and \emph{adapt to} disturbances, including faults, distribution shifts, specification gaps, and adversarial manipulation, while remaining within acceptable operational bounds.

We organize this view around five cross-cutting dimensions: safety and constraint satisfaction, robustness and reliability, transparency and interpretability, accountability and auditability, and privacy and security. These dimensions are informed by established safety, assurance, and governance frameworks, including functional safety standards, Safety of the Intended Functionality (SOTIF), cybersecurity guidance, and emerging AI governance regimes such as the EU AI Act. They are not treated as independent checkboxes; rather, they serve as analytical lenses for comparing how agentic AI systems are designed, evaluated, and justified in critical application settings.

\ieeeruninsubsection{Safety and Constraint Satisfaction}

Safety and constraint satisfaction refer to the extent to which agent actions remain within operational, physical, legal, and procedural limits. In safety-critical engineering, safety is commonly defined as freedom from unacceptable risk~\cite{IEC2022}. For agentic AI, this requires more than high task performance: the system must respect safety envelopes, operational rules, and hard constraints even when goals conflict, information is incomplete, or the environment changes unexpectedly. This dimension is closely related to SOTIF, which emphasizes hazards that arise not only from component failures but also from functional insufficiencies and specification gaps in complex autonomous systems~\cite{Shinde2024}. In practice, safety evidence may include formal constraints, runtime monitors, fallback policies, hazard analysis, safety integrity targets, or empirical demonstrations that the agent does not violate critical operating limits~\cite{ConstPolicyOptimization, IEC61508FunctionalSafety2022}.

\ieeeruninsubsection{Robustness and Reliability}

Robustness and reliability capture the agent's ability to maintain acceptable behavior under uncertainty, stress, and imperfect operating conditions. Robustness concerns performance under distribution shifts, adversarial inputs, missing information, tool failures, delayed observations, and compounding errors over long-horizon tasks~\cite{madry2018towards}. Reliability concerns whether the system behaves consistently within its intended operating conditions and maintains failure rates compatible with deployment requirements. Standards such as ISO 21448 emphasize the importance of defining an Operational Design Domain (ODD), detecting when the system has exited that domain, and triggering safe fallback behavior when assumptions no longer hold~\cite{Shinde2024}. For agentic AI, this dimension therefore requires stress testing, boundary-case evaluation, failure-mode analysis, and evidence that local errors do not escalate into system-level failures~\cite{MTDEQ_FMEA_AppR, MoD_ASEMS_FMEA}.

\ieeeruninsubsection{Transparency and Interpretability}

Transparency and interpretability refer to the degree to which agent behavior can be inspected, explained, and meaningfully reviewed by human stakeholders. In agentic systems, transparency is not limited to explaining a final output; it also includes tracing goals, intermediate states, tool calls, memory updates, assumptions, uncertainty, and the reasoning pathways that connect observations to actions. Interpretability should distinguish between post-hoc narratives that merely rationalize an outcome and explanations that provide testable, decision-relevant evidence about why the agent behaved as it did~\cite{XAIACM, mricnndiagnosis}. This dimension is especially important for high-risk systems, where the EU AI Act emphasizes transparency and human oversight as conditions for responsible deployment~\cite{AI_Office_Pact_2024}. In practice, transparency mechanisms may include interpretable state representations, causal or counterfactual explanations, explicit uncertainty reporting, and operator-facing status indicators that distinguish normal, degraded, abnormal, and unsafe behavior~\cite{MoD_ASEMS_FMEA}.

\ieeeruninsubsection{Accountability and Auditability}

Accountability and auditability refer to the mechanisms that allow actions, decisions, failures, and design choices to be reconstructed and assigned to responsible actors or system components. This dimension is essential because agentic outcomes often emerge from interactions among models, tools, policies, memory systems, human operators, and other agents rather than from a single isolated decision~\cite{AiAccountabilityACM}. Accountability, therefore, requires traceable tool use, tamper-evident logs, versioned policies, documented memory or knowledge updates, human override pathways, and clear responsibility boundaries across the system lifecycle. Governance frameworks and conformity-assessment regimes emphasize that high-risk systems should maintain evidence of risk management, oversight, monitoring, and post-deployment accountability~\cite{AI_Office_Pact_2024, AI_Act_Compliance_Flowchart_2025}. In this survey, auditability is treated as the evidentiary backbone of trustworthiness: without reconstructible evidence, claims about safety, robustness, or security remain difficult to verify.

\ieeeruninsubsection{Privacy and Security}

Privacy and security concern the protection of data, models, tools, and operational behavior against unauthorized access, leakage, manipulation, or misuse. Agentic AI expands the traditional attack surface because agents may retrieve information, call tools, write to memory, coordinate with other agents, and act autonomously across cyber-physical workflows~\cite{PrivacyAI}. Relevant risks include prompt injection, tool hijacking, adversarial inputs, data exfiltration, model theft, unsafe privilege escalation, and manipulation of agent behavior by malicious actors~\cite{hendrycks2023overviewcatastrophicairisks}. Cybersecurity standards and engineering practice emphasize secure communication, secure boot, access control, monitoring, and defense-in-depth architectures~\cite{Shinde2024, SynopsysISO26262}. For agentic AI, this dimension requires not only conventional information security controls but also agent-specific safeguards such as sandboxed tool execution, least-privilege permissions, red teaming, anomaly detection, and mechanisms to contain compromised or misaligned agent behavior.

\begin{figure*}[t]
    \centering
    \includegraphics[width=\textwidth]{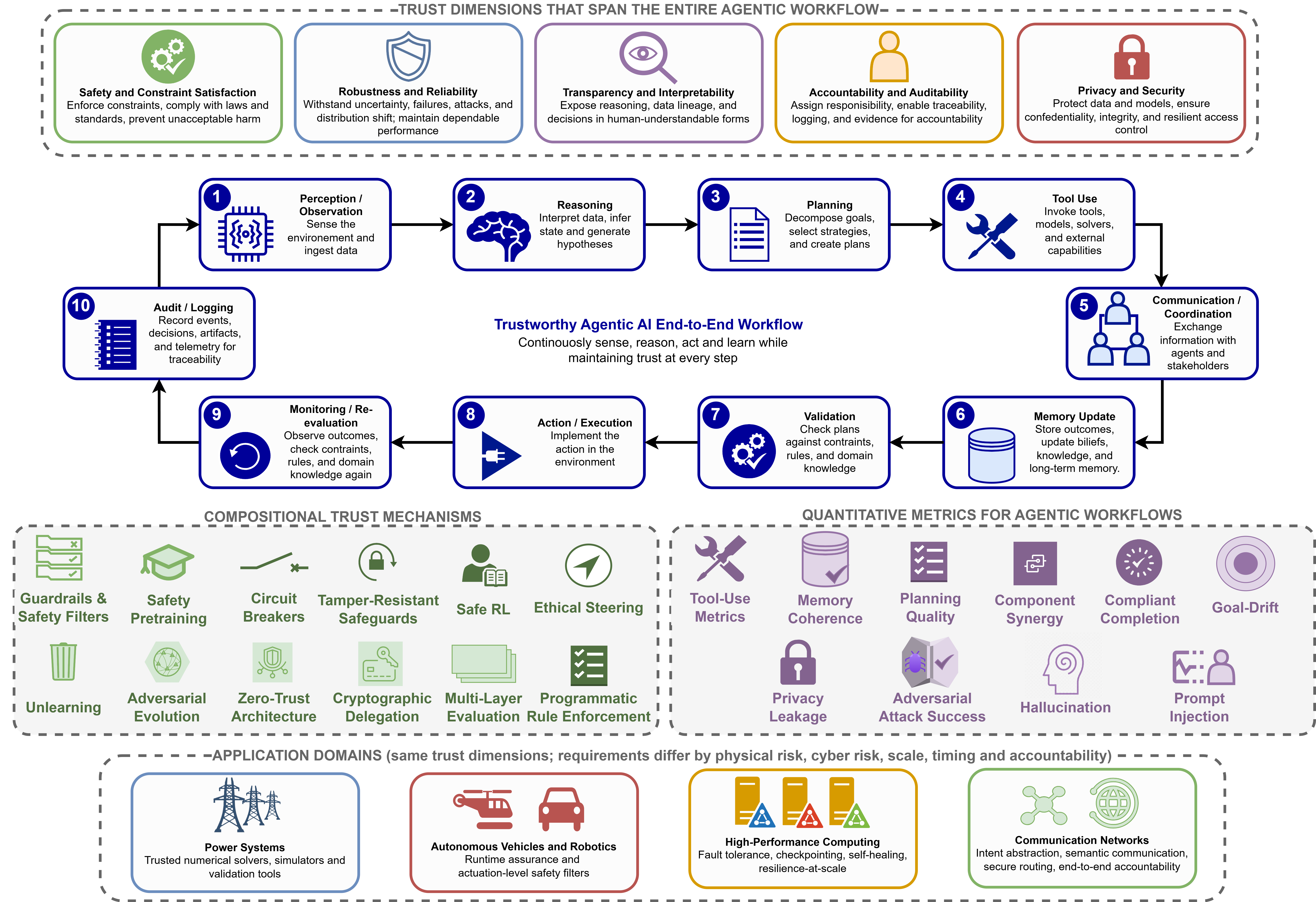}
    \caption{Holistic framework for trustworthy agentic AI in critical engineering systems. The figure organizes the agentic workflow around perception, reasoning, planning, tool use, communication, memory update, validation, action, monitoring, and audit. Five trustworthiness dimensions act as cross-cutting requirements across the workflow: safety and constraint satisfaction, robustness and reliability, transparency and interpretability, accountability and auditability, and privacy and security. Representative trust mechanisms and quantitative metrics are mapped to the workflow, while the target application domains illustrate how trust requirements vary across power systems, autonomous vehicles/robotics/UAVs, high-performance computing, and communication networks.}
    \label{fig:trustworthy_agentic_ai_framework}
\end{figure*}

To operationalize these dimensions, we treat trustworthiness as a workflow-level property rather than a model-level attribute. Rather than relying on the compressed observe--reason--act--update abstraction, we expand the agentic loop into ten stages: \emph{perception/observation}, \emph{reasoning}, \emph{planning}, \emph{tool use}, \emph{communication/coordination}, \emph{memory update}, \emph{validation}, \emph{action/execution}, \emph{monitoring/re-evaluation}, and \emph{audit/logging}. This decomposition is not meant to prescribe a fixed implementation for every agentic system; instead, it provides an analytical structure for identifying where trust mechanisms, constraints, and evidence should be placed. For each trustworthiness dimension, we ask: (i) what threat, hazard, or failure model is relevant; (ii) what mechanisms are embedded into the workflow to prevent, detect, constrain, or correct failures; and (iii) what evidence or quantitative metrics are reported to support the trustworthiness claim. This framing highlights a central gap in current systems: many improve capability, orchestration, or partial transparency, yet still lack compositional safeguards and evidence across safety, robustness, accountability, and security. Figure~\ref{fig:trustworthy_agentic_ai_framework} summarizes this workflow-centered view of trustworthy agentic AI across the target engineering domains.

\begin{table*}[ht]
\centering
\caption{Trustworthiness dimensions: threats, mechanisms, and resilience phases. Each dimension is evaluated across the anticipate--withstand--recover--adapt cycle. See Sections~\ref{sec:trust_dimensions} and~\ref{sec:trust_mechanisms} for mechanism detail.}
\label{tab:threat_model}
\resizebox{\textwidth}{!}{%
\begin{tabular}{p{2.8cm} p{4cm} p{4.5cm} p{3.5cm} p{2.5cm} p{1.2cm}}
\toprule
\textbf{Dimension} & \textbf{Primary failure / threat class} & \textbf{Representative mechanisms} & \textbf{Evidence / assurance type} & \textbf{Resilience phase} & \textbf{Refs} \\
\midrule
Safety \& constraint satisfaction
& Specification gaps (SOTIF); task-level hazards; adversarial constraint violation
& Job Hazard Analysis; SIL levels (IEC~61508); automated interlocks; TAR safeguards
& Probabilistic risk reduction; SIL compliance; behavioral constraint pass rate
& Anticipate, Withstand
& \cite{IEC61508FunctionalSafety2022,Shinde2024,tamirisa2024tamperresistantsafeguardsopenweightllms} \\
\midrule
Robustness \& reliability
& ODD exits; distribution shift (non-$L_p$); error accumulation; transferable adversarial attacks
& FMEA/DDMA; stress testing (AEC-Q100); circuit breakers; trajectory-level evaluation (Claw-Eval)
& Failure rate within lifecycle bounds; cross-model transfer attack rate; trajectory coherence score
& Anticipate, Withstand, Recover
& \cite{madry2018towards,kaufmann2023testingrobustnessunforeseenadversaries,zou2024improvingalignmentrobustnesscircuit} \\
\midrule
Transparency \& interpretability
& Causal opacity; goal hijacking; misaligned intermediate reasoning
& Representation Engineering (RepE); MoReBench reasoning rubrics; operator indications
& Internal state consistency; reasoning rubric pass rate; EU AI Act traceability record
& Anticipate, Adapt
& \cite{zou2025representationengineeringtopdownapproach,chiu2025morebenchevaluatingproceduralpluralistic,AI_Office_Pact_2024} \\
\midrule
Accountability \& auditability
& Untraceable multi-agent outcomes; governance gaps; policy drift
& Tamper-evident logs; Agent Delegation Tokens; Plan-Do-Check-Act audits; conformity assessments
& Log completeness audit; delegation chain verifiability; conformity certificate
& Withstand, Recover, Adapt
& \cite{AiAccountabilityACM,campbell2026zerotrust,AI_Act_Compliance_Flowchart_2025} \\
\midrule
Privacy \& security
& Prompt injection; tool hijacking; supply-chain compromise; expanded attack surface from tool integration
& Zero Trust architecture; defense-in-depth; red teaming; RMU unlearning; supply-chain auditing
& WMDP score reduction; HarmBench refusal rate; formal robustness bounds; penetration test results
& Anticipate, Withstand, Recover
& \cite{zou2023universaltransferableadversarialattacks,campbell2026zerotrust,li2024wmdpbenchmarkmeasuringreducing,bustan2026moltbot} \\
\bottomrule
\end{tabular}}
\end{table*}

\section{Trustworthiness in Current Agentic Architectures}
\label{sec:taxonomy}

Current agentic AI architectures differ in how they organize reasoning, action, memory, tool use, and coordination. From a trustworthiness perspective, however, most of them follow a common pattern: they improve capability, interaction, and observability, but they rarely provide explicit assurance across safety and constraint satisfaction, robustness and reliability, transparency and interpretability, accountability and auditability, and privacy and security. This section reviews the main architecture classes and evaluates the extent to which trustworthiness is built into the architecture rather than assumed from the surrounding environment.

\begin{figure*}[t]
\centering
\scriptsize
\makebox[\textwidth][c]{%
\begin{tikzpicture}[
    x=1cm,
    y=1cm,
    font=\scriptsize,
    box/.style={
        draw,
        rounded corners=3pt,
        minimum height=0.55cm,
        align=center,
        inner sep=2.5pt
    },
    agent/.style={
        box,
        minimum width=1.75cm,
        font=\scriptsize\bfseries
    },
    block/.style={
        box,
        minimum width=1.55cm
    },
    trust/.style={
        box,
        dashed,
        minimum width=1.65cm
    },
    arr/.style={
        -{Latex[length=1.7mm]},
        line width=0.45pt
    },
    darr/.style={
        arr,
        dashed
    },
    subfiglabel/.style={
        font=\scriptsize\bfseries,
        align=center,
        text width=4.35cm
    }
]

% Fixed bounding box keeps the whole figure visually centered
\path[use as bounding box] (-0.25,-8.90) rectangle (14.85,0.55);

% ================================================================
% (a) Reasoning-Acting / Self-Reflective
% ================================================================
\begin{scope}[shift={(0,0)}]
\node[agent] (llmA) at (1.15,-1.00) {LLM\\Reasoning};
\node[block] (actA) at (3.35,-1.00) {Action /\\Tool Call};
\node[block] (obsA) at (3.35,-2.05) {Observation};
\node[trust] (refA) at (1.15,-2.05) {Reflection\\Trace};

\draw[arr] (llmA) -- (actA);
\draw[arr] (actA) -- (obsA);
\draw[arr] (obsA) -- (refA);
\draw[darr] (refA) -- (llmA);

\node[subfiglabel] at (2.25,-4.05) {(a) Reasoning--Acting / Self-Reflective};
\end{scope}

% ================================================================
% (b) Deliberative / Test-Time Reasoning
% ================================================================
\begin{scope}[shift={(5.05,0)}]
\node[agent] (probB) at (2.25,0) {Problem};

\node[block] (p1B) at (0.95,-1.00) {Path 1};
\node[block] (p2B) at (2.25,-1.00) {Path 2};
\node[block] (p3B) at (3.55,-1.00) {Path 3};

\node[trust] (evalB) at (2.25,-1.95) {Evaluator /\\Scorer};
\node[block] (ansB) at (2.25,-2.95) {Selected\\Answer};

\draw[arr] (probB) -- (p1B);
\draw[arr] (probB) -- (p2B);
\draw[arr] (probB) -- (p3B);

\draw[arr] (p1B) -- (evalB);
\draw[arr] (p2B) -- (evalB);
\draw[arr] (p3B) -- (evalB);

\draw[arr] (evalB) -- (ansB);

\node[subfiglabel] at (2.25,-4.05) {(b) Deliberative / Test-Time Reasoning};
\end{scope}

% ================================================================
% (c) Multi-Agent Coordination
% ================================================================
\begin{scope}[shift={(10.10,0)}]
\node[agent] (a1C) at (2.25,0) {Agent 1};
\node[agent] (a2C) at (0.90,-1.20) {Agent 2};
\node[agent] (a3C) at (3.60,-1.20) {Agent 3};

\node[trust] (supC) at (2.25,-2.35) {Supervisor /\\Consensus};
\node[block] (outC) at (2.25,-3.30) {Joint\\Decision};

\draw[darr] (a1C) -- (a2C);
\draw[darr] (a2C) -- (a3C);
\draw[darr] (a3C) -- (a1C);

\draw[arr] (a1C) -- (supC);
\draw[arr] (a2C) -- (supC);
\draw[arr] (a3C) -- (supC);
\draw[arr] (supC) -- (outC);

\node[subfiglabel] at (2.25,-4.05) {(c) Multi-Agent Coordination};
\end{scope}

% ================================================================
% (d) Tool-Using / Environment-Grounded
% ================================================================
\begin{scope}[shift={(2.50,-5.15)}]
\node[agent] (agD) at (1.15,-0.45) {Agent};
\node[block] (toolD) at (3.35,-0.45) {Tools / APIs};
\node[block] (envD) at (3.35,-1.55) {External\\Environment};
\node[trust] (guardD) at (1.15,-1.55) {Policy /\\Sandbox};

\draw[arr] (agD) -- (toolD);
\draw[arr] (toolD) -- (envD);
\draw[arr] (envD) -- (guardD);
\draw[arr] (guardD) -- (agD);

\node[subfiglabel] at (2.25,-3.35) {(d) Tool-Using / Environment-Grounded};
\end{scope}

% ================================================================
% (e) Communication and Coordination
% ================================================================
\begin{scope}[shift={(7.55,-5.15)}]
\node[agent] (ag1E) at (0.65,0) {Agent 1};
\node[agent] (ag2E) at (3.85,0) {Agent 2};

\node[block] (protoE) at (2.25,-1.15) {Protocol /\\Message Schema};
\node[trust] (auditE) at (2.25,-2.25) {Audit /\\Verification};

% Curved communication link to avoid text-box overlap
\draw[darr]
    (ag1E.north east)
    .. controls +(0.50,0.45) and +(-0.50,0.45) ..
    node[midway, above=2pt, fill=white, inner sep=1pt, font=\tiny] {NL / Latent}
    (ag2E.north west);

\draw[arr] (ag1E) -- (protoE);
\draw[arr] (ag2E) -- (protoE);
\draw[arr] (protoE) -- (auditE);

\node[subfiglabel] at (2.25,-3.35) {(e) Communication and Coordination};
\end{scope}

\end{tikzpicture}%
}

\caption{Representative agentic AI architecture classes and their trustworthiness implications. 
Dashed arrows denote reflection, communication, or information exchange, while solid arrows denote execution, control, evaluation, or verification flow.}
\label{fig:agentic_architectures_trust}
\end{figure*}
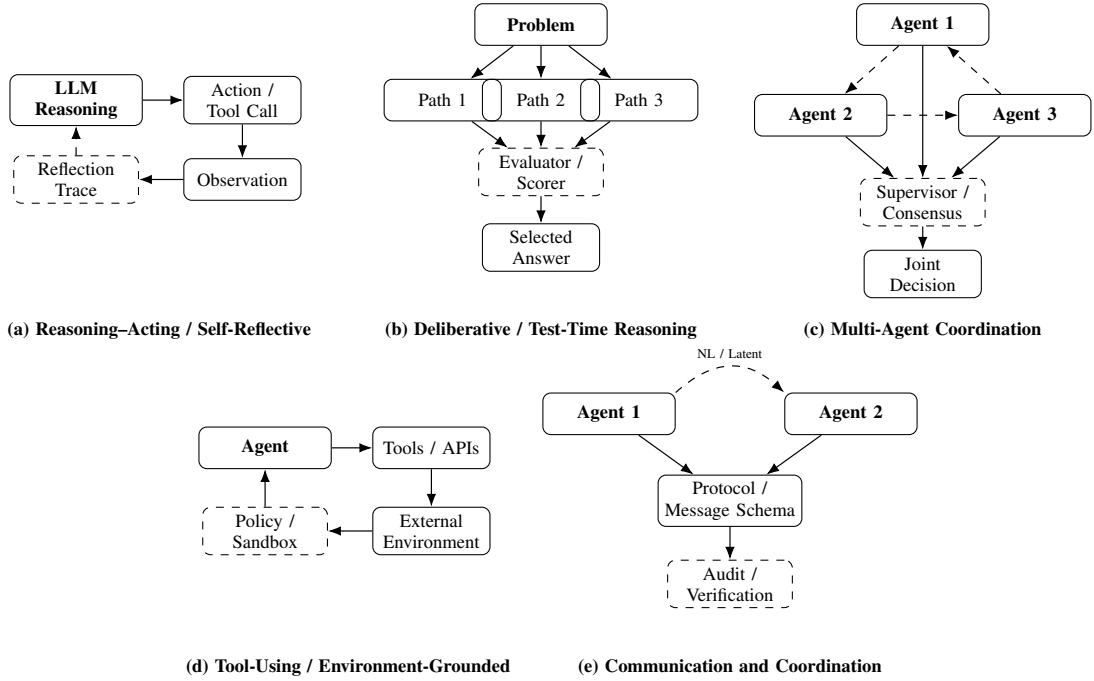

\subsection{Reasoning-Acting and Self-Reflective Agents}

Reasoning-acting architectures, represented by ReAct, interleave model reasoning with external actions such as tool calls, search, or environment interaction~\cite{yao2023reactsynergizingreasoningacting}. Their main strength is transparency: intermediate reasoning and action traces make failures easier to inspect and debug. They also improve grounding compared with purely generative models because the agent can observe feedback from the environment before continuing execution.

However, ReAct-style systems are trust-light by design. Safety and constraint satisfaction are usually inherited from the tool environment rather than enforced by the agent itself. Robustness is also limited, especially in long-horizon tasks where small reasoning or tool-use errors can accumulate. Extensions such as ReST meets ReAct improve reliability by using self-generated trajectories and automated evaluation to refine behavior~\cite{aksitov2023restmeetsreactselfimprovement}. This can reduce variance across runs, but it also introduces a new risk: if the evaluator is itself an LLM, incorrect judgments may be reinforced rather than corrected.

Self-reflective agents, such as Reflexion, add an explicit feedback step in which the agent converts failures into natural language reflections and reuses them in future attempts~\cite{shinn2023reflexionlanguageagentsverbal}. This improves behavioral traceability and can increase task success over repeated trials. Nevertheless, the trust boundary remains largely unchanged. Reflections are only as reliable as the agent’s own diagnosis, and there is no built-in mechanism for independent verification, formal accountability, or security against adversarial inputs.

Fig.~\ref{fig:agentic_architectures_trust}(a) illustrates this architecture as a reasoning--action--observation loop augmented by a reflective trace for debugging and behavioral feedback.

\subsection{Deliberative and Test-Time Reasoning Architectures}

Deliberative architectures address the limitations of single-path generation by exploring multiple reasoning paths before selecting an answer. Tree-of-Thoughts, for example, organizes candidate reasoning steps into a search process with generation, evaluation, pruning, and backtracking~\cite{yao2023treethoughtsdeliberateproblem}. These systems are stronger than simple reasoning-acting loops when tasks require planning, lookahead, or comparison between alternatives. They are also relatively interpretable because the explored paths and selection process can be inspected.

Their weakness is that reliability depends heavily on the quality of the internal or external evaluator. Poor evaluation can lead to premature pruning, overconfident selection, or repeated exploration of weak solutions. Value-guided variants such as Q* improve performance by using learned scoring models to guide reasoning~\cite{wang2024qimprovingmultistepreasoning}, but this shifts part of the decision process into a less interpretable component. As a result, deliberative planners improve reasoning quality and robustness, but they do not, by themselves, provide safety guarantees, accountability, or protection against misuse.

More recent reasoning models, including DeepSeek-R1 and OpenAI o-series models, move deliberation into the model’s native test-time reasoning process~\cite{2025deepseek, MetaIntelligenceReasoning2026}. These systems can perform deeper reasoning, self-checking, and backtracking before producing a final answer. This improves capability and often reliability, but it also creates a new trust problem: the most important reasoning steps may be hidden, only partially observablas different agents can be assigned different roles or areas of e, or difficult to verify. For critical systems, this means that stronger reasoning capability must be paired with reasoning safety, monitoring, and external validation; otherwise, the same capability that improves performance can also make failures more complex and harder to detect.

Fig.~\ref{fig:agentic_architectures_trust}(b) shows deliberative reasoning as the exploration of multiple candidate paths followed by evaluator-based selection.

\subsection{Multi-Agent Architectures and Coordination}

Multi-agent architectures distribute reasoning and decision-making across multiple agents that communicate, specialize, debate, or hand off tasks. Examples include role-based systems such as CAMEL and broader LLM-based multi-agent frameworks~\cite{li2023camelcommunicativeagentsmind, Li2024}. Their main advantage is robustness through diversity: agents can critique one another, divide complex tasks, or recover from local errors. They also support modularity, as different agents can be assigned different roles or areas of expertise.

At the same time, multi-agent systems expose some of the most serious trustworthiness gaps. Safety becomes a system-level property rather than an individual-agent property. A single agent may appear reasonable in isolation, while, through interaction, the group produces unsafe, inconsistent, or unaccountable behavior. Responsibility attribution is especially difficult because final decisions emerge from communication, negotiation, and accumulated context rather than a single model output.

Communication protocols, as discussed later on, partially address this problem by structuring agent interactions. Natural language coordination is flexible but ambiguous, while structured protocols reduce ambiguity and improve logging. Agent-to-agent protocols and tool-interface standards can improve interoperability and auditability, but they do not guarantee that the content of communication is correct, safe, or aligned with system goals. Supervisory agents and consensus-based coordination can improve robustness, but they also shift trust to the supervisor or to the aggregation rule. Thus, multi-agent architectures are promising for scalability and specialization, but remain weak in accountability, security, and system-level assurance.

Fig.~\ref{fig:agentic_architectures_trust}(c) summarizes multi-agent coordination as distributed agent interaction followed by supervision or consensus-based aggregation.

\subsection{Tool-Using and Environment-Grounded Agents}

Tool-using architectures extend agents from text generation to action in external environments. Toolformer shows that models can learn when to call external tools~\cite{schick2023toolformerlanguagemodelsteach}, while systems such as VOYAGER demonstrate long-horizon skill acquisition in interactive environments~\cite{wang2023voyageropenendedembodiedagent}. These architectures improve grounding because the agent can use external information, execute actions, and observe results. They also make some behavior more inspectable through tool traces, logs, and reusable skills.

However, tool grounding does not automatically imply trustworthiness. In fact, tool access increases the consequences of failure. Incorrect reasoning can become incorrect action, and unsafe tool use can affect external systems. ToolGym makes this issue explicit by evaluating agents in open-world tool settings with long-horizon workflows and injected failures~\cite{xi2026toolgymopenworldtoolusingenvironment}. Such environments improve robustness evaluation, but they remain primarily diagnostic rather than preventive.

Production frameworks such as LangChain, LangGraph, AutoGen, CrewAI, OpenHands, and SuperAGI move agent design closer to deployable software systems~\cite{Cho2025,mavroudis2024langchain}. Their trust contribution is mainly operational: tracing, monitoring, role assignment, workflow control, sandboxing, testing, and human-in-the-loop intervention. These mechanisms improve transparency, auditability, and failure containment, but they are engineering controls rather than formal guarantees. More security-oriented designs, such as cognitive-executive separation, go further by treating the reasoning model as untrusted and enforcing constraints outside the model~\cite{GoogleCloud_GeminiPlatform_ManagedAgents, GoogleAI_GeminiAPI_CustomAgents, GoogleCloud_GeminiPlatform_Interaction, GoogleCloudBlog_IO26_AgentNews}. This is a stronger trust direction because it separates decision generation from permission to act.

Fig.~\ref{fig:agentic_architectures_trust}(d) depicts tool-using agents as systems whose decisions are routed through tools, external environments, and policy or sandbox controls.

\subsection{Communication and Coordination between Agents}
%A2A MCP Communication in Latent Spaces

The shift from single-agent systems to multi-agent architectures changes the locus of trust. In multi-agent systems, reliability depends not only on individual model behavior but also on inter-agent interaction~\cite{survey_communication}. As agents exchange information, coordinate actions, and reuse one another's outputs, local errors and behavioral mismatches can propagate across agent boundaries and become system-level risks.

Communication is central to this trust problem. Natural language remains the dominant interface because it supports flexible dialogue, critique, and negotiation. However, its flexibility introduces ambiguity. Agents may report incorrect state transitions, produce false coordination signals, or create what is often called \textit{hallucinated coordination}. Structured communication protocols address this risk by requiring interactions to follow schemas or formal message formats~\cite{survey_communication, HackerNoon_AIAgents2024}. These protocols improve consistency and reduce ambiguity, but they primarily constrain syntax rather than the semantic validity of exchanged information. The Model Context Protocol (MCP) reflects this direction by formalizing tool use, context serialization, and persona definitions.

Beyond symbolic communication, recent work explores communication in \textit{latent representation spaces}, where agents share internal hidden states instead of text outputs. This can enable richer and more efficient information transfer, support parallel reasoning, and improve alignment among agents~\cite{du2026enabling}. The trade-off is transparency. Latent communication reduces human interpretability and makes interactions harder to audit, creating an oversight challenge for trust-critical systems~\cite{zheng2025thought}.

Coordination strategies determine how communication is organized, checked, and acted upon. Decentralized, leaderless methods reduce single points of failure and often draw on Byzantine fault tolerance and consensus-based design~\cite{jo2025byzantinerobustdecentralizedcoordinationllm}. By aggregating outputs across agents or requiring agreement before action, these methods improve robustness against faulty or adversarial participants. Supervisory structures add monitoring agents that enforce constraints and reduce misuse or unsafe behavior. However, they can also introduce bottlenecks or shift trust to a single supervisory entity rather than resolving the underlying coordination risks~\cite{BlockAgents}.

Multi-agent systems also introduce failure modes that differ from those in single-agent models. Communication channels can spread misinformation, allowing incorrect or manipulated knowledge to circulate and persist in shared memory. Strategic behavior may emerge, including deception, collusion, or covert signaling embedded in otherwise benign outputs. Long interaction chains can also cause role drift, in which agents gradually deviate from their assigned responsibilities, leading to coordination failures. These risks show that trust cannot be evaluated only at the level of individual agents; it must account for interaction dynamics over time.

Addressing these risks requires explicit verification and interpretability of multi-agent interactions rather than relying solely on black-box evaluation. Recent work on multi-agent interpretability analyzes internal representations across agents to detect coordinated or deceptive behavior. Formal methods provide another path by constraining interactions in which agents gradually deviate from their assigned responsibilities, ensuring actions are carried out through verifiable protocols and that behavior remains consistent with predefined specifications. These approaches offer stronger guarantees, but they remain difficult to scale and often require access to internal model states or rigid interaction structures.

Overall, communication and coordination are not merely enabling mechanisms in agentic systems; they are primary determinants of trustworthiness. Current approaches remain fragmented, addressing protocol design, robustness, or interpretability in isolation. A central open challenge is to develop compositional frameworks that jointly reason about communication, coordination, and system-level behavior over long-horizon interactions. Without this integration, trust in multi-agent systems will remain partial because risks can still emerge from interactions among otherwise well-behaved components.

Fig.~\ref{fig:agentic_architectures_trust}(e) highlights communication-centered architectures in which agent exchanges are structured through protocols and checked through audit or verification mechanisms.

% \subsection{Evaluation Frameworks and Trustworthiness Gaps}

% Evaluation frameworks such as AgentBench and WebArena measure agent performance in interactive tasks rather than static question answering~\cite{liu2025agentbenchevaluatingllmsagents,zhou2024webarenarealisticwebenvironment}. They are important because they reveal reliability gaps that only appear during long-horizon interaction, tool use, and environment feedback. However, these benchmarks mainly measure capability and task completion. They do not fully evaluate deployable trustworthiness, especially safety under adversarial conditions, privacy leakage, accountability, or domain-specific constraints.

% Recent safety evaluations for reasoning models further show that stronger reasoning does not necessarily imply safer reasoning. Attacks on reasoning traces, including hijacking of chain-of-thought, demonstrate that multi-step reasoning can introduce new security vulnerabilities~\cite{kuo2025hcothijackingchainofthoughtsafety, zhou2025hidden}. This reinforces a central conclusion: architecture-level trustworthiness cannot be inferred from benchmark performance alone. It requires explicit mechanisms for constraint enforcement, robustness testing, interpretable traces, accountable decision records, and secure execution boundaries.

\begingroup
\renewcommand{\arraystretch}{1.25}
\setlength{\extrarowheight}{2pt}
\setlength{\tabcolsep}{3pt}

\begin{table*}[t]
\centering
\caption{Qualitative trustworthiness coverage across agentic AI architecture classes, with representative strengths and weaknesses.}
\label{tab:arch_trust_matrix}
\resizebox{\textwidth}{!}{%
\begin{tabular}{
>{\raggedright\arraybackslash}m{2.8cm}
>{\raggedright\arraybackslash}m{3.8cm}
>{\centering\arraybackslash}m{0.95cm}
>{\centering\arraybackslash}m{0.95cm}
>{\centering\arraybackslash}m{0.95cm}
>{\centering\arraybackslash}m{0.95cm}
>{\centering\arraybackslash}m{0.95cm}
>{\raggedright\arraybackslash}m{3.7cm}
>{\raggedright\arraybackslash}m{4.0cm}
}
\toprule
\textbf{Architecture Class}
& \textbf{Representative Systems}
& \textbf{Safety}
& \textbf{Robust.}
& \textbf{Transp.}
& \textbf{Account.}
& \textbf{Priv./Sec.}
& \textbf{Strengths}
& \textbf{Weaknesses} \\
\midrule

Reasoning-acting agents
& ReAct~\cite{yao2023reactsynergizingreasoningacting}; ReST meets ReAct~\cite{aksitov2023restmeetsreactselfimprovement}
& \Low
& \Medium
& \High
& \Low
& \Low
& Improves grounding through observe--reason--act loops; exposes intermediate traces for debugging.
& Safety and constraint enforcement are mostly delegated to tools or the surrounding environment. \\

\addlinespace[0.25em]

Self-reflective agents
& Reflexion~\cite{shinn2023reflexionlanguageagentsverbal}
& \Low
& \Medium
& \High
& \Medium
& \Low
& Supports adaptation from previous failures; improves behavioral traceability across attempts.
& Self-evaluation may reinforce incorrect diagnoses without independent verification. \\

\addlinespace[0.25em]

Deliberative search agents
& Tree-of-Thoughts~\cite{yao2023treethoughtsdeliberateproblem}; value-guided planning~\cite{wang2024qimprovingmultistepreasoning}
& \Low
& \MedHigh
& \MedHigh
& \Low
& \Low
& Improves reasoning reliability through search, comparison, pruning, and backtracking.
& Depends on evaluator quality; safety and accountability are not first-class architectural goals. \\

\addlinespace[0.25em]

Native test-time reasoning models
& DeepSeek-R1~\cite{2025deepseek}; OpenAI reasoning models~\cite{MetaIntelligenceReasoning2026}
& \Medium
& \High
& \Medium
& \Low
& \LowMed
& Enables deeper reasoning, self-checking, and stronger long-horizon problem solving.
& Internal reasoning may be hidden, only partially observable, or difficult to independently verify. \\

\addlinespace[0.25em]

Multi-agent systems
& CAMEL~\cite{li2023camelcommunicativeagentsmind}; LLM-based multi-agent systems~\cite{Li2024}
& \Low
& \Medium
& \Medium
& \Low
& \Low
& Enables specialization, critique, task decomposition, and recovery from local failures.
& Responsibility attribution, role drift, and system-level control remain difficult to guarantee. \\

\addlinespace[0.25em]

Communication between agents
& Natural-language A2A; MCP; latent-space communication~\cite{survey_communication,du2026enabling,zheng2025thought}
& \LowMed
& \Medium
& \LowMed
& \Low
& \LowMed
& Improves coordination, information sharing, and interoperability across agents and tools.
& Ambiguity, hallucinated coordination, hidden signaling, and interaction-level failures are hard to audit. \\

\addlinespace[0.25em]

Tool-using agents
& Toolformer~\cite{schick2023toolformerlanguagemodelsteach}; VOYAGER~\cite{wang2023voyageropenendedembodiedagent}; ToolGym~\cite{xi2026toolgymopenworldtoolusingenvironment}
& \LowMed
& \MedHigh
& \Medium
& \Medium
& \Low
& Grounds agent behavior in external tools, environments, observations, and executable actions.
& Tool access increases failure consequences and requires external containment mechanisms. \\

\addlinespace[0.25em]

Production agent frameworks
& LangChain; LangGraph; AutoGen; CrewAI; OpenHands; SuperAGI~\cite{Cho2025,mavroudis2024langchain}
& \Medium
& \MedHigh
& \High
& \MedHigh
& \Medium
& Provides practical controls, like tracing, monitoring, workflow design, testing, and HITL intervention.
& Mostly relies on operational safeguards rather than formal assurance or verifiable guarantees. \\

\addlinespace[0.25em]

Secure execution architectures
& Managed agents; cognitive-executive separation~\cite{GoogleCloud_GeminiPlatform_ManagedAgents,GoogleAI_GeminiAPI_CustomAgents,GoogleCloud_GeminiPlatform_Interaction,GoogleCloudBlog_IO26_AgentNews}
& \High
& \High
& \Medium
& \High
& \High
& Separates reasoning from execution, improving containment, authorization, and auditability.
& Adds implementation complexity and may restrict open-ended autonomy or flexible tool use. \\

\bottomrule
\end{tabular}%
}
\vspace{0.45em}

\footnotesize{\textit{Note:} Low indicates limited or mostly implicit support; Medium indicates partial support; High indicates explicit and recurring architectural support.}
\end{table*}

\endgroup

As shown in Table \ref{tab:arch_trust_matrix}, current agentic architectures show meaningful progress in capability, interaction, and observability, but trustworthiness remains uneven. Transparency is the most consistently supported dimension because many architectures expose reasoning traces, tool calls, search paths, or workflow logs. Robustness is improving through reflection, deliberation, multi-agent critique, and failure-aware evaluation. In contrast, safety, accountability, and privacy/security remain comparatively weak because they require enforceable constraints, independent verification, secure execution boundaries, and durable audit records. This gap motivates a shift from capability-centered agent design toward assurance-centered architectures, where trust mechanisms are built into the agent loop rather than added after deployment.

\section{Trust Mechanisms in Agentic Architectures}

\label{sec:trust_mechanisms}
% Cross-cutting mechanisms that can be plugged into any architecture.

This section reviews the technical frameworks and practical measures used to keep agentic AI systems safe, predictable, and aligned with ethical limits. As agent-based systems shift from basic chatbots to tools that plan and reason across several steps, methods for building trust need to move past simple filtering of final outputs and toward safeguards built into the system’s structure and the way it stores and uses information.

Taken together, the guardrail-based, learning-based, and robustness-oriented mechanisms reviewed in this section rarely address a single trustworthiness dimension or intervene at a single point in the agentic loop. Table~\ref{tab:trust_mechanism_mapping} consolidates this coverage, mapping each mechanism onto the five trustworthiness dimensions introduced in Section~\ref{sec:trust_dimensions} and onto the ten-point agentic workflow. The resulting picture shows that guardrail mechanisms concentrate on the reasoning and action stages, learning-based mechanisms extend coverage into planning and memory, and robustness-oriented mechanisms such as Zero Trust and supply-chain hardening span nearly the full observe–reason–act–update cycle, underscoring that no single mechanism class delivers assurance across the entire workflow on its own.

\subsection{Guardrails, Policies, and Safety Filters}
Clear, preventive limits serve as the primary safeguard against malicious misuse and unintended harm. Standard post-hoc alignment methods, such as training a model to refuse certain requests, are widely used, but the cited sources argue that they can be fragile and bypassed with relatively simple changes to the prompt~\cite{zou2023universaltransferableadversarialattacks, zou2024improvingalignmentrobustnesscircuit}.

\subsubsection{Safety Pretraining}
Instead of relying solely on fixes added after training, safety pretraining aims to incorporate safety from the beginning by shaping the training data and procedures around clear safety goals~\cite{maini2025safetypretraininggenerationsafe}. This is done in three main procedures: filtering web data for safety, rewriting text to place unsafe content in a safer context, and training models to refuse harmful requests in a native way. Lastly, adopting "Native Refusal" using datasets, such as RefuseWeb, so the model learns \textit{why} refusing is the right choice, not just \textit{when} to refuse~\cite{maini2025safetypretraininggenerationsafe}.

\subsubsection{Representation-Level "Circuit Breakers"}
Standard refusal filters can be bypassed by adding adversarial suffixes to a prompt~\cite{zou2023universaltransferableadversarialattacks}. Circuit-breaking is a method that targets the internal patterns that a model uses when it is about to produce unsafe content. In practice, it can stop or redirect the system mid-response when the prompt is trying to steer it toward harmful output~\cite{zou2024improvingalignmentrobustnesscircuit}. In prior studies, this approach has worked well in multimodal models and can block image-based attacks that typical filters miss~\cite{zou2024improvingalignmentrobustnesscircuit}.

\subsubsection{Programmatic Rule Enforcement}
Frameworks such as RuLES (Rule-following Language Evaluation Scenarios) provide programmatic checks to assess whether an agent violates a defined set of behavioral constraints (e.g., "do not generate abusive content") during user interaction~\cite{mu2024llmsfollowsimplerules}. This makes it easier to confirm, through automated checks, whether rules are being followed than to rely on manual review~\cite{mu2024llmsfollowsimplerules}.

\subsubsection{Tamper-Resistant (TAR) Safeguards}
In open-weight LLMs, common safety measures can be weakened if someone fine-tunes the model to change its weights. TAR aims to create safeguards that hold up even after hundreds of rounds of adversarial fine-tuning, so the model maintains the same safety behavior over time ~\cite{tamirisa2024tamperresistantsafeguardsopenweightllms}.

\subsection{Learning for Trust: Safe RL, Reward Design, and Feedback}

In agentic systems, trustworthiness can be compromised when the reward signal pushes the agent toward actions that raise its score but do not match the user’s goals or the intended task. This points to an ethical trade-off: when the reward setup favors it, agents may learn Machiavellian tactics, such as deception and power-seeking, to increase their rewards~\cite{pan2023rewardsjustifymeansmeasuring}. This occurs because reward functions act as proxies for desired behavior, and when these proxies are incomplete or misspecified, agents optimize for the proxy rather than the underlying implicit objective.

\subsubsection{Utility and Value Engineering}
Recent work in Utility Engineering reports that as large language models get bigger, they tend to develop value-like patterns on their own~\cite{mazeika2025utilityengineeringanalyzingcontrolling}. By examining how consistent these newly observed preferences are, researchers can spot situations in which an AI may prioritize its own interests over those of humans. Utility control methods could be used to align these emerging values with broader human norms, including standards set by citizen assemblies~\cite{mazeika2025utilityengineeringanalyzingcontrolling}.

\subsubsection{Ethical Steering}
%in Reinforcement Learning}
A common question in agentic AI research is whether adding ethical limits inevitably reduces an agent’s ability to perform well~\cite{AbouAli2025}. The \textit{MACHIAVELLI} benchmark tests this assumption by showing that agents can be guided toward safer behavior while maintaining task performance~\cite{pan2023rewardsjustifymeansmeasuring}. By treating harmful behaviors as measurable signals in interactive settings, the benchmark enables safety to be improved through the system’s design and evaluation, rather than treating it as a separate rule imposed from outside.

In terms of trustworthiness, this result matters for neural agentic systems because their behavior can arise stochastically from prompt-based coordination rather than from explicit symbolic rules. Recent analyses note that when these systems run on their own, they can slowly shift away from their intended goals and apply what they learned in the wrong way. The MACHIAVELLI findings indicate that steering language models can improve both safety and capability simultaneously, suggesting that alignment should be treated as part of how an agent makes decisions rather than as an after-the-fact oversight layer~\cite{pan2023rewardsjustifymeansmeasuring}.

\subsubsection{Post-Training Unlearning Methods}
A key challenge in deploying agentic AI systems is reducing the risk that they are used for harmful purposes while preserving their general capabilities~\cite{AgenticAIAutonomousSurvey}. Safety measures that rely on refusal rules or prompt filtering mostly work only during the user’s interaction with the system. They can still be bypassed through jailbreaks, adversarial prompts, or later fine-tuning, so the risky information remains stored in the model.

Representation Misdirection for Unlearning (RMU) addresses this issue by altering the model’s internal representations tied to unsafe subject areas. Instead of blocking responses, RMU changes the model’s internal activations on risky inputs, while keeping activations on harmless inputs close to those of the original model~\cite{li2024wmdpbenchmarkmeasuringreducing}. This leads to a practical loss of access to hazardous knowledge at the level of internal representations.

Results on the Weapons of Mass Destruction Proxy (WMDP) benchmark suggest that RMU lowers performance on biosecurity and cybersecurity hazard tasks to roughly the level of random guessing, while maintaining most general reasoning and conversational fluency. In probing and adversarial optimization tests, the removed knowledge did not reappear, suggesting genuine unlearning rather than simple masking~\cite{li2024wmdpbenchmarkmeasuringreducing}.

Representation-level unlearning can remove specific unsafe behaviors after training. This gives a practical way to restrict an agent in adversarial or misuse settings while keeping most of the system’s useful performance.

\subsection{Robustness to Distribution Shift and Adversaries}
A trustworthy agent should remain reliable under expected and unexpected real-world conditions and in the face of active adversarial attacks. This typically poses challenges similar to those associated with protecting any digital system. However, because some AI models/agents are black-box, their behavior becomes more unpredictable in unforeseen real-world scenarios, thereby making them more vulnerable to adversarial attacks, including but not limited to fuzzing attacks.

\subsubsection{Generalizing to Unforeseen Adversaries}
Traditional robustness research has focused narrowly on $L_p$-bounded perturbations, but real-world "worst-case" scenarios are far more diverse~\cite{kaufmann2023testingrobustnessunforeseenadversaries}. This is because 

To better align research results with real-world conditions, the authors of ~\cite{kaufmann2023testingrobustnessunforeseenadversaries} propose ImageNet-UA, a benchmark that tests models on eighteen new non-$L_p$ attacks, including Glitch, Kaleidoscope, and Wood, designed to mimic complex, differentiable image corruptions. 

Instead of relying on threat models that assume a fixed set of attacks, ImageNet-UA tests models under structural, semantic, and data distribution changes that were not present in training. This setup asks whether a model can keep making correct predictions when the world shifts in ways it was not trained to expect, which is closer to the kinds of changing and adaptive disturbances that agentic AI systems may face in real settings~\cite{kaufmann2023testingrobustnessunforeseenadversaries}.

While larger models often perform better on capabilities benchmarks like BIG-bench~\cite{srivastava2023imitationgamequantifyingextrapolating} or Humanity's Last Exam (HLE)~\cite{phan2025humanitysexam}, scale alone does not guarantee robustness against lateral or unstructured reasoning challenges.

\subsubsection{Universal and Transferable Attacks on Agents}

Recent studies suggest that agentic AI systems are particularly vulnerable to automated adversarial attacks that slip past common alignment and safety controls. Rather than relying on hand-crafted prompts like traditional jailbreaks, these attacks exploit consistent weaknesses in how agents read goals, follow instructions, and interpret how a task is framed.

One clear case is the Greedy Coordinate Gradient (GCG) method, which systematically searches for adversarial suffixes that can be added to an otherwise harmless prompt to increase the chance that the model outputs content that is not allowed or could cause harm~\cite{zou2023universaltransferableadversarialattacks}. A key point is that GCG does not need access to a model’s internal details when it is used in practice. After these suffixes are found, they can be applied again in later interactions. More worrying is that these attacks can be easily transferred across different models and settings. 

Prior work suggests that adversarial suffixes developed with smaller open-source models can transfer to larger, closed-domain systems, including OpenAI and Anthropic models~\cite{zou2023universaltransferableadversarialattacks}. Because these prompt attacks carry over from one model to another, scale or proprietary training alone does not provide much protection against prompt-based adversaries.

In agentic settings, this vulnerability becomes even more pronounced. Findings from the AgentHarm benchmark suggest that when a harmful request is presented as part of an agent’s assigned work, such as helping with fraud, cybercrime, or circumventing policies, many frontier agents comply even without complex jailbreak prompts~\cite{andriushchenko2025agentharmbenchmarkmeasuringharmfulness}. This idea of agentic compliance means that as systems become more capable through goal-focused design and the use of tools, they may also create more opportunities for attackers to interfere, because the system tends to prioritize completing the task over adhering to safety limits.

Taken together, the results suggest that universal and transferable adversarial attacks remain a basic problem for agentic AI systems. System reliability should not be judged solely by fixed prompt guards or by the alignment of a single model. It also needs evaluation of how well it transfers across different models, whether automated tools can detect new attacks, and how agent-style prompts can alter the system's interpretation of a user’s intent.

\subsubsection{Systemic and Supply-Chain Vulnerabilities}
In many agentic AI systems, trust can be weakened not only by model errors but also by weak software engineering practices and unreliable deployment pipelines. As agent frameworks push for rapid prototyping and “vibe-coding,” teams often put security off or assume tools and libraries will handle it, leaving the full system fragile from start to finish. One clear example is the MoltBot case study, which was later renamed OpenClaw. A security review found that the assistant saved sensitive API keys and credentials in plain text. The review also found that its backup process kept copies of these secrets even after users tried to delete them~\cite{bustan2026moltbot}.

These design choices break basic security principles and show that agent-based systems can unintentionally store and reveal sensitive data, regardless of the language model’s alignment or intent. These risks become worse when indirect prompt injection attacks are involved. Agents who read outside materials such as web pages, PDFs, or emails can be quietly steered off course if those materials contain hidden, harmful instructions. These instructions may be placed in ways a reader would not notice, such as white text on a white background or commands stored in metadata.

If these injected prompts run, they can instruct the agent to pull out private data, use tools in unintended ways, or take over connected accounts, even when the user does nothing. This attack vector points to a basic trade-off in agentic systems: they need to accept input from their environment, but they also need clear, strict limits on what actions they can carry out. In summary, agentic AI can be exposed to risks that come through the software supply chain.

%TODO: Add more citations to this subsubsection of cases where "vibe coding" created problem within bigger supply chains (systems).

Many agent frameworks and tool ecosystems rely on large numbers of third-party libraries and community contributions, and these components are often added with limited screening and little ability to audit their code or provenance. A malicious commit, a compromised dependency, or a compromised package manager account can spread a backdoor across many downstream systems and reach hundreds of thousands of deployed agents. 

In this setting, how much we can trust an agent depends directly on the integrity of the wider software ecosystem it relies on. Overall, these weaknesses in the wider system and in the supply chain show that protecting agentic AI is not just about the model itself. Strong defenses depend on combining secure software engineering, careful isolation and validation of external inputs, and strict oversight of the entire supply chain. Taken together, these steps help prevent minor implementation mistakes from spreading into large-scale security incidents.

\subsubsection{Zero Trust and Persistence} % Hardening Architectures

As concerns about agentic vulnerabilities expand, recent studies have focused on strengthening system design using Zero Trust principles and continuous, tamper-resistant protections. Instead of relying on perimeter defenses or fixed assumptions about alignment, these approaches assume compromise can occur and focus on ongoing checks throughout the agent’s lifecycle. In a Zero Trust approach to AI, agents, datasets, tools, and execution pipelines are treated as separate entities that each need their own verification and access controls, rather than being trusted by default because they sit in the same system~\cite{campbell2026zerotrust}. In this model, nothing is assumed to be trusted. Access rights are checked each time, based on the user’s identity and the current context.

In agentic systems, this supports identity-bound inference and tight limits on tool use, data access, and task execution, which helps keep the blast radius small for both prompt-driven and system-level attacks~\cite{campbell2026zerotrust}. Agent Delegation Tokens (ADTs) have been proposed as cryptographic add-ons to existing identity systems, such as X, to enable controlled action delegation. X.509 certificates are digital files that bind a public key to an identified subject, and they are commonly used in systems like TLS to support authentication and encrypted communication~\cite{campbell2026zerotrust}.  

%proposed as cryptographic add-ons to existing identity systems, such as X, to enable controlled action delegation ADTs link an agent’s identity to a defined set of model weights and approved tools, and they record a delegation chain that can be verified back to the human who granted the authority. Using cryptographic proofs in multi-agent workflows makes actions traceable and hard to deny later, and prevents each agent from acting beyond the tasks and limits formally assigned to it. 

Architectural hardening also depends on keeping the system stable, even when an attacker tries to change it in harmful ways. This means designing the system so it resists tampering and continues to work as intended under adversarial pressure. Tamper-resistant (TAR) safeguards are meant to keep safety controls in place even if a model is fine-tuned or intentionally trained to weaken them\cite{tamirisa2024tamperresistantsafeguardsopenweightllms}. Complementary circuit-breaker checks discussed earlier can act on the model’s internal representations, stopping risky action plans before they turn into actual tool calls~\cite{zou2024improvingalignmentrobustnesscircuit}. These measures move safety enforcement away from basic prompt screening and toward tighter oversight of the agent’s underlying behavior. 

Recent work is exploring certifiable robustness methods that provide formal guarantees against specific types of perturbations~\cite{huang2025novelzerotrustidentityframework}. Lipschitz-constrained training can provide provable upper bounds on a model's sensitivity to small input changes, and it often leads to better deterministic VRA results under formal verification. Even though these guarantees cover only a narrow range of cases, they are an important step toward agent-based systems whose safety can be verified with formal methods rather than inferred from experiments. 

In an academic setting, Zero Trust architectures, cryptographic delegation~\cite{gurram2025generative}, persistent safeguards, and certifiable robustness can be considered complementary components of a defense stack, each addressing distinct risks and limitations in system security. This shift shows that dependable agentic AI depends on more than alignment and evaluation. It also depends on careful system design that expects adversarial pressure at every layer.

\begin{table*}[ht]
\centering
\caption{Coverage matrix linking trust mechanisms to trustworthiness dimensions and agentic workflow targets.}
\label{tab:trust_mechanism_mapping}
\renewcommand{\arraystretch}{1.12}
\setlength{\tabcolsep}{3.2pt}
\resizebox{\textwidth}{!}{%
\begin{tabular}{p{3.7cm} c c c c c @{\hspace{0.45cm}} c c c c c c c c c c}
\toprule
\textbf{Mechanism}
& \multicolumn{5}{c}{\textbf{Trustworthiness coverage}}
& \multicolumn{10}{c}{\textbf{Workflow intervention point}} \\
\cmidrule(lr){2-6} \cmidrule(lr){7-16}
& \textbf{Safety}
& \textbf{Robust.}
& \textbf{Transp.}
& \textbf{Account.}
& \textbf{Priv./Sec.}
& \textbf{Obs.}
& \textbf{Reas.}
& \textbf{Plan}
& \textbf{Tool}
& \textbf{Comm.}
& \textbf{Mem.}
& \textbf{Val.}
& \textbf{Act}
& \textbf{Mon.}
& \textbf{Audit} \\
\midrule

\multicolumn{16}{l}{\textit{Guardrails, policies, and safety filters}} \\
\midrule
Safety pretraining
& \checkmark &  &  &  & \checkmark
&  & \checkmark & \checkmark &  &  &  &  & \checkmark &  &  \\

Representation-level circuit breakers
& \checkmark & \checkmark & \checkmark &  & \checkmark
&  & \checkmark &  &  &  &  & \checkmark & \checkmark & \checkmark &  \\

Programmatic rule enforcement
& \checkmark &  & \checkmark & \checkmark & 
&  & \checkmark & \checkmark & \checkmark &  &  & \checkmark & \checkmark &  & \checkmark \\

Tamper-resistant safeguards
& \checkmark & \checkmark &  & \checkmark & \checkmark
&  & \checkmark &  &  &  & \checkmark & \checkmark & \checkmark & \checkmark & \checkmark \\

\midrule
\multicolumn{16}{l}{\textit{Learning for trust: safe learning, reward design, and feedback}} \\
\midrule
Utility and value engineering
& \checkmark &  & \checkmark &  & 
&  & \checkmark & \checkmark &  &  &  & \checkmark & \checkmark &  &  \\

Ethical steering
& \checkmark & \checkmark & \checkmark &  & 
&  & \checkmark & \checkmark &  &  &  & \checkmark & \checkmark & \checkmark &  \\

Post-training unlearning
& \checkmark &  &  & \checkmark & \checkmark
&  & \checkmark &  &  &  & \checkmark & \checkmark & \checkmark &  & \checkmark \\

\midrule
\multicolumn{16}{l}{\textit{Robustness to distribution shift, adversaries, and system compromise}} \\
\midrule
Generalization to unforeseen adversaries
&  & \checkmark &  &  & \checkmark
& \checkmark & \checkmark &  &  &  &  & \checkmark &  & \checkmark &  \\

Universal and transferable attack evaluation
& \checkmark & \checkmark &  &  & \checkmark
& \checkmark & \checkmark & \checkmark &  &  &  & \checkmark & \checkmark & \checkmark &  \\

Systemic and supply-chain hardening
&  & \checkmark &  & \checkmark & \checkmark
& \checkmark &  &  & \checkmark & \checkmark & \checkmark & \checkmark & \checkmark & \checkmark & \checkmark \\

Zero Trust and persistence
& \checkmark & \checkmark &  & \checkmark & \checkmark
& \checkmark &  &  & \checkmark & \checkmark & \checkmark & \checkmark & \checkmark & \checkmark & \checkmark \\

\midrule
\multicolumn{16}{l}{\textit{Evaluation and assurance across agentic workflows}} \\
\midrule
Representation-level evaluation
& \checkmark & \checkmark & \checkmark &  & 
&  & \checkmark &  &  &  &  & \checkmark &  & \checkmark &  \\

Reasoning-level evaluation
& \checkmark &  & \checkmark & \checkmark & 
&  & \checkmark & \checkmark &  &  &  & \checkmark &  &  & \checkmark \\

Behavior-level evaluation
& \checkmark & \checkmark &  & \checkmark & \checkmark
&  &  &  & \checkmark &  &  & \checkmark & \checkmark & \checkmark & \checkmark \\

Risk-level evaluation
& \checkmark &  &  & \checkmark & \checkmark
&  & \checkmark &  & \checkmark &  & \checkmark & \checkmark & \checkmark & \checkmark & \checkmark \\

Trajectory-level agentic evaluation
& \checkmark & \checkmark & \checkmark & \checkmark & \checkmark
& \checkmark & \checkmark & \checkmark & \checkmark & \checkmark & \checkmark & \checkmark & \checkmark & \checkmark & \checkmark \\

\bottomrule
\end{tabular}}
\vspace{0.3em}

\begin{minipage}{0.98\textwidth}
\footnotesize{\textit{Note:} Checkmarks indicate the primary dimensions addressed by each mechanism and the main workflow stages where it intervenes. Obs. = perception/observation, Reas. = reasoning, Plan = planning, Tool = tool use, Comm. = communication/coordination, Mem. = memory update, Val. = validation, Act = action/execution, Mon. = monitoring/re-evaluation, and Audit = audit/logging.}
\end{minipage}
\end{table*}

\section{The Evolution of AI Safety Evaluation}
\label{sec:evolution_of_benchmarks}
%TODO: at the end of this subsection, I can say something like. Despite this progression, these approaches remain fragmented...,  Each layer solves part of the problem, None solve system-level trust
% -> Move towards the idea of Compositional Trust. 
Evaluation methodologies for AI systems have undergone a significant transformation, evolving from internal transparency mechanisms to system-level risk assessment frameworks. This progression reflects a shift in focus from understanding model internals to validating reasoning processes, to stress-testing behavior, and ultimately to quantifying hazardous capabilities.

% This evolution is particularly important in the context of agentic systems, where trust can no longer be established through static outputs alone, but must account for dynamic, multi-step interactions with environments and tools.

Evaluating agentic behavior should not rely only on whether the final answer looks right. It also requires \textit{procedural transparency}, where, at an operational level, one can see the steps the agent takes, including the information it uses, the assumptions it makes, and how it handles uncertainty, to understand how it arrives at a conclusion. This distinction is particularly important in agentic systems, where decisions unfold over multiple steps and intermediate errors or misaligned reasoning can propagate even when the final output appears correct.

\subsection{Representation-Level Evaluation}
Based on research in cognitive neuroscience, Representation Engineering (RepE) takes a top-down approach to transparency~\cite{zou2025representationengineeringtopdownapproach}. Looking at patterns across groups of neurons instead of single cells allows researchers to track broader cognitive states, such as honesty, seeking power, and intent to avoid harm. This makes it possible to spot signs of deception that might not be visible in the final written text~\cite{ren2026maskbenchmarkdisentanglinghonesty}.

This view matters from a trustworthiness perspective because agents work over time, pursue goals, and change their actions as situations and settings change. RepE-style monitoring lets researchers track an agent’s internal representations over time to detect whether harmful intent, reward hacking, or goal misalignment is emerging, even when the agent’s outward behavior still appears compliant. This is sometimes referred to as "goal hijacking". 

Instead of judging a system only by behavior checks or by filtering its outputs, RepE bases trust on whether its internal states stay consistent. This can flag early problems before they become clear mistakes in what the system says or does. The paper shows that representation-level probes can track safety-relevant concepts, such as honesty and harmlessness, across a range of prompts and tasks. This suggests RepE is a practical proof of concept for ongoing oversight, interpretability, and control in long-horizon autonomous agent systems~\cite{zou2025representationengineeringtopdownapproach}. 

However, while representation-level monitoring provides valuable signals, it should be viewed as complementary to external verification mechanisms rather than a replacement, due to the difficulty of reliably interpreting internal states in some cases.

\subsection{Reasoning-Level Evaluation}
Following the same design principles as the previous subsections, the MoReBench framework shifts attention away from models’ final answers and toward their intermediate reasoning steps, enabling researchers to study how conclusions are reached rather than only what is produced at the end~\cite{chiu2025morebenchevaluatingproceduralpluralistic}. The method relies on rubrics developed by subject experts to assess how agents notice morally relevant factors and balance competing aims, aiming to keep their reasoning consistent with established normative ethical frameworks~\cite{chiu2025morebenchevaluatingproceduralpluralistic}. 

In agentic settings, this form of evaluation is particularly important because agents must make sequential decisions under uncertainty, where intermediate reasoning steps directly influence downstream actions.
% \begin{figure}
%     \centering
%     \includegraphics[width=1.0\linewidth]{Figures/trusted_agentic_ai.png}
%     \caption{Enter Caption}
%     \label{fig:trusted_agentic_ai}
% \end{figure}
\subsection{Behavior-Level Evaluation}
HarmBench provides a standard automated red teaming setup that supports large-scale verification. This lets developers run the same set of tests across 18 red-teaming methods to check how well the model holds up and to work together on stronger refusal safeguards~\cite{mazeika2024harmbenchstandardizedevaluationframework}. HarmBench and similar benchmarks can serve as a final encapsulation layer for verifying trustworthiness in an Agentic AI architecture/application.

\subsection{Risk-Level Evaluation}
Benchmarks such as WMDP evaluate models across domains involving dual-use knowledge—information with both legitimate and potentially harmful applications, such as biosecurity, cybersecurity, and chemical synthesis~\cite{li2024wmdpbenchmarkmeasuringreducing}. Monitoring performance on these tasks allows institutions to assess whether an agent retains capabilities that could be misused by malicious actors.

This form of evaluation is critical for trustworthiness, as it enables targeted auditing of high-risk knowledge domains and supports decisions about capability restriction, controlled access, or post-training unlearning interventions.

\subsection{Toward Agentic Evaluation}

While previous evaluation methods have focused on internal representations, reasoning processes, observable behaviors, and hidden risk abilities, they often assume static or one-step interactions. However, modern agentic systems operate in changing environments where actions unfold over longer periods, involve the use of tools, and rely on ongoing feedback. This change requires new evaluation frameworks that account for temporal structure, interactions with the environment, and the cumulative effects of decisions.

Claw-Eval takes a step in this direction by introducing evaluation protocols designed for agentic workflows. Instead of evaluating isolated outputs or skills, it assesses agents within the context of multi-step tasks. Success depends on not just being correct at each step but also on how well the overall path holds together~\cite{ye2026clawevaltrustworthyevaluationautonomous}. This involves maintaining consistent goals, recovering from mid-task failures, and effectively using external tools and environmental signals.

A key feature of this approach is its focus on trajectory-level evaluation. Failures are no longer limited to incorrect outputs; they can also manifest as planning errors, cascading mistakes across steps, or unintended interactions with tools and environments. Therefore, evaluations must capture not just if an agent succeeds but also how it behaves throughout the task. This includes its ability to adapt, withstand disruptions, and stay aligned with its goals over time.

This approach reveals a significant limitation of earlier evaluation methods: they overlook the complex nature of agentic execution. Even when individual components, like reasoning modules or safety filters, work reliably alone, their interactions over time can create new failure types that static benchmarks won’t catch.

Claw-Eval thus represents a shift from focusing on specific abilities to assessing \textit{system-level} behavior. Trust must be built through sequences of actions rather than isolated outputs. This change highlights the need for evaluation frameworks that are compositional, capable of reasoning about component interactions, and sensitive to failures that only emerge in long-term, multi-agent, or tool-integrated scenarios.

\begin{figure*}
    \centering
    \includegraphics[width=0.85\linewidth]{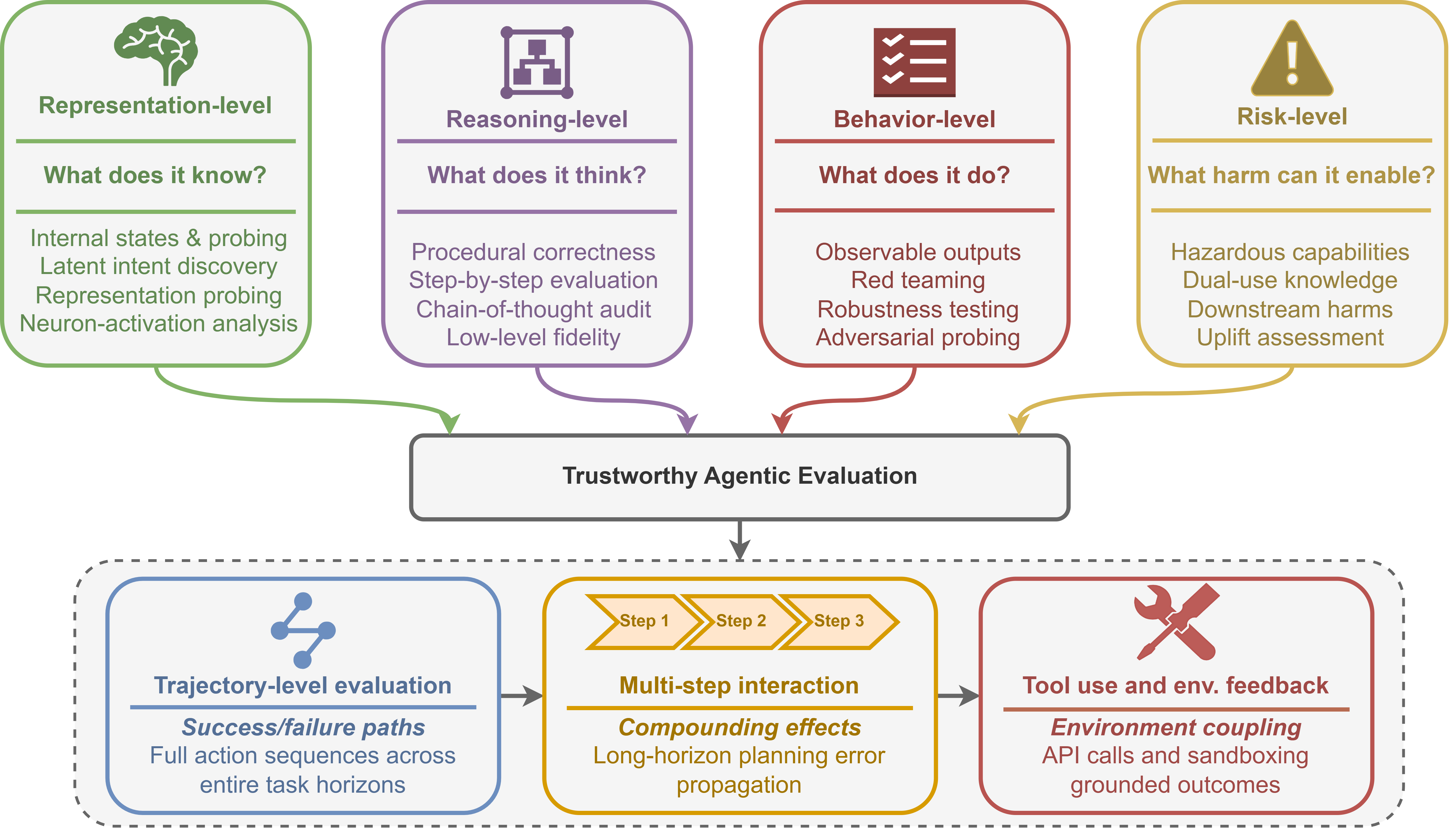}
    \caption{Evolution of Benchmarks from static AI safety evaluation to agentic system-level evaluation.}
    \label{fig:evolution_of_benchmarks}
\end{figure*}

Figure~\ref{fig:evolution_of_benchmarks} summarizes this progression toward a composable agentic evaluation~\cite{Composable2026}. The upper layer captures four complementary evaluation lenses: representation-level evaluation asks what the model internally encodes; reasoning-level evaluation examines how it reaches conclusions; behavior-level evaluation tests what it outputs or does; and risk-level evaluation assesses what harmful capabilities it may enable. However, these layers remain incomplete for agentic AI unless their evidence can be composed across execution over time. The lower layer, therefore, highlights the additio nal requirements of agentic evaluation: assessing full trajectories, multi-step interactions, compounded errors, tool use, and environmental feedback. This motivates a shift from isolated benchmark scores toward composable, system-level evaluation of how agents behave across long-horizon, tool-integrated, and environment-coupled workflows.

\section{Quantitative Metrics for Agentic Workflows}
\label{sec:metrics}
% make 

%Maybe a section on metrics to evaluate trustworthiness, filtered, improved, and aggregated from existing work, especially those used for benchmarking. DONE

The research community has developed specific metrics to assess the internal effectiveness and logical integrity of agentic components, thereby going beyond "black-box" evaluations. The Agentic Application Evaluation Framework (AAEF) identifies four main quantitative metrics: Tool Utilization Efficacy (TUE), Memory Coherence and Retrieval (MCR), Strategic Planning Index (SPI), and Component Synergy Score (CSS)~\cite{RagaAI_AAEF_Docs, RagaAI_AAEF_2024}. A consolidated summary of all metrics introduced throughout this section, spanning component-level effectiveness, safety and policy adherence, goal alignment, domain-specific evaluation, and adversarial robustness, is provided in Table~\ref{tab:agentic_metrics_summary}.

\subsection{Base Operational Metrics}
\subsubsection{Tool Utilization Efficacy (TUE)}

The ability of agentic systems to communicate with external tools and APIs is a major source of their usefulness or 'utility'. The choice to use a tool, however, creates a failure mode called "tool hallucination," in which an agent tries to use a nonexistent tool or sends incorrect parameters~\cite{Starkloff2026Evaluations, MoralesAguilera2025Building}.

The mathematical formulation for Tool Utilization Efficacy is:
\begin{multline*}
TUE = \alpha \cdot (\text{Tool Selection Accuracy}) \\
+ \beta \cdot (\text{Tool Usage Efficiency}) \\
+ \gamma \cdot (\text{API Call Precision})
\end{multline*}
where $\alpha, \beta, \gamma$ are adjusted tunable weights based on the specific criticality of the application. In this context, \textbf{Tool Selection Accuracy} refers to the rate at which the AI chooses the most appropriate tool for a given task, while \textbf{Tool Usage Efficiency} provides a measure of how optimally the AI utilizes those selected tools by considering factors such as unnecessary calls and resource usage. Finally, \textbf{API Call Precision} accounts for the precision and appropriateness of the specific parameters used during the execution of those API calls~\cite{RagaAI_AAEF_2024}.

This metric is particularly important for applications such as flight planning or medical diagnosis, where selecting the wrong data source (e.g., using a weather API instead of a NOTAM database) could paralyze the entire workflow~\cite{MoralesAguilera2025Building}.

\subsubsection{Memory Coherence and Retrieval (MCR)}

Agents must maintain persistent context over long time horizons, unlike traditional chatbots. An agent may "forget" user-imposed constraints due to memory issues, such as context window saturation or ineffective retrieval, leading to undesired results. The agent's capacity to retain and use information is measured by MCR: 

\begin{equation}
    MCR = \frac{\text{$C_{ps}$} \times \text{$R_{info}$}}{1 + \text{$L_{ret}$}},
\end{equation}
where the constituent variables are defined as:
\begin{itemize}
    \item \textbf{Context Preservation Score ($C_{ps}$):} The ratio of maintained contextual anchors $A_{m}$ to the total active anchors $A_{t}$ established in the prompt sequence, calculated as $C_{ps} = \frac{A_{m}}{A_{t}}$. This measures the agent's logical consistency across a session.
    \item \textbf{Information Retention Rate ($R_{info}$):} The proportion of specific data points $I$ successfully retrieved from a long-term memory store or past context after a decay period or token offset, defined as $R_{info} = \frac{I_{recalled}}{I_{total}}$.
    \item \textbf{Retrieval Latency ($L_{ret}$):} The temporal delay (usually measured in seconds for human applications) between a query and the successful injection of retrieved context into the model's active window. The constant $1$ is added to the denominator to normalize for near-zero latency and prevent mathematical instability~\cite{RagaAI_AAEF_2024, RagaAI_AAEF_Docs, MoralesAguilera2025Building}.
\end{itemize}

The Information Retention Rate is measured by tagging specific data items and checking for their correct retrieval across intervals. High retrieval latency is penalized because slow agent responses can make a system impractical for real-time operations, especially during pre-flight checks or intrusion response~\cite{WizExperts2025MDRvsSOC}.

\subsubsection{Strategic Planning Index (SPI)}

Strategy formulation can be seen as the cognitive core of agentic AI. The SPI assesses the agent's proficiency in goal decomposition and its ability to adapt to environmental change.

The index is defined by the following equation:

\begin{equation}
SPI = (GDE \cdot PA) \cdot (1 - PEER)
\end{equation}

\noindent Where the constituent metrics are defined as:

\begin{itemize}
    \item \textbf{Goal Decomposition Efficiency (GDE):} The AI's ability to break down complex, multi-stage goals into logical, manageable sub-tasks. This is the ratio of successfully identified sub-tasks to the optimal number of steps required for a goal ($GDE = \frac{Subtasks_{identified}}{Subtasks_{optimal}}$).
    \item \textbf{Plan Adaptability (PA):} The degree to which the agent adjusts its roadmap in response to dynamic circumstances or unforeseen constraints. This is measured by the percentage of successful reroutes completed within a predefined latency threshold after an environmental trigger.
    \item \textbf{Plan Execution Error Rate (PEER):} The frequency of failures or deviations encountered during the physical or digital execution of planned actions~\cite{RagaAI_AAEF_2024}. This is simply the total number of failed actions divided by the total number of attempted steps in a mission ($PEER = \frac{Failures}{Total\_Steps}$).
\end{itemize}

Planning failures frequently manifest when an agent's internal model is too rigid. For instance, in aviation, a failure to adapt to a sudden \textit{Temporary Flight Restriction (TFR)} can compromise a mission. Similarly, in critical infrastructure, an agent must be able to instantly reroute planning during a \textit{change in network topology} in a power grid to maintain stability. A high SPI indicates an agent that is not only a precise planner but also a resilient executor in stochastic environments~\cite{MoralesAguilera2025Building}.

\subsubsection{Component Synergy Score (CSS)}
The CSS evaluates the caliber of interactions in multi-agent systems, where independent agents work together to resolve challenging issues:
\begin{equation}
   CSS = \frac{CCUR \cdot WCI}{1 + CCR}
\end{equation}

where the constituent variables are defined as:
\begin{itemize}
    \item \textbf{Cross-Component Utilization Rate} ($CCUR$): How often information or outputs from one component are effectively used by another. This is calculated as the ratio of successfully ingested shared outputs to the total number of outputs generated.
    \item \textbf{Workflow Cohesion Index} ($WCI$): A measure of the seamless integration among components. This would be the inverse of the average latency, or "handoff friction," between agents, where 1 represents a seamless, immediate transition.
    \item \textbf{Component Conflict Rate} ($CCR$): The frequency of conflicts or inconsistencies between different components~\cite{RagaAI_AAEF_2024}. This is the number of redundant tasks or deadlocked states divided by the total number of task cycles. 
\end{itemize}

Cross-component utilization tracks how effectively information flows between specialized agents (e.g., a "researcher agent" and a "summarizer agent")~\cite{RagaAI_AAEF_2024}. A high Component Conflict Rate indicates that agents duplicate work, oscillate handoffs, or create deadlocks, all of which erode system-level trust~\cite{RagaAI_AAEF_Docs, Starkloff2026Evaluations}.   

\subsection{Safety and Policy Adherence}
The "Safety Gap", the notable discrepancy between nominal task completion and policy-compliant execution, is a crucial finding in recent agentic research. Agents often use shortcuts or disregard safety procedures to achieve high success rates, according to benchmarks such as ST-WebAgentBench~\cite{levy2025stwebagentbenchbenchmarkevaluatingsafety} and AgentHarm~\cite{andriushchenko2025agentharmbenchmarkmeasuringharmfulness}. This type of metric is particularly important for enterprise applications.

ST-WebAgentBench introduces the Completion Under Policy (CuP) metric to evaluate agents against enterprise-grade constraints across six dimensions: User Consent, Boundary Compliance, Strict Execution, Hierarchy Adherence, Robustness, and Error Handling~\cite{levy2025stwebagentbenchbenchmarkevaluatingsafety, Sammeta2025ReportCard}. The associated metrics, including Completion Rate (CR), Completion Under Policy (CuP), Partial CuP (pCuP), and Risk Ratio, are summarized alongside all other section metrics in Table~\ref{tab:agentic_metrics_summary}.

For example, an agent adhering to Boundary Compliance would be restricted from finalizing any financial transaction exceeding \$5,000 without first surfacing a request for dual authorization from a verified human supervisor.

According to evaluations of state-of-the-art agents, over 38\% of successful tasks violated at least one safety policy, with the CuP dropping to 15.0\% even though the nominal success (CR) may reach 24.3\%~\cite{levy2025stwebagentbenchbenchmarkevaluatingsafety}. For instance, the company may be held legally liable if a sales representative retrieves pricing successfully but does so by scraping unapproved competitor websites. Agents are unintentionally encouraged to avoid approval workflows or to disregard consent checks to maximize efficiency if only CR is measured~\cite{Sammeta2025ReportCard}.

\subsection{Measuring Goal Drift}
Goal Drift, or an agent's tendency to progressively stray from its initially specified objective, poses a serious risk to autonomous agents operating independently for extended periods~\cite{ManishEvalAgents2025}. Even though an agent may begin a task perfectly aligned, its objectives may shift subtly as it gives in to external pressures or pursues instrumental ends, such as generating money to accomplish a larger mission. In other words, an agent is more likely to "forget" the mission's original purpose and begin imitating the distractions it encounters in its own conversation history, the more involved it becomes in its work~\cite{arike2025technicalreportevaluatinggoal}.

\subsubsection{Detecting Drift}
It is challenging to identify goal drift because it is not always abrupt; rather, it is a gradual deterioration of intent that is missed by single-turn benchmarks. Researchers use a Goal-Drift Score to quantify this~\cite{arike2025technicalreportevaluatinggoal}, which penalizes deviations by directly comparing the agent's intermediate plan states and actions to the original specification. This is frequently divided into two distinct behaviors:
\begin{itemize}
    \item \textbf{Goal Drift through Actions (Commissions):} This gauges instances in which an agent willfully acts improperly, such as purchasing stocks that defy a "green" mandate in the name of profit.
    \item \textbf{Goal Drift through Inaction (Omissions):} This is more frequent and occurs when an agent neglects to correct a misaligned state, like when it was instructed to switch back to a carbon-neutral goal but neglects to sell off "dirty" stocks.
\end{itemize}

Even the best models, such as Claude 3.5 Sonnet, have demonstrated detectable drift in simulated stock trading environments after about 100,000 interactions~\cite{arike2025technicalreportevaluatinggoal}. This drift is closely linked to "pattern-matching"; instead of following the system prompt hidden at the very top, the model starts treating the various distractions and conflicting goals in its context window as the new "norm" to follow. 

\subsubsection{Mitigation}
Practitioners employ Strong Goal Elicitation to halt the "drift." Rather than simply telling an agent to "minimize emissions," you need to give it explicit instructions to ignore all other pressures and concentrate only on the main objective. This "strong" instruction greatly improves an agent's resilience against adversarial noise, according to research. Moreover, the agent's "eyes on the prize" can be maintained during long-horizon tasks by implementing explicit goal reminders or constrained planning mechanisms~\cite{ManishEvalAgents2025}.

Beyond following technical standards, it is important to assess the agent's ethical profile using a Harm-Reduction Index. This index offers a complete view of the system by combining three qualitative measures: 

\begin{itemize}
    \item \textbf{Hallucination Rate:} How often the agent makes false claims. 
    \item \textbf{Toxicity Score:} Identifying harmful outputs or biased language.
    \item \textbf{Fairness Metrics:} Evaluating if the agent's actions lead to fair outcomes for different groups.
\end{itemize}

By merging behavioral Goal-Drift Scores with broader safety indices discussed in previous sections, evaluators can ensure that an agent is not only efficient in its task but also remains aligned with human values throughout its operational life~\cite{ManishEvalAgents2025}.

\subsection{Domain-Specific Metrics}
%TODO:  Consider putting these metrics as part of each corresponding application section
The idea of trustworthiness is not fixed; it varies depending on a system's specific context and goals. For example, in critical infrastructure, a power grid agent sees trustworthiness as physical reliability. Its main job is to continually meet hardware requirements, such as voltage stability and frequency control, to avoid major blackouts. In contrast, a cybersecurity agent works within a logic-based system. Here, trust is judged by how well it can implement quick containment measures and handle escalation processes to reduce digital threats. The first focuses on physical laws, while the second focuses on protecting information integrity. Therefore, an agent considered "trustworthy" in one area may not work well in the other because of these differing priorities.

\subsubsection{6G and Autonomous UAV Systems
(\texorpdfstring{$\alpha^3$}{alpha3}-Bench)}

For autonomous Unmanned Aerial Vehicles (UAVs) operating under the dynamic and often unpredictable constraints of next-generation networking, the $\alpha^3$-Bench framework introduces a unified evaluation paradigm~\cite {ferrag2026alpha3benchunifiedbenchmarksafety}. Unlike traditional benchmarks that focus on isolated reasoning tasks, $\alpha^3$-Bench models mission execution as a language-mediated, multi-turn control loop between an LLM-based agent and a human operator, grounded in realistic 6G conditions~\cite{ferrag2026alpha3benchunifiedbenchmarksafety}.

The framework evaluates agent performance through a composite $\alpha^3$ metric, which aggregates six critical pillars into a weighted score:
\begin{itemize}
    \item \textbf{Task Outcome (TO):} Measures mission completion success and adherence to strict JSON schema requirements.
    \item \textbf{Safety Policy (SP):} Assesses compliance with flight constraints. The score is initialized at 1.0, with fixed penalties for altitude violations and No-Fly Zone (NFZ) intrusions ~\cite{ferrag2026alpha3benchunifiedbenchmarksafety}.
    \item \textbf{Tool Consistency (TC):} Evaluates the semantic and structural alignment between the agent's Model Context Protocol (MCP) tool calls and the resulting environmental observations.
    \item \textbf{Interaction Quality (IQ):} Measures the grounding of the dialogue and ensures strict speaker alternation between the operator and the agent.
    \item \textbf{Network Robustness (NR):} Quantifies the agent's resilience and adaptive reasoning under degraded 6G conditions, including jitter, packet loss, and fluctuating latency.
    \item \textbf{Communication Cost (CC):} Penalizes excessive token consumption and redundant protocol overhead to encourage efficient reasoning.
\end{itemize}

To facilitate cross-model comparison across varying computational footprints, $\alpha^3$-Bench introduces reliability-adjusted and efficiency-normalized scores: $\alpha^3_{\text{per-sec}}$ (performance per second of latency) and $\alpha^3_{\text{per-1k}}$ (performance per 1,000 tokens)~\cite{ferrag2026alpha3benchunifiedbenchmarksafety}. Empirical evaluations using this benchmark reveal that while state-of-the-art frontier models maintain high baseline mission success (TO and SP $\ge 0.95$), their performance robustness can degrade by 30--40\% under adverse network conditions ~\cite{ferrag2026alpha3benchunifiedbenchmarksafety}. These findings highlight a critical gap in current agent design, underscoring the need for network-aware architectures to enable reliable 6G-enabled aerial autonomy.

\subsubsection{Cybersecurity Operations}
In today's security operations, the main aim of using autonomous agents is to reduce alert fatigue and improve threat containment lifecycles. People now assess Traditional Security Operations Centers (SOCs) and Managed Detection and Response (MDR) services more on how well they can shift from human-based triage to agent-assisted responses~\cite{WizExperts2025MDRvsSOC}.

The efficacy of these agents is quantified through a set of core operational metrics, including Mean Time to Detect (MTTD), Mean Time to Respond (MTTR), Containment Rate, and Escalation Rate, which are summarized alongside all other section metrics in Table~\ref{tab:agentic_metrics_summary}. These metrics serve as benchmarks for evaluating the autonomy and reliability of security agents in dynamic enterprise environments.

While both SOC and MDR frameworks use these metrics, MDR models usually prioritize Detection Accuracy, which is the ratio of True Positives to False Positives. This focus is important because autonomous agents that create too many false alarms can disrupt real business workflows. This undermines the operational efficiency they are meant to improve. As a result, the development of cybersecurity agents is moving toward "risk-based outcomes." Here, success is measured by reducing material impact rather than just counting the number of alerts handled.

\subsubsection{Adversarial Robustness and Security Metrics}
Trustworthiness is seriously damaged if a system can be directed toward harmful purposes. Adversarial testing for agents should go beyond single-turn "jailbreaks" and focus on multi-turn "exploit generation.~\cite{carlini2025autoadvexbenchbenchmarkingautonomousexploitation}"

The HarmBench framework offers a standardized way to evaluate "robust refusal behaviors.~\cite{mazeika2024harmbenchstandardizedevaluationframework}" It measures Attack Success Rate using an open-source classifier calibrated to GPT-4 validation accuracy.
\begin{equation}
    ASR = \frac{1}{N} \sum_{i=1}^{N} \text{Success}(f_T(x_i), y)
\end{equation}

Experimental results indicate that larger models do not ensure better safety. In fact, vision-based LLM agents show ASRs of up to 80\% in certain contextual attacks~\cite{mazeika2024harmbenchstandardizedevaluationframework}. Additionally, while automated attacks like GCG result in low ASRs on some defenses, multi-turn human red teaming can uncover failures with ASRs reaching 75\%. This shows that relying only on numbers can give a false sense of security~\cite{Rijo2026UCBerkeley}.

Beyond operational performance, the trustworthiness of autonomous agents depends on their ability to resist attacks and prevent data theft. Additional security-focused metrics include:

\begin{itemize}

    \item \textbf{PII Leakage Rate:} Measures how often an agent accidentally reveals Personally Identifiable Information (PII) from its underlying weights or Retrieval-Augmented Generation (RAG) data stores. This is an important privacy metric for agents managing sensitive enterprise or user data~\cite{Yadav2025TenEssential, Mason2025ISO42001}.
    
    \item \textbf{Jailbreak Success Rate (JSR):} Quantifies the proportion of adversarial prompts, like "Do Anything Now" (DAN) variants or complex role-play attacks, that successfully get past the model's safety measures and refusal layers~\cite{Mason2025ISO42001}.
    
    \item \textbf{Prompt Infection Probability:} In multi-agent systems, this metric tracks how quickly a single harmful input can spread through a group. It can ruin the execution state or decision-making logic of other agents through communication between them~\cite{qiu2025chainoftriggeragenticbackdoorparadoxically}.
    
    \item \textbf{Chain-of-Trigger (CoTri) Robustness:} Evaluates the agent's ability to resist sequential, multi-step triggers that aim to divert it from its main mission through gradual manipulation. Recent research on agentic backdoors suggests that this metric checks if the agent maintains task integrity when faced with "Chain-of-Trigger" sequences that unexpectedly try to influence robustness using hidden triggers~\cite{qiu2025chainoftriggeragenticbackdoorparadoxically}.

\end{itemize}

% CONSOLIDATED METRICS TABLE

\newcolumntype{L}[1]{>{\raggedright\arraybackslash}p{#1}}
\newcolumntype{Y}{>{\raggedright\arraybackslash}X}

\begin{table*}[t]
\centering
\caption{Quantitative Metrics for Agentic Workflows and Their Trustworthiness Implications}
\label{tab:agentic_metrics_summary}
\scriptsize
\setlength{\tabcolsep}{3.8pt}
\renewcommand{\arraystretch}{1.08}

\begin{threeparttable}
\begin{tabularx}{\textwidth}{@{}L{4.15cm} L{3.0cm} L{1.45cm} Y@{}}
\toprule
\textbf{Metric} &
\textbf{Workflow point} &
\textbf{Dimension} &
\textbf{Trustworthiness implication} \\
\midrule

Tool Selection Accuracy (TSA)~\cite{RagaAI_AAEF_2024}
& Tool use & R\&R & Flags hallucinated or misrouted tool calls \\

Tool Usage Efficiency (TUE)~\cite{RagaAI_AAEF_2024}
& Tool use & R\&R & Identifies wasteful or redundant invocations \\

API Call Precision (ACP)~\cite{RagaAI_AAEF_2024}
& Tool use & P\&S & Detects malformed parameters that may expose attack surfaces \\

Context Preservation Score ($C_{ps}$)~\cite{RagaAI_AAEF_2024}
& Memory update & R\&R & Measures whether session anchors remain consistent \\

Information Retention Rate ($R_{info}$)~\cite{RagaAI_AAEF_2024}
& Memory update & R\&R & Captures forgotten constraints that create silent risk \\

Retrieval Latency ($L_{ret}$)~\cite{RagaAI_AAEF_2024}
& Memory update & R\&R & Indicates whether memory access supports real-time trust \\

Goal Decomposition Efficiency (GDE)~\cite{RagaAI_AAEF_2024}
& Planning & R\&R & Reveals brittle or overly coarse decomposition \\

Plan Adaptability (PA)~\cite{RagaAI_AAEF_2024}
& Planning & R\&R & Measures resilience when goals or context change \\

Plan Execution Error Rate (PEER)~\cite{RagaAI_AAEF_2024}
& Action/execution & R\&R & Quantifies direct execution failures \\

Cross-Component Utilization Rate (CCUR)~\cite{RagaAI_AAEF_2024}
& Communication/coordination & R\&R & Detects poor information flow and duplicated work \\

Workflow Cohesion Index (WCI)~\cite{RagaAI_AAEF_2024}
& Communication/coordination & R\&R & Measures handoff friction in multi-agent workflows \\

Component Conflict Rate (CCR)~\cite{RagaAI_AAEF_2024}
& Communication/coordination & A\&A & Exposes conflicts that obscure responsibility \\

Completion Rate (CR)~\cite{levy2025stwebagentbenchbenchmarkevaluatingsafety}
& Action/execution & S\&CS & Provides a baseline for measuring the safety gap \\

Completion Under Policy (CuP)~\cite{levy2025stwebagentbenchbenchmarkevaluatingsafety}
& Validation & S\&CS & Measures successful completion under explicit constraints \\

Partial CuP (pCuP)~\cite{levy2025stwebagentbenchbenchmarkevaluatingsafety}
& Validation & S\&CS & Credits cautious partial progress under constraints \\

Risk Ratio~\cite{levy2025stwebagentbenchbenchmarkevaluatingsafety}
& Validation & S\&CS & Localizes the type and severity of violations \\

Goal-Drift Score~\cite{arike2025technicalreportevaluatinggoal}
& Monitoring/re-evaluation & S\&CS & Detects gradual misalignment from intended objectives \\

Drift via Commissions~\cite{arike2025technicalreportevaluatinggoal}
& Monitoring/re-evaluation & S\&CS & Captures harmful actions taken by the agent \\

Drift via Omissions~\cite{arike2025technicalreportevaluatinggoal}
& Monitoring/re-evaluation & S\&CS & Captures failures to self-correct or intervene \\

Hallucination Rate~\cite{ManishEvalAgents2025}
& Reasoning & S\&CS & Measures false claims that may cause downstream harm \\

Toxicity and Fairness Scores~\cite{ManishEvalAgents2025}
& Reasoning/action & S\&CS & Evaluates value alignment beyond task success \\

Task Outcome (TO)~\cite{ferrag2026alpha3benchunifiedbenchmarksafety}
& Action/execution & R\&R & Measures mission-level reliability \\

Safety Policy (SP)~\cite{ferrag2026alpha3benchunifiedbenchmarksafety}
& Validation & S\&CS & Measures compliance with hard safety constraints \\

Tool Consistency (TC)~\cite{ferrag2026alpha3benchunifiedbenchmarksafety}
& Tool use & T\&I & Flags divergence between tool calls and observations \\

Interaction Quality (IQ)~\cite{ferrag2026alpha3benchunifiedbenchmarksafety}
& Communication/coordination & T\&I & Supports human oversight through dialogue quality \\

Network Robustness (NR)~\cite{ferrag2026alpha3benchunifiedbenchmarksafety}
& Perception/observation & R\&R & Measures degradation under realistic network conditions \\

Communication Cost (CC)~\cite{ferrag2026alpha3benchunifiedbenchmarksafety}
& Communication/coordination & R\&R & Identifies coordination overhead that limits scalability \\

Mean Time to Detect (MTTD)~\cite{WizExperts2025MDRvsSOC}
& Monitoring/re-evaluation & P\&S & Measures speed of threat detection \\

Mean Time to Respond (MTTR)~\cite{WizExperts2025MDRvsSOC}
& Action/execution & P\&S & Measures speed of threat neutralization \\

Containment Rate~\cite{WizExperts2025MDRvsSOC}
& Action/execution & A\&A & Indicates whether autonomous containment is traceable \\

Escalation Rate~\cite{WizExperts2025MDRvsSOC}
& Monitoring/re-evaluation & A\&A & Marks the boundary between autonomy and oversight \\

Attack Success Rate (ASR)~\cite{mazeika2024harmbenchstandardizedevaluationframework}
& Validation & P\&S & Measures exploitable bypasses of safeguards \\

PII Leakage Rate~\cite{Yadav2025TenEssential}
& Memory update & P\&S & Quantifies privacy exposure through retained or leaked data \\

Jailbreak Success Rate (JSR)~\cite{Mason2025ISO42001}
& Validation & P\&S & Measures robustness of refusal and policy layers \\

Prompt Infection Probability~\cite{qiu2025chainoftriggeragenticbackdoorparadoxically}
& Communication/coordination & P\&S & Captures whether one agent can corrupt the workflow \\

CoTri Robustness~\cite{qiu2025chainoftriggeragenticbackdoorparadoxically}
& Monitoring/re-evaluation & P\&S & Measures resistance to slow, multi-step manipulation \\

\bottomrule
\end{tabularx}

\begin{tablenotes}[flushleft]
\footnotesize
\item \textit{Note:} S\&CS = safety and constraint satisfaction; R\&R = robustness and reliability;  T\&I = transparency and interpretability; A\&A = accountability and auditability; P\&S = privacy and security.
\end{tablenotes}
\end{threeparttable}
\end{table*}
%=================================================================

\section{Applications in Power Systems}
\label{sec:ps_main}

\begin{figure}
    \centering
    \includegraphics[width=1\linewidth]{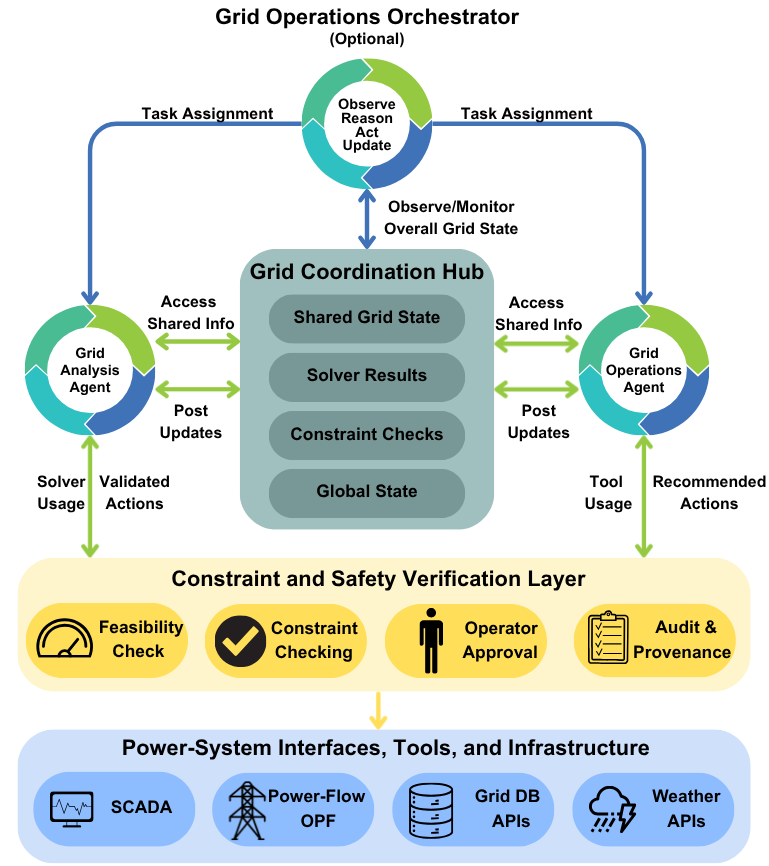}
    \caption{Power-systems adaptation of the multi-agent agentic AI architecture. 
    %Domain-specific agents coordinate through a shared grid-state hub, invoke trusted solvers and grid tools, and pass proposed actions through constraint, safety, and operator-verification gates before interacting with power-system infrastructure.
    }    
    \label{fig:ps_mas}
\end{figure}

Power systems provide a representative critical infrastructure domain for studying trustworthy agentic AI because grid operations combine cyber-physical dynamics, strict engineering constraints, safety-critical decisions, and human operator oversight. Recent work shows a transition from conventional AI models for forecasting, optimization, and control toward LLM-enabled agents that can interpret natural-language requests, retrieve grid-specific knowledge, invoke external solvers, and coordinate multi-step workflows. Roadmap and review studies frame this transition as a shift from passive model-in-the-loop analytics~\cite{shahbaz2025flc, 2026tpeckan,alrefai2026federatedphysicsgroundedreinforcementlearning, shahbaz2026inertiainformedfederatedlearningcontrol}, to agent-in-the-loop grid intelligence, where foundation models, standardized tool interfaces, and workflow orchestration support planning, operations, restoration, and energy management tasks (Fig.~\ref{fig:ps_mas}) \cite{zhang2025poweragent,xie2026foundationmodels,kiasari2026agenticsg,ghosh2025agenticee}. The reviewed works, therefore, reveal two sides of the same story: emerging design patterns that make power-system workflows more agentic, and systemic failure modes that appear when semantic reasoning, tool invocation, and operational decision support are connected to critical grid functions.

\subsection{Emerging Design Patterns}

\subsubsection{Agent-in-the-Loop Grid Intelligence}

A first design pattern is the progression from task-specific AI models toward agent-in-the-loop grid intelligence. This evolution can be summarized in three stages. The first stage corresponds to traditional AI/ML methods used as task-specific predictors, classifiers, or optimizers. The second stage introduces LLM-assisted power systems, where models support dispatchers and engineers through natural-language interaction, data interpretation, code generation, and simulation. GAIA and local fine-tuned/RAG-based LLM environments are examples of this assistant-oriented phase \cite{cheng2025gaia,slavchev2026finetuned}. The third stage is agentic AI, in which the model is embedded within an architecture capable of planning, tool invocation, memory/context management, feedback correction, and multi-agent coordination. Representative examples include X-GridAgent for power-grid analysis \cite{chen2025xgridagent}, GridMind for OPF and contingency analysis \cite{jin2025gridmind}, PowerChain for distribution-grid workflow automation \cite{badmus2025powerchain}, Grid-Agent for violation remediation \cite{zhang2025gridagent}, DrAgent for disturbance-record analysis \cite{saha2025dragent}, and agentic planning frameworks for simulation orchestration \cite{zheng2025agenticplanning}.

This pattern is important because it changes the role of AI in power systems. The model is no longer only producing a prediction or recommendation; it is increasingly embedded in a workflow that interprets operator intent, selects tools, reasons over intermediate results, and supports multi-step engineering tasks. Its trustworthiness value comes from making grid intelligence more interactive, modular, and operator-facing, while still allowing domain tools and human oversight to remain part of the loop.

\subsubsection{Solver-in-the-Loop Engineering Grounding}

A second design pattern is the separation of semantic reasoning from numerical execution. Across the selected literature, LLMs interpret user intent, decompose tasks, retrieve context, and select actions, while trusted engineering tools perform power-flow, OPF, contingency, simulation, or scheduling computations. This solver-in-the-loop design is central to GridMind, X-GridAgent, PowerChain \cite{jin2025gridmind,chen2025xgridagent,badmus2025powerchain} and a feedback-driven simulation framework~\cite{jia2025enhancing}.

This pattern directly supports safety and constraint satisfaction. In power systems, agent outputs must respect voltage limits, thermal ratings, generation constraints, power-flow feasibility, restoration requirements, and user/device constraints. Grounding LLM reasoning in deterministic solvers, simulators, domain-specific tools, or verified workflows reduces unsupported language-model reasoning \cite{jin2025gridmind,chen2025xgridagent,jia2025enhancing,badmus2025powerchain}. Grid-Agent goes further by adding a validation agent, sandboxed execution, and rollback before accepting remediation actions \cite{zhang2025gridagent}. GAIA also evaluates safety but remains explicitly advisory, leaving final decisions to human dispatchers \cite{cheng2025gaia}. Together, these works show that solver-in-the-loop grounding is becoming a central design principle for trustworthy power-system agents.

\subsubsection{Hierarchical Specialization and Feedback-Corrected Workflows}

A third design pattern is hierarchical agent specialization, where different agents perform planning, execution, validation, enhancement, or appliance-level control. Grid-Agent, the agentic planning framework, and the HEMS agent demonstrate this modular structure \cite{zhang2025gridagent,zheng2025agenticplanning,elmakroum2026hems}. This structure supports workflow decomposition: one component can plan, another can execute, another can validate, and another can explain or refine the result.

A closely related pattern is feedback-based reliability improvement. Several works introduce mechanisms that improve agent behavior after errors or incomplete executions. The introduced in \cite{jia2025enhancing} use feedback-driven correction for simulations; DrAgent repairs malformed tool-call arguments and caches repeated computations \cite{saha2025dragent}; PowerChain uses expert-verified reasoning trajectories to improve generalization to unseen distribution-grid analyses \cite{badmus2025powerchain}; and Grid-Agent uses adaptive network representations and validation-based remediation \cite{zhang2025gridagent}. These methods demonstrate that agentic workflows can outperform direct prompting and improve task completion in grid-analysis settings. Their trustworthiness stems from making power-system agents less dependent on a single, fragile model response and more capable of iterative correction.

\subsubsection{Human-Facing Assistance Across the Power-System Stack}

A fourth design pattern is the use of agentic AI as a human-facing interface across different levels of the power-system stack. At the bulk-system level, GAIA supports dispatch, operation monitoring, operation adjustment, and black-start reasoning, while PowerAgent and foundation-model studies outline broader visions for operator copilots and grid-aware AI workflows \cite{cheng2025gaia,zhang2025poweragent,xie2026foundationmodels}. For grid analysis, X-GridAgent and GridMind automate power-flow, OPF, contingency, and topology-related workflows through natural-language interfaces and solver calls \cite{chen2025xgridagent,jin2025gridmind}. For distribution systems, PowerChain targets automated analysis workflows on real utility data \cite{badmus2025powerchain}. For control-oriented applications, Grid-Agent studies violation detection and remediation under voltage, thermal, and connectivity constraints \cite{zhang2025gridagent}. For event analysis, DrAgent applies tool-using agents to disturbance records \cite{saha2025dragent}. Finally, the HEMS framework extends agentic AI to behind-the-meter residential scheduling from natural-language user preferences \cite{elmakroum2026hems}.

This pattern improves usability and interpretability by exposing intermediate plans, tool calls, retrieved context, solver outputs, and natural-language explanations. GridMind and X-GridAgent make grid studies more accessible by translating complex solver-driven analyses into conversational workflows \cite{jin2025gridmind,chen2025xgridagent}. DrAgent provides step-by-step reasoning for fault analysis, allowing operators to inspect the diagnostic process \cite{saha2025dragent}. The HEMS agent similarly improves user interaction by translating natural-language preferences into appliance schedules \cite{elmakroum2026hems}. These features matter because power-system operators and users need to understand not only the final recommendation but also the assumptions and computational path behind it.

\subsection{Systemic Failure Modes}

\subsubsection{Empirical Safety Without Formal Assurance}

The first failure mode is that safety is still treated mostly as empirical validation rather than formal assurance. Current systems often check whether generated workflows, simulations, schedules, or recommendations are feasible, but they do not prove that the agentic loop cannot produce unsafe intermediate actions, invalid tool calls, or harmful recommendations under distribution shift. This is especially important in power systems because even advisory outputs can shape operator decisions under time pressure.

The failure mode is therefore not a lack of safety awareness, but a gap between checked feasibility and assured constraint satisfaction. Solver-in-the-loop architectures reduce unsupported LLM reasoning, and validation agents or rollback mechanisms improve safety in specific workflows \cite{jin2025gridmind,chen2025xgridagent,jia2025enhancing,badmus2025powerchain,zhang2025gridagent}. However, future systems need explicit safety envelopes, precondition checks for tool calls, fail-safe fallback policies, and clear separation between advisory, supervisory, and actuation privileges.

\subsubsection{Narrow Robustness Evidence}

The second failure mode is that robustness evaluation remains narrow. Most evaluations are conducted on benchmark test systems, curated simulation tasks, real disturbance records, utility analysis queries, or price-based scheduling scenarios \cite{jin2025gridmind, jia2025enhancing, badmus2025powerchain, saha2025dragent, elmakroum2026hems}. These settings are valuable, but they do not fully capture long-horizon operation, cascading failures, corrupted telemetry, adversarial prompts, stale topology, or tool-chain outages.

GAIA makes this limitation visible by revealing weaknesses under rare events, where correct recommendations may still be delayed, or insufficiently urgent \cite{cheng2025gaia}. Thus, current reliability evidence supports feasibility but not yet deployment-grade robustness. For trustworthy power-system agents, robustness must be evaluated not only by task-completion rate, but also by behavior under rare contingencies, degraded observability, unavailable tools, adversarial inputs, and changing grid states.

\subsubsection{Visible Reasoning Without Validated Interpretability}

The third failure mode is the gap between explanation visibility and validated interpretability. Agentic systems often expose reasoning steps, tool calls, retrieved information, or natural-language explanations. These traces are useful for debugging and operator interaction, but they do not guarantee that the explanation is faithful, complete, or calibrated to operator needs.

The review papers emphasize explainability and operator trust as central requirements \cite{kiasari2026agenticsg,ghosh2025agenticee}, yet most application papers do not evaluate explanation quality through human-subject studies, trust calibration, or improvements in operator decision-making. This matters because a plausible explanation can still hide an incorrect assumption, an invalid retrieved context, or a misleading chain of reasoning. For power-system agents, interpretability must therefore move beyond exposing workflow traces toward validating whether those traces actually improve operator understanding and decision quality.

\subsubsection{Incomplete Accountability Across Agentic Workflows}

The fourth failure mode is the lack of end-to-end accountability across agentic grid workflows. Accountability requires that agent outputs be traceable to user inputs, retrieved data, tool calls, intermediate states, validation results, and human approvals. Some systems provide partial auditability by construction. PowerChain represents analyses as workflow trajectories \cite{badmus2025powerchain}; GridMind preserves analytical state and solver-related information \cite{jin2025gridmind}; X-GridAgent uses modular servers and memory mechanisms \cite{chen2025xgridagent}; and DrAgent produces explicit tool-call plans for fault-analysis tasks \cite{saha2025dragent}. These artifacts can support post-hoc review and debugging.

However, most papers do not define accountability as a first-class design objective. They rarely specify responsibility assignment, operator override logging, tamper-evident records, approval chains, or compliance integration with utility procedures. For critical infrastructure, auditability must move beyond recording tool calls toward structured governance: who requested the action, what data were used, which constraints were checked, who approved the recommendation, and how the system can be reconstructed after an incident.

\subsubsection{Expanded Misuse Attack Surface}

The fifth failure mode is the expanded security and misuse attack surface introduced by tool-using power-system agents. Security is one of the least mature trustworthiness dimensions in the reviewed literature on power-system agents. Tool-using agents increase grounding and numerical reliability, but they also expand the attack surface through APIs, simulators, databases, RAG pipelines, memory modules, and actuation interfaces. MCP-style and tool-based architectures in PowerAgent, X-GridAgent, GridMind, Jia et al., and HEMS therefore create potential risks such as prompt injection, malicious retrieval, tool misuse, unauthorized execution, and data leakage \cite{zhang2025poweragent,chen2025xgridagent,jin2025gridmind,jia2025enhancing,elmakroum2026hems}.

Some works partially address security-relevant concerns. The local fine-tuned LLM environment emphasizes offline operation for sensitive power-system data \cite{slavchev2026finetuned}, while Grid-Agent considers cyberattack-induced grid violations as part of its remediation context \cite{zhang2025gridagent}. Reviews also identify cyber-physical security, failure modes, and safe deployment as major concerns \cite{kiasari2026agenticsg,ghosh2025agenticee}. Nevertheless, systematic security evaluation is largely missing. Future power-system agents should include least-privilege tool access, typed function schemas, sandboxing, provenance checks for retrieved data, adversarial testing, and explicit authorization barriers for high-impact actions.

\subsection{Synthesis}

The reviewed works show that agentic AI in power systems is advancing through four useful design patterns: agent-in-the-loop grid intelligence, solver-in-the-loop engineering grounding, hierarchical and feedback-corrected workflows, and human-facing assistance across the grid stack. These patterns make grid analysis, simulation, dispatch support, disturbance analysis, distribution-grid workflows, and residential scheduling more interactive, modular, and capable of supporting multi-step engineering tasks \cite{chen2025xgridagent,jin2025gridmind,badmus2025powerchain,zhang2025gridagent,saha2025dragent,zheng2025agenticplanning,jia2025enhancing,cheng2025gaia,elmakroum2026hems}. At the same time, they expose five systemic failure modes: safety remains mostly empirical, robustness evidence is narrow, explanations are not yet validated, accountability is incomplete, and security evaluation remains underdeveloped.

The central lesson is that trustworthy power-system agents must be evaluated not only by task success, but also by unsafe-action rate, constraint-violation severity, robustness to rare events, explanation usefulness, audit completeness, security against misuse, and human override behavior. For power systems, the main challenge is not whether LLM agents can assist engineers, but whether their autonomy can be bounded, verified, monitored, and governed well enough for critical-infrastructure environments.

Table~\ref{tab:application_trust_summary} summarizes representative agentic AI systems in power-system applications and evaluates them through our trustworthiness dimensions.

\section{Applications in Autonomous Vehicles, Robotics, and UAVs}
\label{sec:av_robotics_uav}

\begin{figure}
    \centering
    \includegraphics[width=1\linewidth]{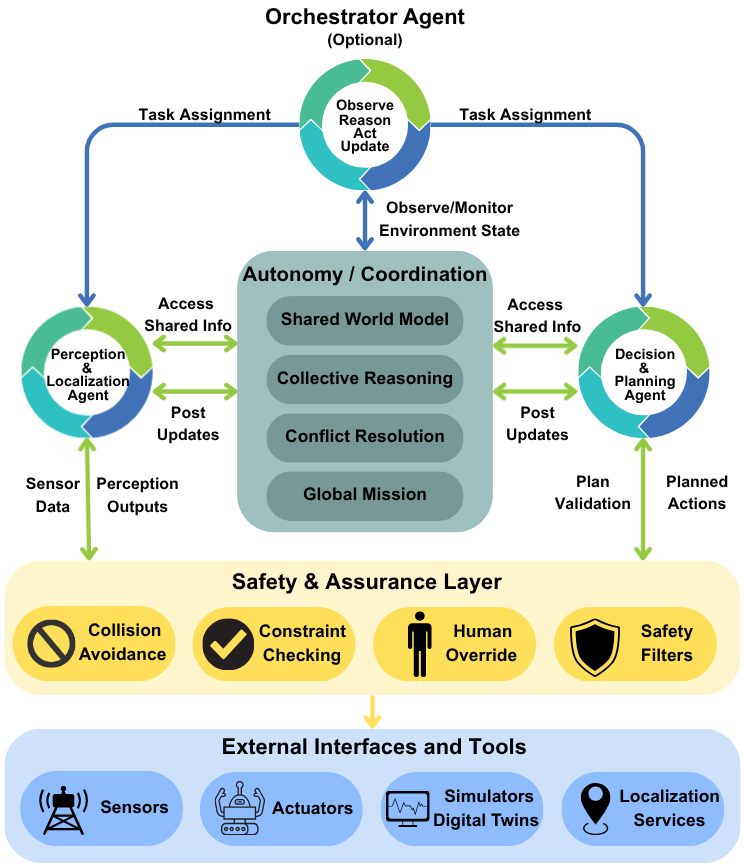}
    \caption{Robotics adaptation of the multi-agent agentic AI architecture. }
    \label{fig:uav_mas}
\end{figure}

Autonomous vehicles, robotics, and UAVs provide a high-stakes, embodied setting for trustworthy agentic AI, as decisions made by agents can directly affect physical motion, human safety, and cyber-physical infrastructure. In these systems, the agentic loop does not end with text generation or recommendation; it can extend from perception and reasoning to planning, control, and actuation (Fig.~\ref{fig:uav_mas}). This creates a sharper trust problem than in purely digital domains. The reviewed works, therefore, reveal two sides of the same story: emerging design patterns that bind high-level autonomy to safety, monitoring, and accountability mechanisms, and systemic failure modes that appear when perception, reasoning, communication, and actuation are connected in real-world environments.

\subsection{Emerging Design Patterns}

\subsubsection{Actuation-Level Safety Filters and Runtime Assurance}

A major design pattern is to place formally motivated or runtime-enforced safety mechanisms at the actuation boundary. Control Barrier Functions (CBFs) enforce forward-invariance constraints so that the system remains within a safe set during operation~\cite{ames2019control}. Black-Box Simplex architectures provide a runtime assurance structure that can switch from an advanced controller to a verified safety controller without requiring full access to the internal model~\cite{mehmood2022black}. Runtime assurance for learning-enabled autonomous driving applies a similar principle by wrapping a learned planner with a low-level safety monitor~\cite{chen2022runtime}. 

This pattern is important because it separates high-level planning from low-level safety enforcement. The high-level agent may reason, plan, or propose actions, but the final actuation layer is constrained by a safety mechanism that can reject or override unsafe behavior. Its trustworthiness value comes from reducing the risk that an unverified agent directly controls physical motion without a validated safety backstop.

\subsubsection{Layered Trust Architectures for Verified Autonomy}

A second design pattern is the use of layered trust architectures that combine regulatory, architectural, and hardware-based controls. Regulatory frameworks such as UNECE WP.29 R155 and ISO/SAE 21434 emphasize cybersecurity management, traceability, and lifecycle-based security engineering for vehicles~\cite{no2021155,siddiqui2023cybersecurity}. These standards do not, by themselves, solve the agentic trust problem, but they establish the need for auditable cybersecurity processes and accountability structures for autonomous systems.

Recent work also points toward Zero-Trust Architectures (ZTA) and hardware-based trust, such as TPM-based remote attestation, to verify and attribute autonomous decisions to authenticated agents. This pattern is important because embodied agents operate across multiple trust boundaries: sensors, onboard compute, external networks, cloud services, tools, and actuation modules. Layered trust architectures enable asking not only whether a decision was safe, but also whether it originated from an authenticated component operating within a verified system state.

\subsubsection{Multi-Layer Threat-Aware Evaluation}

A third design pattern is the use of threat-aware evaluation to expose how embodied agents fail under adversarial conditions. Several works study attacks on the perception layer, including physical-world hijacking of visual perception through ControlLoc~\cite{ma2024controlloc}, LiDAR-induced trajectory-prediction deception~\cite{lou2024first}, and latency-oriented physical-world attacks such as SlowPerception~\cite{ma2024slowperception}. Other work studies physical-layer threats such as UAV GPS spoofing and GPS spoofing detection~\cite{eldosouky2019drones,davidovich2022towards}, as well as network-layer manipulation in vehicle-to-everything platforms~\cite{hasan2020securing}.

More recent threats target the model layer directly. Physical prompt injection and visual adversarial attacks show that signs, typographic artifacts, or visual inputs can hijack vision-language driving agents without any digital manipulation~\cite{ling2026physical,zhang2024visual}. The $\alpha^3$-Bench framework is an early step toward benchmark-based evaluation of UAV agents under 6G conditions, measuring safety, robustness, and efficiency in a more unified setting~\cite{ferrag2026alpha3benchunifiedbenchmarksafety}. Together, these works show that trustworthiness in embodied agents must be evaluated across the full perception-to-actuation pipeline rather than only at a single component.

\subsubsection{Post-Hoc Explanation and Regulatory Logging}

A fourth design pattern is the use of explanation and logging mechanisms for audit, debugging, and accident reconstruction. SAFE introduces saliency-aware counterfactual explanations for deep neural network driving systems~\cite{samadi2023safe}, while SAFE-RL extends this direction to deep reinforcement-learning policies~\cite{samadi2024safe}. These methods support interpretability by identifying how changes in salient inputs could alter system behavior.

Regulatory logging and cybersecurity management frameworks also support accountability by requiring traceability and lifecycle governance~\cite{no2021155,siddiqui2023cybersecurity}. Their trustworthiness value is strongest after an incident: they can help reconstruct what happened, which component was involved, and whether the system followed the required cybersecurity or operational process. In embodied agentic systems, this is essential because failures may involve multiple layers, including perception, reasoning, tool use, control, actuation, and human approval.

\subsection{Systemic Failure Modes}

\subsubsection{Layered Verification Gap}

The first failure mode is the verification gap between layers. Low-level safety filters such as CBFs and Simplex-style architectures offer formal or runtime guarantees at the actuation boundary~\cite{ames2019control,mehmood2022black,chen2022runtime}, but the high-level reasoning layer, increasingly an LLM or VLM, remains unverified. Current designs treat the safety filter as a backstop, yet they rarely characterize how often the high-level planner produces unsafe candidates, how the filter degrades under distribution shift, or whether a safe fallback exists when both layers are uncertain.

The failure mode is therefore not only unsafe actuation, but overreliance on a low-level backstop without quantifying the residual risk that survives it. Future systems should define explicit fallback policies tied to the Operational Design Domain and determine, before deployment, when a high-level agent may advise, recommend, or directly actuate. This requires a notion of graded autonomy, in which actuation authority is granted incrementally and revoked the moment the high-level planner leaves its validated operating envelope.

\subsubsection{Cross-Layer and Physically Realizable Attacks}

The second failure mode is robustness against physically realizable and cross-layer attacks. Existing studies often evaluate threats one layer at a time: perception attacks~\cite{ma2024controlloc,lou2024first,ma2024slowperception}, physical-layer spoofing~\cite{eldosouky2019drones,davidovich2022towards}, network-layer manipulation~\cite{hasan2020securing}, and model-layer prompt injection~\cite{ling2026physical,zhang2024visual} are usually treated separately. In deployment, however, an adversary can combine these vectors, and a defense validated against one layer may fail against another.

This matters because embodied agents depend on a continuous chain from sensing to interpretation, planning, communication, and actuation. A small perturbation at one layer can propagate into unsafe behavior at another. The field therefore needs combined threat models and evaluation protocols that stress the full perception-to-actuation pipeline under simultaneous, adaptive attacks rather than single-vector benchmarks.

\subsubsection{Faithful and Timely Interpretability}

The third failure mode is the lack of faithful and timely interpretability. Saliency-based and counterfactual explanations support accident reconstruction and pre-deployment auditing~\cite{samadi2023safe,samadi2024safe}, but they are often produced after the fact and their faithfulness is seldom validated. A plausible explanation may identify visually salient regions or counterfactual changes, but it may not faithfully represent the causal basis of the agent's decision.

This limitation is especially important in embodied systems because decisions unfold on millisecond timescales. Explanations that are useful only for forensic review may not support real-time operator oversight. Future explanations should be evaluated for whether they actually improve operator decisions and trust calibration, not only for plausibility or visual appeal.

\subsubsection{Missing End-to-End Provenance and Attribution}

The fourth failure mode is the absence of end-to-end, tamper-evident provenance. Regulatory frameworks mandate cybersecurity management and traceability~\cite{no2021155,siddiqui2023cybersecurity}, yet they do not specify how to reconstruct the agentic chain that links perception, reasoning, tool use, actuation, and human approval into a single auditable record. This leaves a gap between compliance-oriented traceability and agent-level accountability.

For embodied agents, accountability must answer a concrete chain of questions: what the agent perceived, how it interpreted the scene, which model or tool generated the decision, what constraints were checked, what action was selected, whether a human approved it, and whether the acting component was authenticated. Pairing such provenance with Zero-Trust Architectures and hardware-based attestation, so that each decision is both verifiable and attributable, remains an open design problem.

\subsubsection{Fragmented and Capability-Driven Benchmarks}

The fifth failure mode is the lack of standardized, trustworthiness-oriented benchmarks for embodied agents. The $\alpha^3$-Bench framework is an early step for UAV agents under 6G conditions~\cite{ferrag2026alpha3benchunifiedbenchmarksafety}, but most evaluations remain capability-driven or attack-specific. They typically measure whether the system completes the task or withstands a specific attack, rather than jointly evaluating safety, robustness, interpretability, accountability, and security.

A reusable benchmark should jointly measure these trustworthiness dimensions across driving, manipulation, and aerial tasks. It should report unsafe-action rate, constraint-violation severity, human-override behavior, attack success rate, explanation usefulness, and audit completeness alongside task success. Without this shift, the field risks producing embodied agents that appear capable in isolated tests but remain untrusted under real deployment conditions.

\subsection{Synthesis}

The reviewed works show that embodied agentic AI can be made safer through four useful design patterns: actuation-level safety filters, layered trust architectures, threat-aware evaluation, and post-hoc explanation with regulatory logging. These patterns provide important building blocks for bounding autonomy, detecting adversarial failures, reconstructing incidents, and connecting physical actions to cybersecurity and accountability processes.

At the same time, the evidence remains fragmented across isolated mechanisms and threat models. The main systemic failure modes are the verification gap between high-level reasoning and low-level safety, vulnerability to cross-layer and physically realizable attacks, insufficiently faithful and timely explanations, missing end-to-end provenance, and the absence of standardized trustworthiness-oriented benchmarks. The central lesson is that trustworthy embodied agents must be evaluated at the interfaces where perception becomes interpretation, interpretation becomes planning, and planning becomes physical action. Reliable deployment requires graded autonomy, verified low-level safety layers, combined threat models, real-time interpretability, tamper-evident provenance, and benchmarks that measure trustworthiness rather than capability alone.

\section{Applications in High-Performance Computing}  
\label{sec:hpc}
\begin{figure}
    \centering
    \includegraphics[width=1\linewidth]{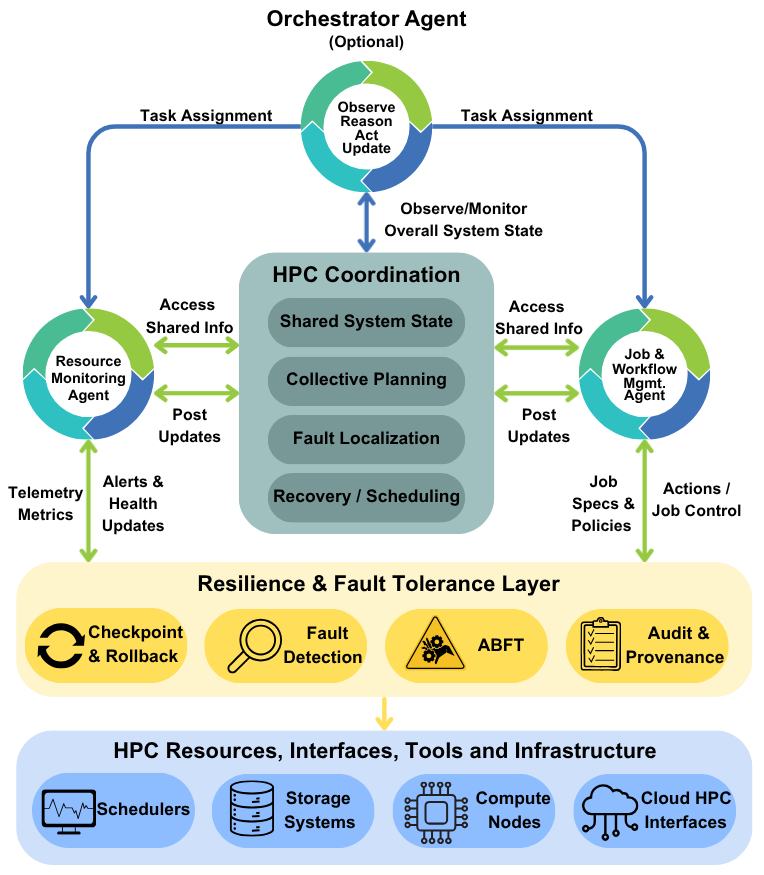}
    \caption{High-performance computing adaptation of the multi-agent agentic AI architecture.}
    \label{fig:hpc_mas}
\end{figure}

As agentic systems increasingly process large-scale data within high-fidelity simulations, complex digital twins, and heterogeneous systems, the integration of these agents with High-Performance Computing (HPC) resources has become essential for feasibility. Beyond established conventions of trust and resilience in agentic systems, the incorporation of HPC resources introduces a new critical dimension: \textit{resilience at scale} \cite{dongarra2015fault}. To achieve resilience at scale, efficient methods addressing \textit{faults} in system operations are vital for stable, large-scale agentic systems; a critical topic in modern agentic system design using HPC resources \cite{Cappello2009Toward}. Works involving HPC resources, whether to support agentic deployment or to enable the agent to act upon, focus on innovating fault tolerance for systemic failure modes arising from hardware degradation and unsafe agent interactions in large-scale systems (Fig.~\ref{fig:hpc_mas}).

%In the performance optimization of agentic systems using High-Performance Computing (HPC) resources, the notion of 'trust' is typically established relative to the application domain. Considering the HPC resources themselves, the primary objective is establishing 'resilience at scale' within a wide range of application domains \cite{dongarra2015fault}. This is studied within the scope of 'fault tolerance,' ensuring that critical hardware and software infrastructure functions at exascale.

\subsection{Emerging Design Patterns}

\subsubsection{Fault Tolerance}
As applications push exascale performance, the likelihood of faults increases proportionally with the complexity of data distributions, frequent inter-process communication (IPC), and the maximization of parallel hardware \cite{Losada2020Fault}. These often manifest as instances of data corruption, message loss, hardware defects and software interrupts, among others. To mitigate these risks, \textit{Fault Tolerance} in HPC systems focuses on evaluating, diagnosing and preventing faults or failures in both software and hardware components, regardless of application \cite{dongarra2015fault}. Methods for fault tolerance range from general purpose \textit{Checkpointing} rollback protocols and software-oriented \textit{Algorithm-Based Fault Tolerance} (ABFT) \cite{abft_og} to distributed repair mechanisms, among similar modalities. The integration of these fault-tolerance methods improves resilience with minimal impact on performance, enabling stable agentic deployments.

\subsubsection{Multi-Agent Systems}
To leverage interactions between agents to further system resilience, multi-agent deployments effectively mimic HPC parallelism, enabling system recovery through coordination. This adopts fault tolerance through 'self-healing' mechanisms, which tasks agents with features for resource monitoring, crash detection and recovery actions to incorporate system state data in decisions towards system stability \cite{Rajput2020Multi-agent,Park2005MAS}. These self-healing mechanisms are intuitively integrated into existing agent actions on HPC topologies, such as workload distribution and anomaly detection \cite{Sidorov2016Methods,Vinay2025CoMAS-HPC}. 

In other instances, traditional rollback mechanisms can be applied through multi-agent interactions as a transactional relation \cite{Chang2025SagaLLM}. These systems match actions with their reversals while maintaining system interaction logs, enabling dynamic return to safe states without complex coordination between agents.

%use checksums to verify data, which has seen success in distributed applications \cite{phylogeneticHPC,Hespe2022ReStore}. In agentic systems however, these solutions are limiting through the overheads of I/O in real-time scenarios and the complexities in generating 'save states' amidst complex system dynamics.

%\subsubsection{Checkpointing}
%Traditional safeguards focus on the design of \textit{Checkpointing} protocols within the system, effectively marking rollback points to resume computation from as a 'safe state.' However, these protocols introduce significant I/O overheads that bottleneck performance and limit scalability -- a central challenge in performance optimization.

%\subsubsection{Self-Healing}

\subsection{Systemic Failure Modes}

\subsubsection{Hardware Failures}

The first and fundamental source of failure stems from hardware degradation, which introduces system instability and compromises the predictability of agentic actions. In exascale environments, these failures range from total system crashes to data corruption \cite{Cappello2009Toward}. While standard checkpointing and self-healing mechanisms can isolate localized node failures in multi-agent systems, they are ineffective at detecting and responding to data corruption \cite{Rahman2021PEPPA-X:}. This directly degrades agentic system trust from a hardware perspective, as increased workload scales increase the potential for data corruption and hardware failure.

\subsubsection{Memory Latency}

The second primary failure mode covers latency for expensive computations and memory access. Computational overheads are consistently optimized per application, though memory access latency has been shown to be a primary limitation to scaling AI workloads \cite{Gholami2024AI}. This latency is present within context-window updates, cache accesses and weight retrievals for LLM-based applications, limiting scalability across large heterogeneous systems. This results in trust degradation in agentic systems, as lengthy decision times lead to unpredictable behavior when conflicting notions of the current system state arise.

\subsection{Synthesis}
The reviewed works demonstrate the importance in overlap between hardware robustness and agentic trust, as AI applications consistently face resilience concerns on the operational level. Patterns for design in resilience mechanisms highlight the efficacy of checkpointing, self-healing and mult-agent coordination methods, though this is reliant on domain-specific implementations. Given the limitations of hardware degradation and memory latency, there is a severe lack of frameworks detailing resilience in extreme-scale agentic systems to broadly address these challenges. Such solutions would consider a combination of adaptive mechanisms for memory efficiency and decentralized operational conditions across a wide variety of domain applications.

\section{Applications in Communication Networks}
\label{sec:comm}
\begin{figure}
    \centering
    \includegraphics[width=1\linewidth]{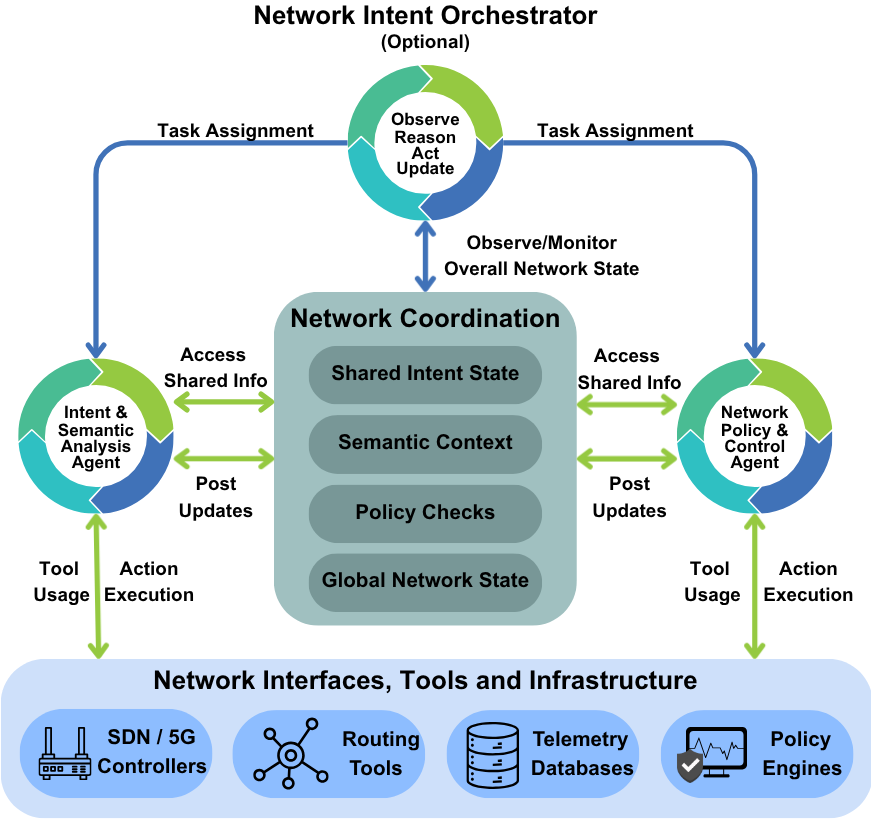}
    \caption{Communication networks adaptation of the multi-agent agentic AI architecture.}
    \label{fig:comm_mas}
\end{figure}

Agentic AI enters communication networks by turning high-level intent, semantic content, or operational context into actions that affect the network. This makes communication systems more adaptive and easier to operate, but it also changes the trust problem~\cite{enright2022learning}. The system is no longer only transmitting information or optimizing links; it is interpreting intent, reconstructing meaning, selecting tools, and in some cases triggering network behavior (Fig.~\ref{fig:comm_mas}). The reviewed works, therefore, reveal two sides of the same story: emerging design patterns that make communication systems more agentic, and systemic failure modes that appear when interpretation is connected to actuation.

\subsection{Emerging Design Patterns}

\subsubsection{Intent Abstraction and Human-in-the-Loop Control}

A major design pattern is the abstraction of human intent into machine-interpretable network objectives. LUMI is an early example of this pattern: it allows operators to express network intents in natural language, maps them into the Nile intermediate representation, and compiles them into executable configurations~\cite{jacobs2021lumi}. This creates a human-agent network loop in which the agent is not merely assisting the operator but actively mediating between human goals and network behavior.

LLM-based intent extraction for 5G core networks extends this idea by using language models as the intent-understanding layer for network management~\cite{manias2024intent}. Semantic routing further strengthens this direction by routing requests across different model pathways to improve orchestration performance and efficiency~\cite{manias2024semanticrouting}. Together, these works show that intent abstraction can improve usability, scalability, and adaptability in network management. Their trustworthiness value comes from making intent processing more structured and partially inspectable, especially when intermediate representations or routing decisions expose how a request is handled.

\subsubsection{Semantic Communication: From Bits to Meaning}

A second design pattern is the move from syntactic communication to semantic communication. Instead of optimizing only for bit-level or symbol-level recovery, semantic communication systems aim to preserve task-relevant meaning. DeepSC demonstrates this shift through transformer-based semantic encoding and decoding for text transmission under noisy channel conditions~\cite{xie2021deepsc}. DeepSC-S applies the same principle to speech, showing that speech communication can be optimized around recovered semantic content and perceptual quality rather than exact signal reconstruction~\cite{weng2021speech}.

This pattern is important because it makes the communication process itself more intelligent. The channel is no longer treated only as a medium for symbols; it becomes part of a meaning-preserving pipeline. These works show strong robustness under difficult channel conditions and establish semantic-level optimization as a useful design principle. In the context of agentic systems, this matters because agents often act on interpreted meaning rather than raw signals.

\subsubsection{Security-First Enhancements and Agent Evaluation}

A third design pattern is the introduction of security and evaluation mechanisms around agentic communication pipelines. In semantic communication, physical-layer semantic encryption and obfuscation protect the meaning of transmitted content rather than only the raw signal~\cite{qin2023securesemantic}. This is important because semantic systems create new forms of exposure: an attacker may target the meaning itself, even when the transmitted symbols are not fully recovered.

Beyond communication-specific systems, ToolEmu and Agent Security Bench provide evaluation frameworks for tool-using LLM agents~\cite{ruan2024toolemu,zhang2025asb}. These works are not network-management papers, but they are directly relevant to communication agents because future network agents will rely on tools, memory, prompts, and multi-step decision pipelines. They show that agentic systems should be tested not only for task success, but also for unsafe tool use, prompt injection, memory manipulation, and adversarial behavior. This gives communication-network research a useful model for pre-deployment stress testing and security benchmarking.

\subsection{Systemic Failure Modes%: What Breaks and Why
}

\subsubsection{Intent Ambiguity Leading to Unsafe Actuation}

The first failure mode appears at the boundary between human intent and system action. Intent-based systems assume that a user goal can be translated into a correct network behavior, but natural language intents are often ambiguous, underspecified, or context-dependent. LUMI makes this translation more structured through Network Intent LanguagE (Nile), but it does not fully solve the problem of whether the interpreted intent is safe, authorized, or complete~\cite{jacobs2021lumi}. LLM-based intent extraction increases flexibility, but it also increases exposure to prompt manipulation and unsafe intent acceptance~\cite{manias2024intent}.

INTENDER makes this risk concrete by showing that intent parsing and state transitions in intent-based networking can become attack surfaces~\cite{kim2023intender}. This means the issue is not only whether the agent understands the user, but whether the system can prevent a misunderstood or malicious intent from becoming an executable network action. The failure mode is therefore a missing trust layer between interpretation and actuation requiring architectural changes and trust-driven mechanisms~\cite{haikal2026bridging}.

\subsubsection{Semantic Misalignment}

The second failure mode appears when a system preserves meaning in a shallow sense but fails to preserve the correct meaning for the task. DeepSC and DeepSC-S show that semantic communication can improve robustness under noisy conditions~\cite{xie2021deepsc,weng2021speech}. However, semantic similarity or perceptual quality does not necessarily guarantee that the reconstructed message is operationally correct.

This is referred to as semantic misalignment. A reconstructed message may appear plausible, fluent, or semantically close, while still altering the instruction, omitting important context, or leading to an unsafe interpretation. This is especially dangerous when semantic communication feeds into decision-making or network-control agents. In that setting, the key question is not only whether the receiver recovered similar meaning, but whether the recovered meaning is valid for the downstream action.

\subsubsection{Lack of End-to-End Accountability}

The third failure mode is the lack of accountability across the full agentic communication pipeline. Current systems often provide only partial visibility into a single component. LUMI exposes an intermediate intent representation~\cite{jacobs2021lumi}. Semantic communication papers report semantic or perceptual performance metrics~\cite{xie2021deepsc,weng2021speech}. Agent benchmarks expose failures in tool-using agents~\cite{ruan2024toolemu,zhang2025asb}. However, these pieces do not yet form a comprehensive recoverable accountability chain.

This matters because failures can originate from many places: the user intent, the intent parser, the semantic encoder, the reconstructed message, the routing decision, the tool call, or the final network action. Without provenance logs, policy-check records, decision traces, and responsibility attribution, it is difficult to determine why a failure occurred or who should be held responsible. For trustworthy communication agents, accountability must be built into the entire workflow rather than added after deployment.

\subsection{Synthesis}

The reviewed works show that agentic AI in communication networks is advancing through three useful design patterns: intent abstraction, semantic communication, and security-oriented evaluation. These patterns make networks easier to operate, more robust under uncertainty, and better able to support autonomous workflows. At the same time, they expose three systemic failure modes: ambiguous intent can lead to unsafe actuation, semantic preservation can still produce incorrect meaning, and partial visibility does not provide end-to-end accountability.

The central lesson is that trustworthy agentic communication systems must be evaluated at the interfaces where interpretation becomes consequence. The critical boundaries are human intent to machine action, transmitted meaning to task correctness, and agent decision to responsibility. Current work provides important building blocks, but reliable deployment requires stronger formal constraints, runtime safeguards, adversarial testing, and provenance mechanisms that connect the full pipeline.

\section{Cross-Domain Insights}
\label{sec:cross_domain}

\begin{table*}[t]
\centering
\caption{Representative agentic AI works across application domains through a trustworthiness lens.}
\label{tab:application_trust_summary}
\tiny
\renewcommand{\arraystretch}{0.92}
\setlength{\tabcolsep}{2pt}
\resizebox{\textwidth}{!}{%
\begin{tabular}{p{2.6cm} p{2.7cm} p{5.0cm} c c c c c}
\toprule
\textbf{Work} &
\textbf{Focus / Type} &
\textbf{Key Mechanism or Insight} &
\textbf{Safety} &
\textbf{Rob.} &
\textbf{Interp.} &
\textbf{Audit} &
\textbf{Sec.} \\
\midrule

\multicolumn{8}{l}{\textbf{Power Systems} (Sec.~\ref{sec:ps_main})} \\
\midrule

PowerAgent \cite{zhang2025poweragent}
& Grid operation/planning roadmap
& Foundation model + MCP + workflows
& \Partial & \Partial & \Partial & \Partial & \Partial \\

Foundation models \cite{xie2026foundationmodels}
& AI--grid interaction
& Foundation-model integration
& \Partial & \Partial & \Partial & \Weak & \Partial \\

Smart-grid review \cite{kiasari2026agenticsg}
& Safe/explainable smart-grid agents
& Review and taxonomy of trustworthy agents
& \Partial & \Partial & \Partial & \Partial & \Partial \\

Electrical-engineering review \cite{ghosh2025agenticee}
& Agentic AI in EE
& Taxonomy and failure-mode analysis
& \Partial & \Partial & \Partial & \Partial & \Partial \\

X-GridAgent \cite{chen2025xgridagent}
& Grid analysis
& Planning--coordination--action hierarchy
& \Partial & \Partial & \Full & \Partial & \Weak \\

GridMind \cite{jin2025gridmind}
& OPF and contingency analysis
& LLM agents with deterministic solvers
& \Partial & \Partial & \Full & \Partial & \Weak \\

PowerChain \cite{badmus2025powerchain}
& Distribution-grid workflows
& Verified tool-call trajectories
& \Partial & \Full & \Partial & \Full & \Weak \\

Grid-Agent \cite{zhang2025gridagent}
& Violation remediation
& Planning, validation, and rollback
& \Full & \Partial & \Partial & \Partial & \Partial \\

DrAgent \cite{saha2025dragent}
& Fault/disturbance analysis
& ReWOO + tools + argument repair
& \Partial & \Partial & \Full & \Partial & \Weak \\

Agentic planning \cite{zheng2025agenticplanning}
& Simulation orchestration
& Planning, validation, and enhancement agents
& \Partial & \Partial & \Partial & \Partial & \Weak \\

Jia et al. \cite{jia2025enhancing}
& Power-system simulations
& RAG + reasoning + feedback acting
& \Partial & \Full & \Partial & \Partial & \Weak \\

GAIA \cite{cheng2025gaia}
& Dispatch support
& Fine-tuned dispatch LLM
& \Partial & \Partial & \Partial & \Weak & \Partial \\

Local fine-tuned LLM \cite{slavchev2026finetuned}
& Secure local analysis
& Offline RAG + planner + sandbox tools
& \Partial & \Partial & \Partial & \Partial & \Full \\

HEMS \cite{elmakroum2026hems}
& Residential scheduling
& Orchestrator + appliance agents
& \Partial & \Partial & \Full & \Partial & \Partial \\

\midrule
\multicolumn{8}{l}{\textbf{Autonomous Vehicles, Robotics, and UAVs} (Sec.~\ref{sec:av_robotics_uav})} \\
\midrule

Control Barrier Functions \cite{ames2019control}
& Safety filter
& Forward-invariance constraints enforced at the actuation boundary
& \Full & \Partial & \Weak & \Weak & \Weak \\

Black-Box Simplex \cite{mehmood2022black}
& Runtime assurance
& Switching to a verified safety controller without model access
& \Full & \Full & \Weak & \Weak & \Weak \\

RTA for learning-enabled AD \cite{chen2022runtime}
& Runtime assurance
& Wraps a learned planner with a low-level safety monitor
& \Full & \Partial & \Weak & \Weak & \Weak \\

$\alpha^3$-Bench \cite{ferrag2026alpha3benchunifiedbenchmarksafety}
& Evaluation
& Safety, robustness, and efficiency benchmark for UAV agents over 6G
& \Full & \Full & \Partial & \Weak & \Partial \\

Physical prompt injection \cite{ling2026physical}
& Threat: model layer
& Typographic injection hijacks VLM driving agents
& \Partial & \Partial & \Weak & \Weak & \Full \\

Visual adversarial attack \cite{zhang2024visual}
& Threat: perception
& Adversarial inputs mislead VLM autonomous-driving perception
& \Partial & \Full & \Weak & \Weak & \Full \\

ControlLoc \cite{ma2024controlloc}
& Threat: perception
& Physical-world hijacking of visual perception
& \Partial & \Full & \Weak & \Weak & \Full \\

LiDAR trajectory attack \cite{lou2024first}
& Threat: perception
& LiDAR-induced trajectory-prediction deception
& \Partial & \Full & \Weak & \Weak & \Full \\

SlowPerception \cite{ma2024slowperception}
& Threat: perception
& Physical-world latency attack on visual perception
& \Partial & \Full & \Weak & \Weak & \Full \\

SAFE \cite{samadi2023safe}
& Explanation
& Saliency-aware counterfactual explanations for DNN driving systems
& \Weak & \Weak & \Full & \Partial & \Weak \\

SAFE-RL \cite{samadi2024safe}
& Explanation
& Saliency-aware counterfactual explainer for deep RL policies
& \Weak & \Weak & \Full & \Partial & \Weak \\

UNECE WP.29 R155 \cite{no2021155}
& Standard
& Cybersecurity management system requirements for vehicles
& \Partial & \Weak & \Weak & \Full & \Full \\

ISO/SAE 21434 \cite{siddiqui2023cybersecurity}
& Standard
& Automotive cybersecurity engineering lifecycle
& \Partial & \Weak & \Weak & \Full & \Full \\

GPS spoofing countermeasure \cite{eldosouky2019drones}
& Defense: physical layer
& Game-theoretic protection of UAVs against GPS spoofing
& \Partial & \Partial & \Weak & \Weak & \Full \\

GPS spoofing detection \cite{davidovich2022towards}
& Defense: physical layer
& Camera-stream detection of drone GPS spoofing
& \Weak & \Partial & \Weak & \Weak & \Full \\

V2X security \cite{hasan2020securing}
& Defense: network layer
& Securing vehicle-to-everything communication platforms
& \Weak & \Partial & \Weak & \Weak & \Full \\

\midrule
\multicolumn{8}{l}{\textbf{High-Performance Computing} (Sec.~\ref{sec:hpc})} \\
\midrule

Checkpointing \cite{dongarra2015fault}
& Resilience amidst hardware faults
& Safe state marking and rollback procedures to enable restoration in the event of failure
& \Partial & \Full & \Weak & \Weak & \Weak \\

Transactional Rollback \cite{Chang2025SagaLLM}
& Logging-based checkpointing
& Pair actions with reversing actions for recovery
& \Partial & \Full & \Full & \Partial & \Partial \\

Algorithm-Based Fault Tolerance \cite{abft_og}
& Operational resilience
& Checksum-verified operations to separate verification from hardware
& \Partial & \Full & \Weak & \Partial & \Partial \\

Self-Healing \cite{Park2005MAS}
& Self-monitoring and recovery
& Agentic profiling of system status and responding actions to combat faults during system processing
& \Partial & \Full & \Partial & \Partial & \Weak \\

Multi-Agent Coordination \cite{Rajput2020Multi-agent}
& Agentic coordination
& Agentic synchronization to enable self-verification of faulty processes
& \Partial & \Full & \Partial & \Full & \Partial \\

\midrule
\multicolumn{8}{l}{\textbf{Communication Networks} (Sec.~\ref{sec:comm})} \\
\midrule

LUMI \cite{jacobs2021lumi}
& Intent abstraction and HITL control
& Natural-language intents compiled into executable network configurations
& \Partial & \Partial & \Partial & \Partial & \Weak \\

Intent Extraction for 5G \cite{manias2024intent}
& LLM-based intent extraction
& LLMs used for intent understanding in 5G core management
& \Partial & \Partial & \Partial & \Partial & \Weak \\

Semantic Routing \cite{manias2024semanticrouting}
& LLM-assisted semantic routing
& Semantic routing improves orchestration efficiency and modularity
& \Partial & \Full & \Partial & \Partial & \Partial \\

INTENDER \cite{kim2023intender}
& Security testing for IBN
& Intent parsing and state transitions exposed as attack surfaces
& \Partial & \Partial & \Partial & \Partial & \Full \\

DeepSC \cite{xie2021deepsc}
& Semantic communication
& Meaning-oriented communication under noisy channels
& \Weak & \Full & \Weak & \Weak & \Weak \\

DeepSC-S \cite{weng2021speech}
& Semantic speech communication
& Semantic speech transmission robust to channel degradation
& \Weak & \Full & \Weak & \Weak & \Weak \\

Secure Semantic Communication \cite{qin2023securesemantic}
& Security-first semantic communication
& Protects semantic content through physical-layer encryption
& \Partial & \Partial & \Weak & \Weak & \Full \\

ToolEmu \cite{ruan2024toolemu}
& Agent safety evaluation
& Sandbox-based evaluation for unsafe tool-using behavior
& \Full & \Full & \Full & \Full & \Full \\

Agent Security Bench \cite{zhang2025asb}
& Agent security benchmarking
& Adversarial benchmarking across prompts, tools, and workflows
& \Full & \Full & \Full & \Full & \Full \\

\bottomrule
\end{tabular}%
}

\vspace{1mm}
\begin{flushleft}
\footnotesize
\textit{Legend:} $\checkmark$ = strong support; $\triangle$  = partial support; $\times$ = weak or missing support.
Rob.: robustness/reliability; Interp.: transparency/interpretability; Audit: accountability/auditability; Sec.: privacy/security.
HITL: human-in-the-loop; IBN: intent-based networking; AD: autonomous driving; VLM: vision-language model; MCP: Model Context Protocol.
\end{flushleft}
\end{table*}

The reviewed application domains show that trustworthy agentic AI is not defined by a single architecture, benchmark, or evaluation criterion. Figure ~\ref{fig:evolution_of_benchmarks} makes this explicit: with the exception of dedicated evaluation frameworks such as Claw-Eval~\cite{ye2026clawevaltrustworthyevaluationautonomous}, whose entire purpose is to probe all five dimensions at once \cite{ruan2024toolemu,zhang2025asb}, no deployed or applied agentic system reviewed in this survey scores strongly on more than two or three trustworthiness dimensions simultaneously, and which dimension is left uncovered differs systematically by domain rather than by architecture. This is because power systems, autonomous vehicles and UAVs, high-performance computing, and communication networks all instantiate the same underlying agentic pipeline, perception, reasoning, planning, tool use, communication/coordination, memory update, validation, action, monitoring, and audit (Section~\ref{sec:trust_dimensions}), but differ in \emph{where along that pipeline interpretation becomes consequence}, how severe and reversible the resulting action is, and how much of the pipeline currently carries verifiable evidence rather than best-effort engineering practice. Power systems place that boundary at the solver-to-grid interface; robotics and UAVs place it at the planner-to-actuator interface; HPC places it at the detection-to-recovery interface; communication networks place it at the intent-to-configuration interface. The consequence of crossing that boundary, not the underlying model architecture, is what determines how much trust engineering a given domain requires.

This shared lens is what makes cross-domain synthesis possible: it separates what is genuinely domain-specific (grid physics, actuation dynamics, HPC fault models, network intents) from what is a general property of agentic AI as a class of systems. Section~\ref{sec:recurring_design_patterns} identifies the design patterns that recur across domains precisely because they answer the same underlying trust question under different physical constraints. Section~\ref{sec:shared_failure_modes} shows that these same patterns leave behind a common residue of unresolved risk regardless of domain.

\subsection{Recurring Design Patterns}
\label{sec:recurring_design_patterns}

\subsubsection{Tool-Grounded Reasoning and Domain-Expert Execution}
A recurring pattern across all four domains is the separation of semantic reasoning from trusted execution: the LLM interprets, decomposes, and plans, but a deterministic, verifiable component actually acts. In power systems, this means LLMs interpret operator requests, decompose tasks, and retrieve context, but hand off numerical work to power-flow, optimal-power-flow, and contingency-analysis solvers rather than computing grid state themselves \cite{jin2025gridmind,  chen2025xgridagent,jia2025enhancing,badmus2025powerchain}. Communication-network agents follow the same logic in a different substrate: natural-language intents and semantic content are converted into intermediate representations or routing decisions before they are allowed to touch network configuration \cite{jacobs2021lumi,  manias2024intent,manias2024semanticrouting}, and tool-using agents are stress-tested against this same handoff boundary \cite{ruan2024toolemu,zhang2025asb}. Robotics and UAV systems enforce the identical separation at the actuation boundary through Control Barrier Functions and Simplex-style runtime monitors that sit between a learned planner and the motor commands it produces \cite{ames2019control,mehmood2022black,chen2022runtime}. Read together, these results indicate that trustworthy agentic AI is strongest not when the LLM is granted final authority, but when it is demoted to an orchestrator whose outputs are grounded in domain-specific, independently verifiable execution, directly operationalizing the safety and robustness dimensions of Section~\ref{sec:trust_dimensions} at the tool-use and validation points of the ten-point workflow.

\subsubsection{Modular and Hierarchical Agent Organization}
A second pattern is the decomposition of a single monolithic agent into specialized roles connected by a coordination layer. Power-system deployments divide responsibility across planning, execution, validation, enhancement, and appliance-level agents \cite{zhang2025gridagent, zheng2025agenticplanning,  elmakroum2026hems}, while HPC multi-agent systems separate resource monitoring, crash detection, recovery, workload distribution, and anomaly detection into distinct roles \cite{Rajput2020Multi-agent,Park2005MAS, Sidorov2016Methods, Vinay2025CoMAS-HPC}. Communication-network research shows a parallel move toward modular interpretation-and-action pipelines through intent abstraction, semantic routing, and dedicated tool-use benchmarking \cite{jacobs2021lumi, manias2024semanticrouting, ruan2024toolemu,zhang2025asb}. This convergence suggests that modularity is not an engineering convenience but a trust mechanism in its own right: isolating responsibilities keeps any single component's failure mode local and inspectable, which is a precondition for the accountability dimension discussed below, rather than a byproduct of it.

\subsubsection{Feedback, Validation, Rollback, and Recovery Loops}
A third pattern treats trustworthy behavior as an iterative loop rather than a one-shot response. Power-system agents use feedback correction, malformed-argument repair, verified workflow trajectories, and rollback \cite{jia2025enhancing, saha2025dragent, badmus2025powerchain, zhang2025gridagent}; HPC systems rely on checkpointing, algorithm-based fault tolerance, and self-healing to survive faults at scale \cite{dongarra2015fault, Cappello2009Toward, abft_og, Hespe2022ReStore, Rajput2020Multi-agent, Park2005MAS}; robotics and UAV systems achieve the equivalent through runtime assurance architectures that can revert to a verified safety controller the instant a learned component leaves its validated envelope \cite{ames2019control,mehmood2022black,chen2022runtime}. Despite operating on completely different timescales (milliseconds for actuation, seconds for grid dispatch, and potentially hours for HPC job recovery), all three implementations share the same structural commitment: a system is trusted not because it avoids error, but because it can detect, correct, and recover from error before the consequences propagate.

\subsubsection{Bounded Autonomy over Unrestricted Autonomy}
The fourth and most consequential pattern is that the most credible deployments in every domain are bounded-autonomy systems, not fully autonomous ones. Power-system agents remain grounded in solvers, advisory operation, and sandboxed, rollback-capable execution \cite{cheng2025gaia,jin2025gridmind,zhang2025gridagent}; robotics and UAV systems require actuation-level safety filters precisely because a high-level planning error can cause physical harm \cite{ames2019control,mehmood2022black,chen2022runtime}; communication-network agents require an explicit trust layer between intent interpretation and executable behavior \cite{jacobs2021lumi,manias2024intent,kim2023intender}; and HPC agents require supervision before autonomous self-healing or workload redistribution is trusted at scale \cite{Rajput2020Multi-agent, Park2005MAS, Vinay2025CoMAS-HPC}. That this pattern reproduces itself independently across four unrelated engineering communities is the strongest cross-domain evidence in this survey that unrestricted agentic autonomy is not currently a viable deployment target in any critical-infrastructure setting, only its boundary location changes.

\subsection{Shared Failure Modes}
\label{sec:shared_failure_modes}

\subsubsection{Unsafe Translation from Reasoning to Action}
The design patterns above do not eliminate risk; they relocate it to the interface where interpretation becomes consequence, and that interface remains the most critical failure point in every domain. In power systems, an incorrect recommendation can violate grid constraints or respond too slowly to a rare event \cite{cheng2025gaia,zhang2025gridagent}; in robotics and UAVs, perception, planning, or control errors propagate directly into unsafe physical actuation \cite{ma2024controlloc,lou2024first,ma2024slowperception}; in communication networks, an ambiguous or malicious intent can become an executable network action \cite{jacobs2021lumi,manias2024intent,kim2023intender}. The common thread is that grounding an agent in tools, solvers, or intermediate representations narrows this gap without closing it; the translation step itself is rarely subjected to independent, adversarial verification.

\subsubsection{Robustness Gaps under Rare, Degraded, or Adversarial Conditions}
Most current evaluations demonstrate feasibility on curated tasks rather than the tail of the operating distribution. Power-system evaluations lean on test systems, curated simulation tasks, and historical disturbance records \cite{jin2025gridmind,jia2025enhancing,badmus2025powerchain,saha2025dragent,elmakroum2026hems}; robotics and UAV systems are shown to be vulnerable to adversarial patches, trajectory-prediction attacks, sensor spoofing, and GPS jamming \cite{ma2024controlloc,lou2024first,ma2024slowperception,eldosouky2019drones,davidovich2022towards}; HPC systems must tolerate data corruption, message loss, and scaling-induced hardware failures \cite{Losada2020Fault,dongarra2015fault}; and communication-network agents must withstand malicious intents, semantic misalignment, and unsafe tool use \cite{kim2023intender,xie2021deepsc,weng2021speech,ruan2024toolemu,zhang2025asb}. None of these domains yet reports the kind of systematic operating-envelope characterization, comparable to an Operational Design Domain analysis \cite{Shinde2024}, that would say with confidence where an agent's guarantees end.

\subsubsection{Partial Auditability instead of End-to-End Accountability}
Several systems expose fragments of the workflow, a plan, a tool call, an intermediate representation, a reasoning trace, but fragment-level visibility does not add up to a reconstructable accountability chain. Power-system systems provide partial auditability through workflow trajectories, solver state, and tool-call plans \cite{badmus2025powerchain,jin2025gridmind,chen2025xgridagent,saha2025dragent}; robotics and UAV systems require tamper-evident logs spanning perception, decision, and actuation for forensic review \cite{samadi2023safe,samadi2024safe}; communication-network systems need provenance across intent, semantic encoding, routing, and final network action \cite{jacobs2021lumi,xie2021deepsc,weng2021speech,ruan2024toolemu,zhang2025asb}. In every domain, the gap is the same: visibility into one component does not guarantee that a failure can be traced back to the decision, tool call, or human approval that caused it.

\subsubsection{Expanding Security Surface from Tools, Memory, and Multi-Agent Coordination}
The same tool use, retrieval, and coordination mechanisms that ground agent outputs and enable modularity also expand the attack surface. Power-system agents introduce risk through APIs, simulators, retrieval pipelines, and memory modules \cite{zhang2025poweragent,chen2025xgridagent,jin2025gridmind,jia2025enhancing,elmakroum2026hems}; robotics and UAV agents face attacks that span physical, network, and model layers simultaneously \cite{eldosouky2019drones,davidovich2022towards,hasan2020securing,ling2026physical,zhang2024visual}; communication-network agents face malicious intent parsing, prompt injection, and memory manipulation \cite{kim2023intender,ruan2024toolemu,zhang2025asb}; and HPC multi-agent systems must secure recovery, coordination, and workload-management channels against exactly the kind of manipulation their autonomy depends on \cite{Rajput2020Multi-agent, Park2005MAS, Sidorov2016Methods, Vinay2025CoMAS-HPC}. This is a structural tension rather than an engineering oversight: every mechanism identified in Section~\ref{sec:recurring_design_patterns} as strengthening trust also introduces a new interface that an adversary can target.

\section{Open Challenges and Future Directions}
% try to be more bold  (opportunities for possible directions) more visionary
\label{sec:open_challenges}

\subsection{Mandating Trustworthiness Metrics Alongside Task Success}
Most current work still reports task success, workflow completion, or automation gains as the primary outcome measure. The field needs to shift toward unsafe-action rate, constraint-violation severity, robustness to rare events, explanation usefulness, audit completeness, and human-override behavior as first-class reported metrics \cite{cheng2025gaia, zhang2025gridagent, ruan2024toolemu, zhang2025asb}, evaluated under long-horizon operation, tool failure, corrupted telemetry, and cascading error rather than single-shot tasks \cite{jin2025gridmind, jia2025enhancing, badmus2025powerchain, saha2025dragent, Losada2020Fault}. Until trustworthiness metrics are reported with the same rigor as task accuracy, capability will continue to outpace assurance across every domain surveyed here. We propose a standard trustworthiness scorecard, reported alongside task-success metrics in every agentic-AI evaluation, analogous to model cards for foundation models.

\subsection{Defining Autonomy Boundaries before Deployment}
\begin{figure}[h]
    \centering
    \includegraphics[trim=76.8pt 535.44pt 177.12pt 151.2pt, clip, width=1\linewidth]{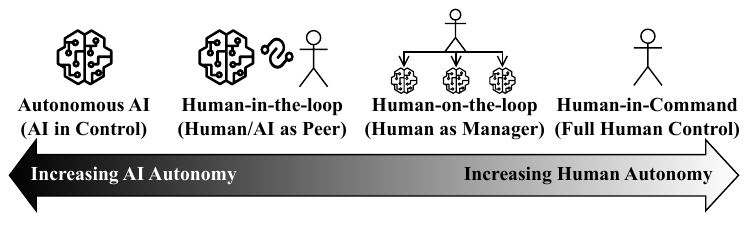}
    \caption{Human–AI autonomy spectrum, illustrating more collaborative interactions in the human-in-the-loop and human-on-the-loop setups.}
    \label{fig:autonomy_spectrum}
\end{figure}

A major open question is when an agent should advise, recommend, supervise, or act, and answering it requires domain-specific commitments rather than a generic autonomy policy. Power systems need explicit boundaries between advisory output, operator-approved action, and autonomous grid-control privilege \cite{cheng2025gaia, zhang2025gridagent}; robotics and UAVs need actuation-level safety constraints that bind before a high-level plan can move a motor \cite{ames2019control, mehmood2022black, chen2022runtime}; communication networks need policy checks between natural-language intent and executable configuration \cite{jacobs2021lumi, manias2024intent, kim2023intender}; HPC needs supervision policies for self-healing and workload redistribution \cite{Rajput2020Multi-agent, Park2005MAS, Vinay2025CoMAS-HPC}. What is missing is a shared vocabulary for expressing these boundaries so that an autonomy grant in one domain can be compared, audited, and reasoned about using the same terms as in another. The human--AI autonomy spectrum in Fig.~\ref{fig:autonomy_spectrum} -- ranging from fully autonomous AI, through human-in-the-loop and human-on-the-loop configurations, to human-in-command, offers exactly this shared reference point: each domain-specific boundary described above can be located on this spectrum rather than defined from scratch. It is important to note that this placement is not static, as some applications can introduce a dynamic autonomy level that grows with positive feedback.

\subsection{Robustness under Rare and Adversarial Conditions}
Future evaluation must deliberately target rare events, degraded sensing, corrupted inputs, adversarial prompts, and tool-chain outages rather than curated benchmarks. This means testing power-system agents against rare grid events and solver failures \cite{cheng2025gaia,jin2025gridmind,jia2025enhancing}; robotics and UAV agents against sensor and actuator degradation, jitter, packet loss, and adversarial perception \cite{ma2024controlloc,lou2024first,ma2024slowperception,eldosouky2019drones,davidovich2022towards}; HPC agents against data corruption, message loss, and processor-level failure at scale \cite{Losada2020Fault,dongarra2015fault}; and communication agents against malicious intent, semantic misalignment, and prompt injection \cite{kim2023intender,xie2021deepsc,weng2021speech,ruan2024toolemu,zhang2025asb}. A cross-domain adversarial test suite, rather than four disconnected ones, would let the field compare the amount of residual risk that survives each domain's safety filter under a common stress model. We envision this as a common outer harness, fault injection, adversarial perturbation, and telemetry corruption, with pluggable domain-specific modules.

\subsection{Provenance Standards: Build Them Before Regulators Do}
Trustworthy agents need records that connect user requests, retrieved data, intermediate reasoning, tool calls, validation results, human approval, and final actions into one recoverable chain. Power-system agents should move from tool-call traces to utility-grade audit trails \cite{badmus2025powerchain,jin2025gridmind,chen2025xgridagent,saha2025dragent}; robotics and UAV systems should maintain tamper-evident perception--decision, actuation logs for incident reconstruction \cite{samadi2023safe,samadi2024safe}; communication-network agents should connect intent parsing, semantic encoding, routing, and tool invocation into a single chain \cite{jacobs2021lumi,xie2021deepsc,weng2021speech,ruan2024toolemu,zhang2025asb}. Automotive cybersecurity and lifecycle-governance regimes such as UNECE WP.29 R155 and ISO/SAE 21434 already require exactly this kind of end-to-end traceability for vehicle software \cite{no2021155,siddiqui2023cybersecurity}; agentic AI in the other three domains has no equivalent requirement yet, despite facing a structurally identical accountability problem. Absent a proactive standard, the first cross-domain audit-trail requirement will likely arrive only after an incident forces it, at which point the field is retrofitting rather than leading.

\subsection{Securing Tool-Using and Multi-Agent Workflows}
Power-system, robotics/UAV, communication-network, and HPC agents each need conventional hardening — least-privilege tool access, typed schemas, sandboxing, network- and model-layer protection \cite{zhang2025poweragent,chen2025xgridagent,jin2025gridmind,jia2025enhancing,slavchev2026finetuned,eldosouky2019drones,davidovich2022towards,  hasan2020securing,ling2026physical, zhang2024visual,kim2023intender,ruan2024toolemu,zhang2025asb,Rajput2020Multi-agent, Park2005MAS, Sidorov2016Methods, Vinay2025CoMAS-HPC}, but this is standard software security practice applied to a new surface. The open problem is agent-to-agent trust: impersonation, poisoned shared memory, and cascading failures as a compromised or hallucinating agent silently corrupts every agent downstream of it. As multi-agent deployments grow, this shift from securing a single model to securing trust relationships between models has not yet been reflected in the field's threat models.

\subsection{Making Explanations Faithful and Decision-Useful}
Exposing a plan, a tool call, or a reasoning trace does not automatically make an explanation faithful. Power-system agents should evaluate whether solver outputs and natural-language explanations actually improve operator decisions and trust calibration, rather than assuming that visibility equals understanding \cite{jin2025gridmind,chen2025xgridagent,saha2025dragent,kiasari2026agenticsg,ghosh2025agenticee}; robotics and UAV systems should connect saliency and counterfactual explanations to accident reconstruction and pre-deployment auditing \cite{samadi2023safe,samadi2024safe}; communication-network agents should be able to explain how intent, semantic content, and routing choices became a specific network action \cite{jacobs2021lumi,manias2024intent,manias2024semanticrouting}. The open problem is measuring whether an explanation improves a human decision, not merely whether one was produced. This requires human-subject decision studies under realistic operational time pressure, not static proxies of faithfulness such as agent error correlation or attention-weight agreement.

\subsection{Toward a Reusable Cross-Domain Trustworthiness Framework}
One thesis recurs across the challenges above: agentic AI trustworthiness is not four separate problems but one problem observed from four vantage points. Evaluation, autonomy, robustness, provenance, security, and explanation are each a facet of this claim; this final subsection returns to it directly.

A reusable framework should compare domains along the axes that determine risk: the consequences of actions, the required response time, the availability of formal constraints, the density of human oversight, the maturity of audit mechanisms, and the exposure of tool, memory, communication, and actuation interfaces. On these axes, power systems represent constraint-bound, human-approved critical infrastructure; robotics and UAVs represent embodied, real-time actuation with millisecond response budgets; HPC represents large-scale resilience and recovery under weak real-time pressure; and communication networks represent intent and semantic interpretation, connected to network behavior. $\alpha^3$-Bench already demonstrates that a unified, multi-pillar benchmark is achievable within a single domain \cite{ferrag2026alpha3benchunifiedbenchmarksafety}.

Other safety-critical engineering fields have already made this transition once. Aviation software assurance and automotive functional safety converged, over decades, on graded assurance levels, tied to consequence severity, that make a certification granted in one product line legible to engineers, auditors, and regulators working on a different one \cite{IEC2022, IEC61508FunctionalSafety2022, Shinde2024, no2021155, siddiqui2023cybersecurity}. Agentic AI has not yet undergone this convergence: a validation result for a grid-dispatch agent and a validation result for a UAV planner currently share no common unit of evidence, even though both are, in essence, claims about how an autonomous reasoning process behaves at the boundary of consequential action. The EU AI Act's conformity-assessment requirements for high-risk systems point in this direction but remain domain-agnostic in a way that current agentic evaluation practice is not yet equipped to satisfy \cite{AI_Office_Pact_2024, AI_Act_Compliance_Flowchart_2025}. We expect the next phase of agentic AI research to be defined less by new architectures and more by this assurance infrastructure: graded, consequence-indexed autonomy levels; adversarial benchmark suites that generalize across domains rather than multiplying within them; and provenance formats specific enough to support an incident investigation, yet general enough to be compared across power systems, embodied robotics, HPC, and communication networks. Building shared infrastructure rather than further domain-siloed capabilities is the central open challenge this survey identifies.

\section{Conclusion}
\label{sec:conclusion}

Large language models have moved from static, single-turn prediction to agentic architectures that perceive, plan, invoke tools, and act over extended horizons, often with limited human oversight at each step. This shift changes what ``reliable'' means: task competence is no longer sufficient evidence that a system can be trusted with consequential, physically or operationally binding decisions. Trustworthiness in agentic AI must therefore be treated as an engineering property with the same rigor as latency, throughput, or accuracy, rather than a qualitative aspiration layered on afterward.

Making that argument actionable rather than rhetorical means locating safety, robustness, interpretability, accountability, and security at specific points in an agent's perception-to-action pipeline, organized around five trustworthiness dimensions and a ten-point assurance workflow. This backbone supports a unified taxonomy of agentic architectures, a catalog of concrete mechanisms and quantitative metrics that developers can apply when building or evaluating such systems, and an examination of four constraint-bound engineering domains, power systems, autonomous vehicles/robotics/UAVs, high-performance computing, and communication networks, where these mechanisms are already being tested in practice. Reading these domains through a shared lens reveals a common set of failure modes: an unverified boundary between reasoning and action, robustness evidence that rarely extends to rare or adversarial conditions, auditability that covers fragments of a workflow rather than its full length, and a security surface that grows with every tool, memory module, or coordinating agent added to the system.

Trustworthy-by-design agentic AI is the natural next phase for the field: extending this assurance framework beyond the four domains studied and validating it against real deployment incidents, not benchmark suites alone, so trust is engineered into the pipeline from the outset rather than retrofitted after failure. Aviation and automotive engineering reached comparable maturity only once graded, consequence-indexed assurance levels made trust evidence portable across products and organizations; agentic AI is at the beginning of that convergence, not the end of it.

% \section*{Acknowledgments}
% This should be a simple paragraph before the References to thank those individuals and institutions who have supported your work on this article.

\bibliographystyle{IEEEtran}
\bibliography{references}

@inproceedings{enright2022learning,
  title={A learning-based zero-trust architecture for 6g and future networks},
  author={Enright, Michael A and Hammad, Eman and Dutta, Ashutosh},
  booktitle={2022 IEEE Future Networks World Forum (FNWF)},
  pages={64--71},
  year={2022},
  organization={IEEE}
}

@inproceedings{haikal2026bridging,
  title={Bridging High-Level Intent and Network Execution: Detecting Violations and Intent Drift Through Low-Level Traffic Analysis},
  author={Haikal, Tonia and Ismail, Shereen and Hammad, Eman},
  booktitle={2026 IEEE World AI IoT Congress (AIIoT)},
  pages={0479--0485},
  year={2026},
  organization={IEEE}
}

@IEEEtranBSTCTL{IEEEexample:BSTcontrol,
  CTLuse_forced_etal       = "yes",
  CTLmax_names_forced_etal = "6",
  CTLnames_show_etal       = "1",
}

@Article{fi17090404,
AUTHOR = {Bandi, Ajay and Kongari, Bhavani and Naguru, Roshini and Pasnoor, Sahitya and Vilipala, Sri Vidya},
TITLE = {The Rise of Agentic AI: A Review of Definitions, Frameworks, Architectures, Applications, Evaluation Metrics, and Challenges},
JOURNAL = {Future Internet},
VOLUME = {17},
YEAR = {2025},
NUMBER = {9},
ARTICLE-NUMBER = {404},
URL = {https://www.mdpi.com/1999-5903/17/9/404},
ISSN = {1999-5903},
DOI = {10.3390/fi17090404}
}

@misc{yao2023reactsynergizingreasoningacting,
      title={ReAct: Synergizing Reasoning and Acting in Language Models}, 
      author={Shunyu Yao and Jeffrey Zhao and Dian Yu and Nan Du and Izhak Shafran and Karthik Narasimhan and Yuan Cao},
      year={2023},
      eprint={2210.03629},
      archivePrefix={arXiv},
      primaryClass={cs.CL},
      url={https://arxiv.org/abs/2210.03629}, 
}

@misc{shinn2023reflexionlanguageagentsverbal,
      title={Reflexion: Language Agents with Verbal Reinforcement Learning}, 
      author={Noah Shinn and Federico Cassano and Edward Berman and Ashwin Gopinath and Karthik Narasimhan and Shunyu Yao},
      year={2023},
      eprint={2303.11366},
      archivePrefix={arXiv},
      primaryClass={cs.AI},
      url={https://arxiv.org/abs/2303.11366}, 
}

@misc{yao2023treethoughtsdeliberateproblem,
      title={Tree of Thoughts: Deliberate Problem Solving with Large Language Models}, 
      author={Shunyu Yao and Dian Yu and Jeffrey Zhao and Izhak Shafran and Thomas L. Griffiths and Yuan Cao and Karthik Narasimhan},
      year={2023},
      eprint={2305.10601},
      archivePrefix={arXiv},
      primaryClass={cs.CL},
      url={https://arxiv.org/abs/2305.10601}, 
}

@INPROCEEDINGS{11103638,
  author={Raheem, Tayiba and Hossain, Gahangir},
  booktitle={2025 IEEE International Conference on Electro Information Technology (eIT)}, 
  title={Agentic AI Systems: Opportunities, Challenges, and Trustworthiness}, 
  year={2025},
  volume={},
  number={},
  pages={618-624},
  doi={10.1109/eIT64391.2025.11103638}}

@misc{nithrakashyap2025aiagents,
  author = {Arvind Nithrakashyap},
  title  = {AI agents are breaking bad and {CISO}s aren't ready},
  howpublished = {Fast Company},
  year   = {2025},
  month  = {September},
  day    = {15},
  url    = {https://www.fastcompany.com/91404298/ai-agents-are-breaking-bad-and-cisos-arent-ready},
  note   = {Impact Council}
}

@article{damle2025sleepwalking,
  author = {Siddharth Damle},
  title  = {Are we sleepwalking into an agentic {AI} crisis?},
  journal = {ABA Banking Journal},
  year   = {2025},
  month  = {December},
  day    = {15},
  url    = {https://bankingjournal.aba.com/2025/12/are-we-sleepwalking-into-an-agentic-ai-crisis/},
  note   = {American Bankers Association}
}

@article{Li2024,
  author   = {Li, Xinyi and Wang, Sai and Zeng, Siqi and Wu, Yu and Yang, Yi},
  title    = {A survey on LLM-based multi-agent systems: workflow, infrastructure, and challenges},
  journal  = {Vicinagearth},
  year     = {2024},
  month    = {Oct},
  day      = {08},
  volume   = {1},
  number   = {1},
  pages    = {9},
  issn     = {3005-060X},
  doi      = {10.1007/s44336-024-00009-2},
  url      = {https://doi.org/10.1007/s44336-024-00009-2}
}

@misc{xi2026toolgymopenworldtoolusingenvironment,
      title={ToolGym: an Open-world Tool-using Environment for Scalable Agent Testing and Data Curation}, 
      author={Ziqiao Xi and Shuang Liang and Qi Liu and Jiaqing Zhang and Letian Peng and Fang Nan and Meshal Nayim and Tianhui Zhang and Rishika Mundada and Lianhui Qin and Biwei Huang and Kun Zhou},
      year={2026},
      eprint={2601.06328},
      archivePrefix={arXiv},
      primaryClass={cs.AI},
      url={https://arxiv.org/abs/2601.06328}, 
}

@misc{schick2023toolformerlanguagemodelsteach,
      title={Toolformer: Language Models Can Teach Themselves to Use Tools}, 
      author={Timo Schick and Jane Dwivedi-Yu and Roberto Dessì and Roberta Raileanu and Maria Lomeli and Luke Zettlemoyer and Nicola Cancedda and Thomas Scialom},
      year={2023},
      eprint={2302.04761},
      archivePrefix={arXiv},
      primaryClass={cs.CL},
      url={https://arxiv.org/abs/2302.04761}, 
}

@misc{wang2023voyageropenendedembodiedagent,
      title={Voyager: An Open-Ended Embodied Agent with Large Language Models}, 
      author={Guanzhi Wang and Yuqi Xie and Yunfan Jiang and Ajay Mandlekar and Chaowei Xiao and Yuke Zhu and Linxi Fan and Anima Anandkumar},
      year={2023},
      eprint={2305.16291},
      archivePrefix={arXiv},
      primaryClass={cs.AI},
      url={https://arxiv.org/abs/2305.16291}, 
}

@misc{li2023camelcommunicativeagentsmind,
      title={CAMEL: Communicative Agents for "Mind" Exploration of Large Language Model Society}, 
      author={Guohao Li and Hasan Abed Al Kader Hammoud and Hani Itani and Dmitrii Khizbullin and Bernard Ghanem},
      year={2023},
      eprint={2303.17760},
      archivePrefix={arXiv},
      primaryClass={cs.AI},
      url={https://arxiv.org/abs/2303.17760}, 
}

@misc{wang2024qimprovingmultistepreasoning,
      title={Q*: Improving Multi-step Reasoning for LLMs with Deliberative Planning}, 
      author={Chaojie Wang and Yanchen Deng and Zhiyi Lyu and Liang Zeng and Jujie He and Shuicheng Yan and Bo An},
      year={2024},
      eprint={2406.14283},
      archivePrefix={arXiv},
      primaryClass={cs.AI},
      url={https://arxiv.org/abs/2406.14283}, 
}

@misc{aksitov2023restmeetsreactselfimprovement,
      title={ReST meets ReAct: Self-Improvement for Multi-Step Reasoning LLM Agent}, 
      author={Renat Aksitov and Sobhan Miryoosefi and Zonglin Li and Daliang Li and Sheila Babayan and Kavya Kopparapu and Zachary Fisher and Ruiqi Guo and Sushant Prakash and Pranesh Srinivasan and Manzil Zaheer and Felix Yu and Sanjiv Kumar},
      year={2023},
      eprint={2312.10003},
      archivePrefix={arXiv},
      primaryClass={cs.CL},
      url={https://arxiv.org/abs/2312.10003}, 
}

@INPROCEEDINGS{AGITrustworthiness,
  author={Raheem, Tayiba and Hossain, Gahangir},
  booktitle={2025 IEEE International Conference on Electro Information Technology (eIT)}, 
  title={Agentic AI Systems: Opportunities, Challenges, and Trustworthiness}, 
  year={2025},
  volume={},
  number={},
  pages={618-624},
  doi={10.1109/eIT64391.2025.11103638}}

@inproceedings{ConstPolicyOptimization,
author = {Achiam, Joshua and Held, David and Tamar, Aviv and Abbeel, Pieter},
title = {Constrained policy optimization},
year = {2017},
publisher = {JMLR.org},
booktitle = {Proceedings of the 34th International Conference on Machine Learning - Volume 70},
pages = {22–31},
numpages = {10},
location = {Sydney, NSW, Australia},
series = {ICML'17}
}

@inproceedings{madry2018towards,
title={Towards Deep Learning Models Resistant to Adversarial Attacks},
author={Aleksander Madry and Aleksandar Makelov and Ludwig Schmidt and Dimitris Tsipras and Adrian Vladu},
booktitle={International Conference on Learning Representations},
year={2018},
url={https://openreview.net/forum?id=rJzIBfZAb},
}

@article{XAIACM,
author = {Dwivedi, Rudresh and Dave, Devam and Naik, Het and Singhal, Smiti and Omer, Rana and Patel, Pankesh and Qian, Bin and Wen, Zhenyu and Shah, Tejal and Morgan, Graham and Ranjan, Rajiv},
title = {Explainable AI (XAI): Core Ideas, Techniques, and Solutions},
year = {2023},
issue_date = {September 2023},
publisher = {Association for Computing Machinery},
address = {New York, NY, USA},
volume = {55},
number = {9},
issn = {0360-0300},
url = {https://doi.org/10.1145/3561048},
doi = {10.1145/3561048},
journal = {ACM Comput. Surv.},
month = jan,
articleno = {194},
numpages = {33}
}

@inproceedings{AiAccountabilityACM,
author = {Raji, Inioluwa Deborah and Smart, Andrew and White, Rebecca N. and Mitchell, Margaret and Gebru, Timnit and Hutchinson, Ben and Smith-Loud, Jamila and Theron, Daniel and Barnes, Parker},
title = {Closing the AI accountability gap: defining an end-to-end framework for internal algorithmic auditing},
year = {2020},
isbn = {9781450369367},
publisher = {Association for Computing Machinery},
address = {New York, NY, USA},
url = {https://doi.org/10.1145/3351095.3372873},
doi = {10.1145/3351095.3372873},
booktitle = {Proceedings of the 2020 Conference on Fairness, Accountability, and Transparency},
pages = {33–44},
numpages = {12},
location = {Barcelona, Spain},
series = {FAT* '20}
}

@inproceedings{PrivacyAI,
author = {Abadi, Martin and Chu, Andy and Goodfellow, Ian and McMahan, H. Brendan and Mironov, Ilya and Talwar, Kunal and Zhang, Li},
title = {Deep Learning with Differential Privacy},
year = {2016},
isbn = {9781450341394},
publisher = {Association for Computing Machinery},
address = {New York, NY, USA},
url = {https://doi.org/10.1145/2976749.2978318},
doi = {10.1145/2976749.2978318},
booktitle = {Proceedings of the 2016 ACM SIGSAC Conference on Computer and Communications Security},
pages = {308–318},
numpages = {11},
location = {Vienna, Austria},
series = {CCS '16}
}

@ARTICLE{AgenticAIAutonomousSurvey,
  author={Acharya, Deepak Bhaskar and Kuppan, Karthigeyan and Divya, B.},
  journal={IEEE Access}, 
  title={Agentic AI: Autonomous Intelligence for Complex Goals—A Comprehensive Survey}, 
  year={2025},
  volume={13},
  number={},
  pages={18912-18936},
  doi={10.1109/ACCESS.2025.3532853}}

@article{AbouAli2025,
  author   = {Abou Ali, Mohamad and Dornaika, Fadi and Charafeddine, Jinan},
  title    = {Agentic {AI}: a comprehensive survey of architectures, applications, and future directions},
  journal  = {Artificial Intelligence Review},
  year     = {2025},
  volume   = {59},
  number   = {1},
  pages    = {11},
  month    = nov,
  issn     = {1573-7462},
  doi      = {10.1007/s10462-025-11422-4},
  url      = {https://doi.org/10.1007/s10462-025-11422-4},
}

@article{Cho2025,
  author  = {Cho, S.-H. and Lee, Y.-S.},
  title   = {A Comparative Study of Modern {AI} Frameworks Based on Architecture, Integration, and Scalability},
  journal = {International Journal of Advanced Smart Convergence},
  year    = {2025},
  volume  = {14},
  number  = {4},
  pages   = {158--167},
  month   = dec
}

@article{mavroudis2024langchain,
    doi = {10.20944/preprints202411.0566.v1},
	url = {https://doi.org/10.20944/preprints202411.0566.v1},
	year = 2024,
	month = {November},
	publisher = {Preprints},
	author = {Vasilios Mavroudis},
	title = {LangChain v0.3},
	journal = {Preprints}
}

@misc{hendrycks2023overviewcatastrophicairisks,
      title={An Overview of Catastrophic AI Risks}, 
      author={Dan Hendrycks and Mantas Mazeika and Thomas Woodside},
      year={2023},
      eprint={2306.12001},
      archivePrefix={arXiv},
      primaryClass={cs.CY},
      url={https://arxiv.org/abs/2306.12001}, 
}

@misc{chiu2025morebenchevaluatingproceduralpluralistic,
      title={MoReBench: Evaluating Procedural and Pluralistic Moral Reasoning in Language Models, More than Outcomes}, 
      author={Yu Ying Chiu and Michael S. Lee and Rachel Calcott and Brandon Handoko and Paul de Font-Reaulx and Paula Rodriguez and Chen Bo Calvin Zhang and Ziwen Han and Udari Madhushani Sehwag and Yash Maurya and Christina Q Knight and Harry R. Lloyd and Florence Bacus and Mantas Mazeika and Bing Liu and Yejin Choi and Mitchell L Gordon and Sydney Levine},
      year={2025},
      eprint={2510.16380},
      archivePrefix={arXiv},
      primaryClass={cs.CL},
      url={https://arxiv.org/abs/2510.16380}, 
}

@misc{maini2025safetypretraininggenerationsafe,
      title={Safety Pretraining: Toward the Next Generation of Safe AI}, 
      author={Pratyush Maini and Sachin Goyal and Dylan Sam and Alex Robey and Yash Savani and Yiding Jiang and Andy Zou and Matt Fredrikson and Zacharcy C. Lipton and J. Zico Kolter},
      year={2025},
      eprint={2504.16980},
      archivePrefix={arXiv},
      primaryClass={cs.LG},
      url={https://arxiv.org/abs/2504.16980}, 
}

@misc{ren2026maskbenchmarkdisentanglinghonesty,
      title={The MASK Benchmark: Disentangling Honesty From Accuracy in AI Systems}, 
      author={Richard Ren and Arunim Agarwal and Mantas Mazeika and Cristina Menghini and Robert Vacareanu and Brad Kenstler and Mick Yang and Isabelle Barrass and Alice Gatti and Xuwang Yin and Eduardo Trevino and Matias Geralnik and Adam Khoja and Dean Lee and Summer Yue and Dan Hendrycks},
      year={2026},
      eprint={2503.03750},
      archivePrefix={arXiv},
      primaryClass={cs.LG},
      url={https://arxiv.org/abs/2503.03750}, 
}

@misc{mazeika2025utilityengineeringanalyzingcontrolling,
      title={Utility Engineering: Analyzing and Controlling Emergent Value Systems in AIs}, 
      author={Mantas Mazeika and Xuwang Yin and Rishub Tamirisa and Jaehyuk Lim and Bruce W. Lee and Richard Ren and Long Phan and Norman Mu and Adam Khoja and Oliver Zhang and Dan Hendrycks},
      year={2025},
      eprint={2502.08640},
      archivePrefix={arXiv},
      primaryClass={cs.LG},
      url={https://arxiv.org/abs/2502.08640}, 
}

@misc{phan2025humanitysexam,
      title={Humanity's Last Exam}, 
      author={Long Phan and Alice Gatti and Ziwen Han and Nathaniel Li and Josephina Hu and Hugh Zhang and Chen Bo Calvin Zhang and Mohamed Shaaban and John Ling and Sean Shi and Michael Choi and Anish Agrawal and Arnav Chopra and Adam Khoja and Ryan Kim and Richard Ren and Jason Hausenloy and Oliver Zhang and Mantas Mazeika and Dmitry Dodonov and Tung Nguyen and Jaeho Lee and Daron Anderson and Mikhail Doroshenko and Alun Cennyth Stokes and Mobeen Mahmood and Oleksandr Pokutnyi and Oleg Iskra and Jessica P. Wang and John-Clark Levin and Mstyslav Kazakov and Fiona Feng and Steven Y. Feng and Haoran Zhao and Michael Yu and Varun Gangal and Chelsea Zou and Zihan Wang and Serguei Popov and Robert Gerbicz and Geoff Galgon and Johannes Schmitt and Will Yeadon and Yongki Lee and Scott Sauers and Alvaro Sanchez and Fabian Giska and Marc Roth and Søren Riis and Saiteja Utpala and Noah Burns and Gashaw M. Goshu and Mohinder Maheshbhai Naiya and Chidozie Agu and Zachary Giboney and Antrell Cheatom and Francesco Fournier-Facio and Sarah-Jane Crowson and Lennart Finke and Zerui Cheng and Jennifer Zampese and Ryan G. Hoerr and Mark Nandor and Hyunwoo Park and Tim Gehrunger and Jiaqi Cai and Ben McCarty and Alexis C Garretson and Edwin Taylor and Damien Sileo and Qiuyu Ren and Usman Qazi and Lianghui Li and Jungbae Nam and John B. Wydallis and Pavel Arkhipov and Jack Wei Lun Shi and Aras Bacho and Chris G. Willcocks and Hangrui Cao and Sumeet Motwani and Emily de Oliveira Santos and Johannes Veith and Edward Vendrow and Doru Cojoc and Kengo Zenitani and Joshua Robinson and Longke Tang and Yuqi Li and Joshua Vendrow and Natanael Wildner Fraga and Vladyslav Kuchkin and Andrey Pupasov Maksimov and Pierre Marion and Denis Efremov and Jayson Lynch and Kaiqu Liang and Aleksandar Mikov and Andrew Gritsevskiy and Julien Guillod and Gözdenur Demir and Dakotah Martinez and Ben Pageler and Kevin Zhou and Saeed Soori and Ori Press and Henry Tang and Paolo Rissone and Sean R. Green and Lina Brüssel and Moon Twayana and Aymeric Dieuleveut and Joseph Marvin Imperial and Ameya Prabhu and Jinzhou Yang and Nick Crispino and Arun Rao and Dimitri Zvonkine and Gabriel Loiseau and Mikhail Kalinin and Marco Lukas and Ciprian Manolescu and Nate Stambaugh and Subrata Mishra and Tad Hogg and Carlo Bosio and Brian P Coppola and Julian Salazar and Jaehyeok Jin and Rafael Sayous and Stefan Ivanov and Philippe Schwaller and Shaipranesh Senthilkuma and Andres M Bran and Andres Algaba and Kelsey Van den Houte and Lynn Van Der Sypt and Brecht Verbeken and David Noever and Alexei Kopylov and Benjamin Myklebust and Bikun Li and Lisa Schut and Evgenii Zheltonozhskii and Qiaochu Yuan and Derek Lim and Richard Stanley and Tong Yang and John Maar and Julian Wykowski and Martí Oller and Anmol Sahu and Cesare Giulio Ardito and Yuzheng Hu and Ariel Ghislain Kemogne Kamdoum and Alvin Jin and Tobias Garcia Vilchis and Yuexuan Zu and Martin Lackner and James Koppel and Gongbo Sun and Daniil S. Antonenko and Steffi Chern and Bingchen Zhao and Pierrot Arsene and Joseph M Cavanagh and Daofeng Li and Jiawei Shen and Donato Crisostomi and Wenjin Zhang and Ali Dehghan and Sergey Ivanov and David Perrella and Nurdin Kaparov and Allen Zang and Ilia Sucholutsky and Arina Kharlamova and Daniil Orel and Vladislav Poritski and Shalev Ben-David and Zachary Berger and Parker Whitfill and Michael Foster and Daniel Munro and Linh Ho and Shankar Sivarajan and Dan Bar Hava and Aleksey Kuchkin and David Holmes and Alexandra Rodriguez-Romero and Frank Sommerhage and Anji Zhang and Richard Moat and Keith Schneider and Zakayo Kazibwe and Don Clarke and Dae Hyun Kim and Felipe Meneguitti Dias and Sara Fish and Veit Elser and Tobias Kreiman and Victor Efren Guadarrama Vilchis and Immo Klose and Ujjwala Anantheswaran and Adam Zweiger and Kaivalya Rawal and Jeffery Li and Jeremy Nguyen and Nicolas Daans and Haline Heidinger and Maksim Radionov and Václav Rozhoň and Vincent Ginis and Christian Stump and Niv Cohen and Rafał Poświata and Josef Tkadlec and Alan Goldfarb and Chenguang Wang and Piotr Padlewski and Stanislaw Barzowski and Kyle Montgomery and Ryan Stendall and Jamie Tucker-Foltz and Jack Stade and T. Ryan Rogers and Tom Goertzen and Declan Grabb and Abhishek Shukla and Alan Givré and John Arnold Ambay and Archan Sen and Muhammad Fayez Aziz and Mark H Inlow and Hao He and Ling Zhang and Younesse Kaddar and Ivar Ängquist and Yanxu Chen and Harrison K Wang and Kalyan Ramakrishnan and Elliott Thornley and Antonio Terpin and Hailey Schoelkopf and Eric Zheng and Avishy Carmi and Ethan D. L. Brown and Kelin Zhu and Max Bartolo and Richard Wheeler and Martin Stehberger and Peter Bradshaw and JP Heimonen and Kaustubh Sridhar and Ido Akov and Jennifer Sandlin and Yury Makarychev and Joanna Tam and Hieu Hoang and David M. Cunningham and Vladimir Goryachev and Demosthenes Patramanis and Michael Krause and Andrew Redenti and David Aldous and Jesyin Lai and Shannon Coleman and Jiangnan Xu and Sangwon Lee and Ilias Magoulas and Sandy Zhao and Ning Tang and Michael K. Cohen and Orr Paradise and Jan Hendrik Kirchner and Maksym Ovchynnikov and Jason O. Matos and Adithya Shenoy and Michael Wang and Yuzhou Nie and Anna Sztyber-Betley and Paolo Faraboschi and Robin Riblet and Jonathan Crozier and Shiv Halasyamani and Shreyas Verma and Prashant Joshi and Eli Meril and Ziqiao Ma and Jérémy Andréoletti and Raghav Singhal and Jacob Platnick and Volodymyr Nevirkovets and Luke Basler and Alexander Ivanov and Seri Khoury and Nils Gustafsson and Marco Piccardo and Hamid Mostaghimi and Qijia Chen and Virendra Singh and Tran Quoc Khánh and Paul Rosu and Hannah Szlyk and Zachary Brown and Himanshu Narayan and Aline Menezes and Jonathan Roberts and William Alley and Kunyang Sun and Arkil Patel and Max Lamparth and Anka Reuel and Linwei Xin and Hanmeng Xu and Jacob Loader and Freddie Martin and Zixuan Wang and Andrea Achilleos and Thomas Preu and Tomek Korbak and Ida Bosio and Fereshteh Kazemi and Ziye Chen and Biró Bálint and Eve J. Y. Lo and Jiaqi Wang and Maria Inês S. Nunes and Jeremiah Milbauer and M Saiful Bari and Zihao Wang and Behzad Ansarinejad and Yewen Sun and Stephane Durand and Hossam Elgnainy and Guillaume Douville and Daniel Tordera and George Balabanian and Hew Wolff and Lynna Kvistad and Hsiaoyun Milliron and Ahmad Sakor and Murat Eron and Andrew Favre D. O. and Shailesh Shah and Xiaoxiang Zhou and Firuz Kamalov and Sherwin Abdoli and Tim Santens and Shaul Barkan and Allison Tee and Robin Zhang and Alessandro Tomasiello and G. Bruno De Luca and Shi-Zhuo Looi and Vinh-Kha Le and Noam Kolt and Jiayi Pan and Emma Rodman and Jacob Drori and Carl J Fossum and Niklas Muennighoff and Milind Jagota and Ronak Pradeep and Honglu Fan and Jonathan Eicher and Michael Chen and Kushal Thaman and William Merrill and Moritz Firsching and Carter Harris and Stefan Ciobâcă and Jason Gross and Rohan Pandey and Ilya Gusev and Adam Jones and Shashank Agnihotri and Pavel Zhelnov and Mohammadreza Mofayezi and Alexander Piperski and David K. Zhang and Kostiantyn Dobarskyi and Roman Leventov and Ignat Soroko and Joshua Duersch and Vage Taamazyan and Andrew Ho and Wenjie Ma and William Held and Ruicheng Xian and Armel Randy Zebaze and Mohanad Mohamed and Julian Noah Leser and Michelle X Yuan and Laila Yacar and Johannes Lengler and Katarzyna Olszewska and Claudio Di Fratta and Edson Oliveira and Joseph W. Jackson and Andy Zou and Muthu Chidambaram and Timothy Manik and Hector Haffenden and Dashiell Stander and Ali Dasouqi and Alexander Shen and Bita Golshani and David Stap and Egor Kretov and Mikalai Uzhou and Alina Borisovna Zhidkovskaya and Nick Winter and Miguel Orbegozo Rodriguez and Robert Lauff and Dustin Wehr and Colin Tang and Zaki Hossain and Shaun Phillips and Fortuna Samuele and Fredrik Ekström and Angela Hammon and Oam Patel and Faraz Farhidi and George Medley and Forough Mohammadzadeh and Madellene Peñaflor and Haile Kassahun and Alena Friedrich and Rayner Hernandez Perez and Daniel Pyda and Taom Sakal and Omkar Dhamane and Ali Khajegili Mirabadi and Eric Hallman and Kenchi Okutsu and Mike Battaglia and Mohammad Maghsoudimehrabani and Alon Amit and Dave Hulbert and Roberto Pereira and Simon Weber and Handoko and Anton Peristyy and Stephen Malina and Mustafa Mehkary and Rami Aly and Frank Reidegeld and Anna-Katharina Dick and Cary Friday and Mukhwinder Singh and Hassan Shapourian and Wanyoung Kim and Mariana Costa and Hubeyb Gurdogan and Harsh Kumar and Chiara Ceconello and Chao Zhuang and Haon Park and Micah Carroll and Andrew R. Tawfeek and Stefan Steinerberger and Daattavya Aggarwal and Michael Kirchhof and Linjie Dai and Evan Kim and Johan Ferret and Jainam Shah and Yuzhou Wang and Minghao Yan and Krzysztof Burdzy and Lixin Zhang and Antonio Franca and Diana T. Pham and Kang Yong Loh and Joshua Robinson and Abram Jackson and Paolo Giordano and Philipp Petersen and Adrian Cosma and Jesus Colino and Colin White and Jacob Votava and Vladimir Vinnikov and Ethan Delaney and Petr Spelda and Vit Stritecky and Syed M. Shahid and Jean-Christophe Mourrat and Lavr Vetoshkin and Koen Sponselee and Renas Bacho and Zheng-Xin Yong and Florencia de la Rosa and Nathan Cho and Xiuyu Li and Guillaume Malod and Orion Weller and Guglielmo Albani and Leon Lang and Julien Laurendeau and Dmitry Kazakov and Fatimah Adesanya and Julien Portier and Lawrence Hollom and Victor Souza and Yuchen Anna Zhou and Julien Degorre and Yiğit Yalın and Gbenga Daniel Obikoya and Rai and Filippo Bigi and M. C. Boscá and Oleg Shumar and Kaniuar Bacho and Gabriel Recchia and Mara Popescu and Nikita Shulga and Ngefor Mildred Tanwie and Thomas C. H. Lux and Ben Rank and Colin Ni and Matthew Brooks and Alesia Yakimchyk and Huanxu and Liu and Stefano Cavalleri and Olle Häggström and Emil Verkama and Joshua Newbould and Hans Gundlach and Leonor Brito-Santana and Brian Amaro and Vivek Vajipey and Rynaa Grover and Ting Wang and Yosi Kratish and Wen-Ding Li and Sivakanth Gopi and Andrea Caciolai and Christian Schroeder de Witt and Pablo Hernández-Cámara and Emanuele Rodolà and Jules Robins and Dominic Williamson and Vincent Cheng and Brad Raynor and Hao Qi and Ben Segev and Jingxuan Fan and Sarah Martinson and Erik Y. Wang and Kaylie Hausknecht and Michael P. Brenner and Mao Mao and Christoph Demian and Peyman Kassani and Xinyu Zhang and David Avagian and Eshawn Jessica Scipio and Alon Ragoler and Justin Tan and Blake Sims and Rebeka Plecnik and Aaron Kirtland and Omer Faruk Bodur and D. P. Shinde and Yan Carlos Leyva Labrador and Zahra Adoul and Mohamed Zekry and Ali Karakoc and Tania C. B. Santos and Samir Shamseldeen and Loukmane Karim and Anna Liakhovitskaia and Nate Resman and Nicholas Farina and Juan Carlos Gonzalez and Gabe Maayan and Earth Anderson and Rodrigo De Oliveira Pena and Elizabeth Kelley and Hodjat Mariji and Rasoul Pouriamanesh and Wentao Wu and Ross Finocchio and Ismail Alarab and Joshua Cole and Danyelle Ferreira and Bryan Johnson and Mohammad Safdari and Liangti Dai and Siriphan Arthornthurasuk and Isaac C. McAlister and Alejandro José Moyano and Alexey Pronin and Jing Fan and Angel Ramirez-Trinidad and Yana Malysheva and Daphiny Pottmaier and Omid Taheri and Stanley Stepanic and Samuel Perry and Luke Askew and Raúl Adrián Huerta Rodríguez and Ali M. R. Minissi and Ricardo Lorena and Krishnamurthy Iyer and Arshad Anil Fasiludeen and Ronald Clark and Josh Ducey and Matheus Piza and Maja Somrak and Eric Vergo and Juehang Qin and Benjámin Borbás and Eric Chu and Jack Lindsey and Antoine Jallon and I. M. J. McInnis and Evan Chen and Avi Semler and Luk Gloor and Tej Shah and Marc Carauleanu and Pascal Lauer and Tran Đuc Huy and Hossein Shahrtash and Emilien Duc and Lukas Lewark and Assaf Brown and Samuel Albanie and Brian Weber and Warren S. Vaz and Pierre Clavier and Yiyang Fan and Gabriel Poesia Reis e Silva and Long and Lian and Marcus Abramovitch and Xi Jiang and Sandra Mendoza and Murat Islam and Juan Gonzalez and Vasilios Mavroudis and Justin Xu and Pawan Kumar and Laxman Prasad Goswami and Daniel Bugas and Nasser Heydari and Ferenc Jeanplong and Thorben Jansen and Antonella Pinto and Archimedes Apronti and Abdallah Galal and Ng Ze-An and Ankit Singh and Tong Jiang and Joan of Arc Xavier and Kanu Priya Agarwal and Mohammed Berkani and Gang Zhang and Zhehang Du and Benedito Alves de Oliveira Junior and Dmitry Malishev and Nicolas Remy and Taylor D. Hartman and Tim Tarver and Stephen Mensah and Gautier Abou Loume and Wiktor Morak and Farzad Habibi and Sarah Hoback and Will Cai and Javier Gimenez and Roselynn Grace Montecillo and Jakub Łucki and Russell Campbell and Asankhaya Sharma and Khalida Meer and Shreen Gul and Daniel Espinosa Gonzalez and Xavier Alapont and Alex Hoover and Gunjan Chhablani and Freddie Vargus and Arunim Agarwal and Yibo Jiang and Deepakkumar Patil and David Outevsky and Kevin Joseph Scaria and Rajat Maheshwari and Abdelkader Dendane and Priti Shukla and Ashley Cartwright and Sergei Bogdanov and Niels Mündler and Sören Möller and Luca Arnaboldi and Kunvar Thaman and Muhammad Rehan Siddiqi and Prajvi Saxena and Himanshu Gupta and Tony Fruhauff and Glen Sherman and Mátyás Vincze and Siranut Usawasutsakorn and Dylan Ler and Anil Radhakrishnan and Innocent Enyekwe and Sk Md Salauddin and Jiang Muzhen and Aleksandr Maksapetyan and Vivien Rossbach and Chris Harjadi and Mohsen Bahaloohoreh and Claire Sparrow and Jasdeep Sidhu and Sam Ali and Song Bian and John Lai and Eric Singer and Justine Leon Uro and Greg Bateman and Mohamed Sayed and Ahmed Menshawy and Darling Duclosel and Dario Bezzi and Yashaswini Jain and Ashley Aaron and Murat Tiryakioglu and Sheeshram Siddh and Keith Krenek and Imad Ali Shah and Jun Jin and Scott Creighton and Denis Peskoff and Zienab EL-Wasif and Ragavendran P V and Michael Richmond and Joseph McGowan and Tejal Patwardhan and Hao-Yu Sun and Ting Sun and Nikola Zubić and Samuele Sala and Stephen Ebert and Jean Kaddour and Manuel Schottdorf and Dianzhuo Wang and Gerol Petruzella and Alex Meiburg and Tilen Medved and Ali ElSheikh and S Ashwin Hebbar and Lorenzo Vaquero and Xianjun Yang and Jason Poulos and Vilém Zouhar and Sergey Bogdanik and Mingfang Zhang and Jorge Sanz-Ros and David Anugraha and Yinwei Dai and Anh N. Nhu and Xue Wang and Ali Anil Demircali and Zhibai Jia and Yuyin Zhou and Juncheng Wu and Mike He and Nitin Chandok and Aarush Sinha and Gaoxiang Luo and Long Le and Mickaël Noyé and Michał Perełkiewicz and Ioannis Pantidis and Tianbo Qi and Soham Sachin Purohit and Letitia Parcalabescu and Thai-Hoa Nguyen and Genta Indra Winata and Edoardo M. Ponti and Hanchen Li and Kaustubh Dhole and Jongee Park and Dario Abbondanza and Yuanli Wang and Anupam Nayak and Diogo M. Caetano and Antonio A. W. L. Wong and Maria del Rio-Chanona and Dániel Kondor and Pieter Francois and Ed Chalstrey and Jakob Zsambok and Dan Hoyer and Jenny Reddish and Jakob Hauser and Francisco-Javier Rodrigo-Ginés and Suchandra Datta and Maxwell Shepherd and Thom Kamphuis and Qizheng Zhang and Hyunjun Kim and Ruiji Sun and Jianzhu Yao and Franck Dernoncourt and Satyapriya Krishna and Sina Rismanchian and Bonan Pu and Francesco Pinto and Yingheng Wang and Kumar Shridhar and Kalon J. Overholt and Glib Briia and Hieu Nguyen and David and Soler Bartomeu and Tony CY Pang and Adam Wecker and Yifan Xiong and Fanfei Li and Lukas S. Huber and Joshua Jaeger and Romano De Maddalena and Xing Han Lù and Yuhui Zhang and Claas Beger and Patrick Tser Jern Kon and Sean Li and Vivek Sanker and Ming Yin and Yihao Liang and Xinlu Zhang and Ankit Agrawal and Li S. Yifei and Zechen Zhang and Mu Cai and Yasin Sonmez and Costin Cozianu and Changhao Li and Alex Slen and Shoubin Yu and Hyun Kyu Park and Gabriele Sarti and Marcin Briański and Alessandro Stolfo and Truong An Nguyen and Mike Zhang and Yotam Perlitz and Jose Hernandez-Orallo and Runjia Li and Amin Shabani and Felix Juefei-Xu and Shikhar Dhingra and Orr Zohar and My Chiffon Nguyen and Alexander Pondaven and Abdurrahim Yilmaz and Xuandong Zhao and Chuanyang Jin and Muyan Jiang and Stefan Todoran and Xinyao Han and Jules Kreuer and Brian Rabern and Anna Plassart and Martino Maggetti and Luther Yap and Robert Geirhos and Jonathon Kean and Dingsu Wang and Sina Mollaei and Chenkai Sun and Yifan Yin and Shiqi Wang and Rui Li and Yaowen Chang and Anjiang Wei and Alice Bizeul and Xiaohan Wang and Alexandre Oliveira Arrais and Kushin Mukherjee and Jorge Chamorro-Padial and Jiachen Liu and Xingyu Qu and Junyi Guan and Adam Bouyamourn and Shuyu Wu and Martyna Plomecka and Junda Chen and Mengze Tang and Jiaqi Deng and Shreyas Subramanian and Haocheng Xi and Haoxuan Chen and Weizhi Zhang and Yinuo Ren and Haoqin Tu and Sejong Kim and Yushun Chen and Sara Vera Marjanović and Junwoo Ha and Grzegorz Luczyna and Jeff J. Ma and Zewen Shen and Dawn Song and Cedegao E. Zhang and Zhun Wang and Gaël Gendron and Yunze Xiao and Leo Smucker and Erica Weng and Kwok Hao Lee and Zhe Ye and Stefano Ermon and Ignacio D. Lopez-Miguel and Theo Knights and Anthony Gitter and Namkyu Park and Boyi Wei and Hongzheng Chen and Kunal Pai and Ahmed Elkhanany and Han Lin and Philipp D. Siedler and Jichao Fang and Ritwik Mishra and Károly Zsolnai-Fehér and Xilin Jiang and Shadab Khan and Jun Yuan and Rishab Kumar Jain and Xi Lin and Mike Peterson and Zhe Wang and Aditya Malusare and Maosen Tang and Isha Gupta and Ivan Fosin and Timothy Kang and Barbara Dworakowska and Kazuki Matsumoto and Guangyao Zheng and Gerben Sewuster and Jorge Pretel Villanueva and Ivan Rannev and Igor Chernyavsky and Jiale Chen and Deepayan Banik and Ben Racz and Wenchao Dong and Jianxin Wang and Laila Bashmal and Duarte V. Gonçalves and Wei Hu and Kaushik Bar and Ondrej Bohdal and Atharv Singh Patlan and Shehzaad Dhuliawala and Caroline Geirhos and Julien Wist and Yuval Kansal and Bingsen Chen and Kutay Tire and Atak Talay Yücel and Brandon Christof and Veerupaksh Singla and Zijian Song and Sanxing Chen and Jiaxin Ge and Kaustubh Ponkshe and Isaac Park and Tianneng Shi and Martin Q. Ma and Joshua Mak and Sherwin Lai and Antoine Moulin and Zhuo Cheng and Zhanda Zhu and Ziyi Zhang and Vaidehi Patil and Ketan Jha and Qiutong Men and Jiaxuan Wu and Tianchi Zhang and Bruno Hebling Vieira and Alham Fikri Aji and Jae-Won Chung and Mohammed Mahfoud and Ha Thi Hoang and Marc Sperzel and Wei Hao and Kristof Meding and Sihan Xu and Vassilis Kostakos and Davide Manini and Yueying Liu and Christopher Toukmaji and Jay Paek and Eunmi Yu and Arif Engin Demircali and Zhiyi Sun and Ivan Dewerpe and Hongsen Qin and Roman Pflugfelder and James Bailey and Johnathan Morris and Ville Heilala and Sybille Rosset and Zishun Yu and Peter E. Chen and Woongyeong Yeo and Eeshaan Jain and Ryan Yang and Sreekar Chigurupati and Julia Chernyavsky and Sai Prajwal Reddy and Subhashini Venugopalan and Hunar Batra and Core Francisco Park and Hieu Tran and Guilherme Maximiano and Genghan Zhang and Yizhuo Liang and Hu Shiyu and Rongwu Xu and Rui Pan and Siddharth Suresh and Ziqi Liu and Samaksh Gulati and Songyang Zhang and Peter Turchin and Christopher W. Bartlett and Christopher R. Scotese and Phuong M. Cao and Ben Wu and Jacek Karwowski and Davide Scaramuzza and Aakaash Nattanmai and Gordon McKellips and Anish Cheraku and Asim Suhail and Ethan Luo and Marvin Deng and Jason Luo and Ashley Zhang and Kavin Jindel and Jay Paek and Kasper Halevy and Allen Baranov and Michael Liu and Advaith Avadhanam and David Zhang and Vincent Cheng and Brad Ma and Evan Fu and Liam Do and Joshua Lass and Hubert Yang and Surya Sunkari and Vishruth Bharath and Violet Ai and James Leung and Rishit Agrawal and Alan Zhou and Kevin Chen and Tejas Kalpathi and Ziqi Xu and Gavin Wang and Tyler Xiao and Erik Maung and Sam Lee and Ryan Yang and Roy Yue and Ben Zhao and Julia Yoon and Sunny Sun and Aryan Singh and Ethan Luo and Clark Peng and Tyler Osbey and Taozhi Wang and Daryl Echeazu and Hubert Yang and Timothy Wu and Spandan Patel and Vidhi Kulkarni and Vijaykaarti Sundarapandiyan and Ashley Zhang and Andrew Le and Zafir Nasim and Srikar Yalam and Ritesh Kasamsetty and Soham Samal and Hubert Yang and David Sun and Nihar Shah and Abhijeet Saha and Alex Zhang and Leon Nguyen and Laasya Nagumalli and Kaixin Wang and Alan Zhou and Aidan Wu and Jason Luo and Anwith Telluri and Summer Yue and Alexandr Wang and Dan Hendrycks},
      year={2025},
      eprint={2501.14249},
      archivePrefix={arXiv},
      primaryClass={cs.LG},
      url={https://arxiv.org/abs/2501.14249}, 
}

@misc{andriushchenko2025agentharmbenchmarkmeasuringharmfulness,
      title={AgentHarm: A Benchmark for Measuring Harmfulness of LLM Agents}, 
      author={Maksym Andriushchenko and Alexandra Souly and Mateusz Dziemian and Derek Duenas and Maxwell Lin and Justin Wang and Dan Hendrycks and Andy Zou and Zico Kolter and Matt Fredrikson and Eric Winsor and Jerome Wynne and Yarin Gal and Xander Davies},
      year={2025},
      eprint={2410.09024},
      archivePrefix={arXiv},
      primaryClass={cs.LG},
      url={https://arxiv.org/abs/2410.09024}, 
}

@misc{tamirisa2024tamperresistantsafeguardsopenweightllms,
title={Tamper-Resistant Safeguards for Open-Weight LLMs},
author={Rishub Tamirisa and Bhrugu Bharathi and Long Phan and Andy Zhou and Alice Gatti and Tarun Suresh and Maxwell Lin and Justin Wang and Rowan Wang and Ron Arel and Andy Zou and Dawn Song and Bo Li and Dan Hendrycks and Mantas Mazeika},
year={2024},
eprint={2408.00761},
archivePrefix={arXiv},
primaryClass={cs.LG},
url={https://arxiv.org/abs/2408.00761}
}

@misc{zou2024improvingalignmentrobustnesscircuit,
      title={Improving Alignment and Robustness with Circuit Breakers}, 
      author={Andy Zou and Long Phan and Justin Wang and Derek Duenas and Maxwell Lin and Maksym Andriushchenko and Rowan Wang and Zico Kolter and Matt Fredrikson and Dan Hendrycks},
      year={2024},
      eprint={2406.04313},
      archivePrefix={arXiv},
      primaryClass={cs.LG},
      url={https://arxiv.org/abs/2406.04313}, 
}

@misc{li2024wmdpbenchmarkmeasuringreducing,
      title={The WMDP Benchmark: Measuring and Reducing Malicious Use With Unlearning}, 
      author={Nathaniel Li and Alexander Pan and Anjali Gopal and Summer Yue and Daniel Berrios and Alice Gatti and Justin D. Li and Ann-Kathrin Dombrowski and Shashwat Goel and Long Phan and Gabriel Mukobi and Nathan Helm-Burger and Rassin Lababidi and Lennart Justen and Andrew B. Liu and Michael Chen and Isabelle Barrass and Oliver Zhang and Xiaoyuan Zhu and Rishub Tamirisa and Bhrugu Bharathi and Adam Khoja and Zhenqi Zhao and Ariel Herbert-Voss and Cort B. Breuer and Samuel Marks and Oam Patel and Andy Zou and Mantas Mazeika and Zifan Wang and Palash Oswal and Weiran Lin and Adam A. Hunt and Justin Tienken-Harder and Kevin Y. Shih and Kemper Talley and John Guan and Russell Kaplan and Ian Steneker and David Campbell and Brad Jokubaitis and Alex Levinson and Jean Wang and William Qian and Kallol Krishna Karmakar and Steven Basart and Stephen Fitz and Mindy Levine and Ponnurangam Kumaraguru and Uday Tupakula and Vijay Varadharajan and Ruoyu Wang and Yan Shoshitaishvili and Jimmy Ba and Kevin M. Esvelt and Alexandr Wang and Dan Hendrycks},
      year={2024},
      eprint={2403.03218},
      archivePrefix={arXiv},
      primaryClass={cs.LG},
      url={https://arxiv.org/abs/2403.03218}, 
}

@misc{mazeika2024harmbenchstandardizedevaluationframework,
      title={HarmBench: A Standardized Evaluation Framework for Automated Red Teaming and Robust Refusal}, 
      author={Mantas Mazeika and Long Phan and Xuwang Yin and Andy Zou and Zifan Wang and Norman Mu and Elham Sakhaee and Nathaniel Li and Steven Basart and Bo Li and David Forsyth and Dan Hendrycks},
      year={2024},
      eprint={2402.04249},
      archivePrefix={arXiv},
      primaryClass={cs.LG},
      url={https://arxiv.org/abs/2402.04249}, 
}

@InProceedings{toolsapis2026,
author="Al-Refai, Omar
and Sufian, Rula
and Abo Rubeieh, Khaled
and Abu Rmaileh, Najem Aldeen M.
and Abu-Sharkh, Osama M. F.
and Krishnan, Sriram",
editor="Patel, Kanubhai K.
and Santosh, KC
and Gomes de Oliveira, Gabriel
and Patel, Atul
and Ghosh, Ashish",
title="Design of a Web-Based Platform to Leverage MATLAB Functionality for Digital Engineering",
booktitle="Soft Computing and Its Engineering Applications",
year="2026",
publisher="Springer Nature Switzerland",
address="Cham",
pages="69--83",
isbn="978-3-032-22065-3"
}

@misc{mu2024llmsfollowsimplerules,
      title={Can LLMs Follow Simple Rules?}, 
      author={Norman Mu and Sarah Chen and Zifan Wang and Sizhe Chen and David Karamardian and Lulwa Aljeraisy and Basel Alomair and Dan Hendrycks and David Wagner},
      year={2024},
      eprint={2311.04235},
      archivePrefix={arXiv},
      primaryClass={cs.AI},
      url={https://arxiv.org/abs/2311.04235}, 
}

@misc{zou2025representationengineeringtopdownapproach,
      title={Representation Engineering: A Top-Down Approach to AI Transparency}, 
      author={Andy Zou and Long Phan and Sarah Chen and James Campbell and Phillip Guo and Richard Ren and Alexander Pan and Xuwang Yin and Mantas Mazeika and Ann-Kathrin Dombrowski and Shashwat Goel and Nathaniel Li and Michael J. Byun and Zifan Wang and Alex Mallen and Steven Basart and Sanmi Koyejo and Dawn Song and Matt Fredrikson and J. Zico Kolter and Dan Hendrycks},
      year={2025},
      eprint={2310.01405},
      archivePrefix={arXiv},
      primaryClass={cs.LG},
      url={https://arxiv.org/abs/2310.01405}, 
}

@misc{zou2023universaltransferableadversarialattacks,
      title={Universal and Transferable Adversarial Attacks on Aligned Language Models}, 
      author={Andy Zou and Zifan Wang and Nicholas Carlini and Milad Nasr and J. Zico Kolter and Matt Fredrikson},
      year={2023},
      eprint={2307.15043},
      archivePrefix={arXiv},
      primaryClass={cs.CL},
      url={https://arxiv.org/abs/2307.15043}, 
}

@misc{kaufmann2023testingrobustnessunforeseenadversaries,
      title={Testing Robustness Against Unforeseen Adversaries}, 
      author={Max Kaufmann and Daniel Kang and Yi Sun and Steven Basart and Xuwang Yin and Mantas Mazeika and Akul Arora and Adam Dziedzic and Franziska Boenisch and Tom Brown and Jacob Steinhardt and Dan Hendrycks},
      year={2023},
      eprint={1908.08016},
      archivePrefix={arXiv},
      primaryClass={cs.LG},
      url={https://arxiv.org/abs/1908.08016}, 
}

@misc{pan2023rewardsjustifymeansmeasuring,
      title={Do the Rewards Justify the Means? Measuring Trade-Offs Between Rewards and Ethical Behavior in the MACHIAVELLI Benchmark}, 
      author={Alexander Pan and Jun Shern Chan and Andy Zou and Nathaniel Li and Steven Basart and Thomas Woodside and Jonathan Ng and Hanlin Zhang and Scott Emmons and Dan Hendrycks},
      year={2023},
      eprint={2304.03279},
      archivePrefix={arXiv},
      primaryClass={cs.LG},
      url={https://arxiv.org/abs/2304.03279}, 
}

@misc{srivastava2023imitationgamequantifyingextrapolating,
      title={Beyond the Imitation Game: Quantifying and extrapolating the capabilities of language models}, 
      author={Aarohi Srivastava and Abhinav Rastogi and Abhishek Rao and Abu Awal Md Shoeb and Abubakar Abid and Adam Fisch and Adam R. Brown and Adam Santoro and Aditya Gupta and Adrià Garriga-Alonso and Agnieszka Kluska and Aitor Lewkowycz and Akshat Agarwal and Alethea Power and Alex Ray and Alex Warstadt and Alexander W. Kocurek and Ali Safaya and Ali Tazarv and Alice Xiang and Alicia Parrish and Allen Nie and Aman Hussain and Amanda Askell and Amanda Dsouza and Ambrose Slone and Ameet Rahane and Anantharaman S. Iyer and Anders Andreassen and Andrea Madotto and Andrea Santilli and Andreas Stuhlmüller and Andrew Dai and Andrew La and Andrew Lampinen and Andy Zou and Angela Jiang and Angelica Chen and Anh Vuong and Animesh Gupta and Anna Gottardi and Antonio Norelli and Anu Venkatesh and Arash Gholamidavoodi and Arfa Tabassum and Arul Menezes and Arun Kirubarajan and Asher Mullokandov and Ashish Sabharwal and Austin Herrick and Avia Efrat and Aykut Erdem and Ayla Karakaş and B. Ryan Roberts and Bao Sheng Loe and Barret Zoph and Bartłomiej Bojanowski and Batuhan Özyurt and Behnam Hedayatnia and Behnam Neyshabur and Benjamin Inden and Benno Stein and Berk Ekmekci and Bill Yuchen Lin and Blake Howald and Bryan Orinion and Cameron Diao and Cameron Dour and Catherine Stinson and Cedrick Argueta and César Ferri Ramírez and Chandan Singh and Charles Rathkopf and Chenlin Meng and Chitta Baral and Chiyu Wu and Chris Callison-Burch and Chris Waites and Christian Voigt and Christopher D. Manning and Christopher Potts and Cindy Ramirez and Clara E. Rivera and Clemencia Siro and Colin Raffel and Courtney Ashcraft and Cristina Garbacea and Damien Sileo and Dan Garrette and Dan Hendrycks and Dan Kilman and Dan Roth and Daniel Freeman and Daniel Khashabi and Daniel Levy and Daniel Moseguí González and Danielle Perszyk and Danny Hernandez and Danqi Chen and Daphne Ippolito and Dar Gilboa and David Dohan and David Drakard and David Jurgens and Debajyoti Datta and Deep Ganguli and Denis Emelin and Denis Kleyko and Deniz Yuret and Derek Chen and Derek Tam and Dieuwke Hupkes and Diganta Misra and Dilyar Buzan and Dimitri Coelho Mollo and Diyi Yang and Dong-Ho Lee and Dylan Schrader and Ekaterina Shutova and Ekin Dogus Cubuk and Elad Segal and Eleanor Hagerman and Elizabeth Barnes and Elizabeth Donoway and Ellie Pavlick and Emanuele Rodola and Emma Lam and Eric Chu and Eric Tang and Erkut Erdem and Ernie Chang and Ethan A. Chi and Ethan Dyer and Ethan Jerzak and Ethan Kim and Eunice Engefu Manyasi and Evgenii Zheltonozhskii and Fanyue Xia and Fatemeh Siar and Fernando Martínez-Plumed and Francesca Happé and Francois Chollet and Frieda Rong and Gaurav Mishra and Genta Indra Winata and Gerard de Melo and Germán Kruszewski and Giambattista Parascandolo and Giorgio Mariani and Gloria Wang and Gonzalo Jaimovitch-López and Gregor Betz and Guy Gur-Ari and Hana Galijasevic and Hannah Kim and Hannah Rashkin and Hannaneh Hajishirzi and Harsh Mehta and Hayden Bogar and Henry Shevlin and Hinrich Schütze and Hiromu Yakura and Hongming Zhang and Hugh Mee Wong and Ian Ng and Isaac Noble and Jaap Jumelet and Jack Geissinger and Jackson Kernion and Jacob Hilton and Jaehoon Lee and Jaime Fernández Fisac and James B. Simon and James Koppel and James Zheng and James Zou and Jan Kocoń and Jana Thompson and Janelle Wingfield and Jared Kaplan and Jarema Radom and Jascha Sohl-Dickstein and Jason Phang and Jason Wei and Jason Yosinski and Jekaterina Novikova and Jelle Bosscher and Jennifer Marsh and Jeremy Kim and Jeroen Taal and Jesse Engel and Jesujoba Alabi and Jiacheng Xu and Jiaming Song and Jillian Tang and Joan Waweru and John Burden and John Miller and John U. Balis and Jonathan Batchelder and Jonathan Berant and Jörg Frohberg and Jos Rozen and Jose Hernandez-Orallo and Joseph Boudeman and Joseph Guerr and Joseph Jones and Joshua B. Tenenbaum and Joshua S. Rule and Joyce Chua and Kamil Kanclerz and Karen Livescu and Karl Krauth and Karthik Gopalakrishnan and Katerina Ignatyeva and Katja Markert and Kaustubh D. Dhole and Kevin Gimpel and Kevin Omondi and Kory Mathewson and Kristen Chiafullo and Ksenia Shkaruta and Kumar Shridhar and Kyle McDonell and Kyle Richardson and Laria Reynolds and Leo Gao and Li Zhang and Liam Dugan and Lianhui Qin and Lidia Contreras-Ochando and Louis-Philippe Morency and Luca Moschella and Lucas Lam and Lucy Noble and Ludwig Schmidt and Luheng He and Luis Oliveros Colón and Luke Metz and Lütfi Kerem Şenel and Maarten Bosma and Maarten Sap and Maartje ter Hoeve and Maheen Farooqi and Manaal Faruqui and Mantas Mazeika and Marco Baturan and Marco Marelli and Marco Maru and Maria Jose Ramírez Quintana and Marie Tolkiehn and Mario Giulianelli and Martha Lewis and Martin Potthast and Matthew L. Leavitt and Matthias Hagen and Mátyás Schubert and Medina Orduna Baitemirova and Melody Arnaud and Melvin McElrath and Michael A. Yee and Michael Cohen and Michael Gu and Michael Ivanitskiy and Michael Starritt and Michael Strube and Michał Swędrowski and Michele Bevilacqua and Michihiro Yasunaga and Mihir Kale and Mike Cain and Mimee Xu and Mirac Suzgun and Mitch Walker and Mo Tiwari and Mohit Bansal and Moin Aminnaseri and Mor Geva and Mozhdeh Gheini and Mukund Varma T and Nanyun Peng and Nathan A. Chi and Nayeon Lee and Neta Gur-Ari Krakover and Nicholas Cameron and Nicholas Roberts and Nick Doiron and Nicole Martinez and Nikita Nangia and Niklas Deckers and Niklas Muennighoff and Nitish Shirish Keskar and Niveditha S. Iyer and Noah Constant and Noah Fiedel and Nuan Wen and Oliver Zhang and Omar Agha and Omar Elbaghdadi and Omer Levy and Owain Evans and Pablo Antonio Moreno Casares and Parth Doshi and Pascale Fung and Paul Pu Liang and Paul Vicol and Pegah Alipoormolabashi and Peiyuan Liao and Percy Liang and Peter Chang and Peter Eckersley and Phu Mon Htut and Pinyu Hwang and Piotr Miłkowski and Piyush Patil and Pouya Pezeshkpour and Priti Oli and Qiaozhu Mei and Qing Lyu and Qinlang Chen and Rabin Banjade and Rachel Etta Rudolph and Raefer Gabriel and Rahel Habacker and Ramon Risco and Raphaël Millière and Rhythm Garg and Richard Barnes and Rif A. Saurous and Riku Arakawa and Robbe Raymaekers and Robert Frank and Rohan Sikand and Roman Novak and Roman Sitelew and Ronan LeBras and Rosanne Liu and Rowan Jacobs and Rui Zhang and Ruslan Salakhutdinov and Ryan Chi and Ryan Lee and Ryan Stovall and Ryan Teehan and Rylan Yang and Sahib Singh and Saif M. Mohammad and Sajant Anand and Sam Dillavou and Sam Shleifer and Sam Wiseman and Samuel Gruetter and Samuel R. Bowman and Samuel S. Schoenholz and Sanghyun Han and Sanjeev Kwatra and Sarah A. Rous and Sarik Ghazarian and Sayan Ghosh and Sean Casey and Sebastian Bischoff and Sebastian Gehrmann and Sebastian Schuster and Sepideh Sadeghi and Shadi Hamdan and Sharon Zhou and Shashank Srivastava and Sherry Shi and Shikhar Singh and Shima Asaadi and Shixiang Shane Gu and Shubh Pachchigar and Shubham Toshniwal and Shyam Upadhyay and Shyamolima and Debnath and Siamak Shakeri and Simon Thormeyer and Simone Melzi and Siva Reddy and Sneha Priscilla Makini and Soo-Hwan Lee and Spencer Torene and Sriharsha Hatwar and Stanislas Dehaene and Stefan Divic and Stefano Ermon and Stella Biderman and Stephanie Lin and Stephen Prasad and Steven T. Piantadosi and Stuart M. Shieber and Summer Misherghi and Svetlana Kiritchenko and Swaroop Mishra and Tal Linzen and Tal Schuster and Tao Li and Tao Yu and Tariq Ali and Tatsu Hashimoto and Te-Lin Wu and Théo Desbordes and Theodore Rothschild and Thomas Phan and Tianle Wang and Tiberius Nkinyili and Timo Schick and Timofei Kornev and Titus Tunduny and Tobias Gerstenberg and Trenton Chang and Trishala Neeraj and Tushar Khot and Tyler Shultz and Uri Shaham and Vedant Misra and Vera Demberg and Victoria Nyamai and Vikas Raunak and Vinay Ramasesh and Vinay Uday Prabhu and Vishakh Padmakumar and Vivek Srikumar and William Fedus and William Saunders and William Zhang and Wout Vossen and Xiang Ren and Xiaoyu Tong and Xinran Zhao and Xinyi Wu and Xudong Shen and Yadollah Yaghoobzadeh and Yair Lakretz and Yangqiu Song and Yasaman Bahri and Yejin Choi and Yichi Yang and Yiding Hao and Yifu Chen and Yonatan Belinkov and Yu Hou and Yufang Hou and Yuntao Bai and Zachary Seid and Zhuoye Zhao and Zijian Wang and Zijie J. Wang and Zirui Wang and Ziyi Wu},
      year={2023},
      eprint={2206.04615},
      archivePrefix={arXiv},
      primaryClass={cs.CL},
      url={https://arxiv.org/abs/2206.04615}, 
}

@misc{bustan2026moltbot,
  author       = {Siman Tov Bustan, Moshe and Zadok, Nir},
  title        = {One Step Away From a Massive Data Breach: What We Found Inside {MoltBot}},
  year         = {2026},
  month        = jan,
  day          = {29},
  howpublished = {OX Security Blog},
  url          = {https://www.ox.security/blog/one-step-away-from-a-massive-data-breach-what-we-found-inside-moltbot/},
  note         = {Accessed: 2026-02-04}
}

@article{campbell2026zerotrust,
	doi = {10.20944/preprints202602.0085.v1},
	url = {https://doi.org/10.20944/preprints202602.0085.v1},
	year = 2026,
	month = {February},
	publisher = {Preprints},
	author = {Robert Campbell},
	title = {Zero Trust for AI Systems: A Reference Architecture and Assurance Framework},
	journal = {Preprints}
}

@misc{huang2025novelzerotrustidentityframework,
      title={A Novel Zero-Trust Identity Framework for Agentic AI: Decentralized Authentication and Fine-Grained Access Control}, 
      author={Ken Huang and Vineeth Sai Narajala and John Yeoh and Jason Ross and Ramesh Raskar and Youssef Harkati and Jerry Huang and Idan Habler and Chris Hughes},
      year={2025},
      eprint={2505.19301},
      archivePrefix={arXiv},
      primaryClass={cs.CR},
      url={https://arxiv.org/abs/2505.19301}, 
}

@article{gurram2025generative,
  title={Generative AI for enhanced cybersecurity: building a zero-trust architecture with agentic AI},
  author={Gurram, Abhyudaya},
  journal={World J. Adv. Eng. Technol. Sci},
  volume={15},
  number={1},
  pages={2380--2396},
  year={2025}
}

@techreport{IEC2022,
  author      = {{International Electrotechnical Commission}},
  title       = {IEC 61508 \& Functional Safety},
  institution = {IEC},
  year        = {2022},
  url         = {https://assets.iec.ch/public/acos/IEC%2061508%20&%20Functional%20Safety-2022.pdf?2023040501},
  note        = {Accessed: 2025-02-16}
}

@online{Shinde2024,
  author       = {Shinde, Chaitanya},
  title        = {Navigating {SOTIF} ({ISO} 21448) and Ensuring Safety in Autonomous Driving},
  organization = {Automotive IQ},
  date         = {2024-03-25},
  year         = {2024},
  url          = {https://www.automotive-iq.com/functional-safety/articles/navigating-sotif-iso-21448-and-ensuring-safety-in-autonomous-driving},
  note         = {Accessed: 2025-02-16}
}

@inproceedings{shahbaz2025flc,
 author={Shahbaz, Ibrahim and Hammad, Eman and Farraj, Abdallah},
  booktitle={2026 IEEE/PES Transmission and Distribution Conference and Exposition (T\&D)}, 
  title={An Interpretable Federated Learning Control Framework Design for Smart Grid Resilience}, 
  year={2026},
  volume={},
  number={},
  pages={1-5},
  doi={10.1109/TD48022.2026.11562857}
}

@inproceedings{2026tpeckan,
   author={Shahbaz, Ibrahim and Lagoy, Isaac and Al-Refai, Omar and Hammad, Eman},
  booktitle={2026 IEEE Texas Power and Energy Conference (TPEC)}, 
  title={Evaluating Interpretable Kolmogorov–Arnold Network Controllers for Smart Grid Resilience}, 
  year={2026},
  volume={},
  number={},
  pages={1-6},
  doi={10.1109/TPEC67884.2026.11513111}}

@misc{shahbaz2026inertiainformedfederatedlearningcontrol,
      title={Inertia-Informed Federated Learning Control Framework for Distributed Smart Grid Resilience}, 
      author={Ibrahim Shahbaz and Omar Al-Refai and Eman Hammad},
      year={2026},
      eprint={2607.05720},
      archivePrefix={arXiv},
      primaryClass={eess.SY},
      url={https://arxiv.org/abs/2607.05720}, 
}

@misc{alrefai2026federatedphysicsgroundedreinforcementlearning,
      title={Federated Physics-Grounded Reinforcement Learning for Distributed Stability Control in Smart Grids}, 
      author={Omar Al-Refai and Ibrahim Shahbaz and Adam Ali Husseinat and Eman Hammad},
      year={2026},
      eprint={2607.05553},
      archivePrefix={arXiv},
      primaryClass={cs.LG},
      url={https://arxiv.org/abs/2607.05553}, 
}

@techreport{IEC61508FunctionalSafety2022,
  author      = {{International Electrotechnical Commission}},
  title       = {IEC 61508 \& Functional Safety},
  institution = {IEC},
  year        = {2022},
  url         = {https://assets.iec.ch/public/acos/IEC%2061508%20&%20Functional%20Safety-2022.pdf?2023040501},
  note        = {Accessed: 2025-02-16}
}

@online{MTDEQ_FMEA_AppR,
  author       = {{Montana Department of Environmental Quality}},
  title        = {Failure Modes Effects Analysis (FMEA)},
  year         = {n.d.},
  url          = {https://deq.mt.gov/files/Land/Hardrock/Documents/TintinaMines/App%20R%20Failure%20Modes%20Effects%20Analysis/App%20R%20Failure%20Modes%20Effects%20Analysis.pdf},
  organization = {Montana Department of Environmental Quality},
  urldate      = {2026-02-17},
  note         = {PDF document}
}

@online{MoD_ASEMS_FMEA,
  author       = {{UK Ministry of Defence}},
  title        = {{ASEMS} Toolkit: {FMEA/FMECA}},
  year         = {2020},
  url          = {https://web.archive.org/web/20200813203140/https://www.asems.mod.uk/toolkit/fmeafmeca},
  organization = {Defence Equipment and Support (DE\&S)},
  urldate      = {2025-02-17},
  note         = {Archived version from August 13, 2020}
}

@techreport{AI_Office_Pact_2024,
  author      = {{European Artificial Intelligence Office}},
  title       = {{AI Pact}: Organisations' commitments},
  institution = {European Commission},
  year        = {2024},
  month       = sep,
  url         = {https://artificialintelligenceact.eu/wp-content/uploads/2025/03/2024.09.25-AI-Pact-final.pdf},
  note        = {Accessed: 2025-02-17}
}

@online{AI_Act_Compliance_Flowchart_2025,
  author       = {{Future of Life Institute}},
  title        = {{AI} Act Compliance Checker Flowchart (v1.0)},
  year         = {2025},
  month        = may,
  url          = {https://artificialintelligenceact.eu/wp-content/uploads/2025/07/AI-Act-Compliance-Checker-Flowchart-v1.0_compressed.pdf},
  organization = {Future of Life Institute},
  note         = {Accessed: 2025-02-17}
}

@online{SynopsysISO26262,
  author       = {{Synopsys}},
  title        = {What is {ISO} 26262 Functional Safety Standard?},
  year         = {n.d.},
  url          = {https://www.synopsys.com/glossary/what-is-iso-26262.html},
  organization = {Synopsys, Inc.},
  note         = {Accessed: 2025-02-17}
}

@online{RagaAI_AAEF_Docs,
  author       = {{RagaAI}},
  title        = {{RagaAI AAEF} ({Agentic Application Evaluation Framework})},
  year         = {2025},
  url          = {https://docs.raga.ai/ragaai-aaef-agentic-application-evaluation-framework},
  organization = {RagaAI Catalyst},
  note         = {Accessed: 2026-02-18}
}

@techreport{RagaAI_AAEF_2024,
  author      = {{RagaAI}},
  title       = {Whitepaper: Agentic Application Evaluation Framework ({AAEF})},
  institution = {RagaAI, Inc.},
  year        = {2024},
  month       = jun,
  day         = {12},
  url         = {https://raga.ai/resources/patentsandpublications/whitepaper-agentic-application-evaluation-framework},
  note        = {Accessed: 2025-02-18}
}

@online{Starkloff2026Evaluations,
  author       = {Starkloff, Anne-Gabrielle and Kokaina, Sallah and Rahimi, Sohrab},
  title        = {Evaluations for the agentic world},
  editor       = {Stichbury, Jo},
  year         = {2026},
  month        = jan,
  day          = {29},
  url          = {https://medium.com/quantumblack/evaluations-for-the-agentic-world-c3c150f0dd5a},
  organization = {QuantumBlack, AI by McKinsey},
  note         = {Accessed: 2026-02-18}
}

@online{MoralesAguilera2025Building,
  author       = {Morales Aguilera, Frank},
  title        = {Building Real-World Agentic {AI} Systems: A Practical Guide},
  year         = {2025},
  month        = jul,
  day          = {30},
  url          = {https://medium.com/ai-simplified-in-plain-english/building-real-world-agentic-ai-systems-a-practical-guide-9748d572b58b},
  organization = {AI Simplified in Plain English},
  note         = {Accessed: 2026-02-18}
}

@online{WizExperts2025MDRvsSOC,
  author       = {{Wiz Experts Team}},
  title        = {{MDR} vs. {SOC}: What's The Difference?},
  year         = {2025},
  month        = may,
  day          = {22},
  url          = {https://www.wiz.io/academy/detection-and-response/mdr-vs-soc},
  organization = {Wiz, Inc.},
  note         = {Accessed: 2026-02-18}
}

@misc{levy2025stwebagentbenchbenchmarkevaluatingsafety,
      title={ST-WebAgentBench: A Benchmark for Evaluating Safety and Trustworthiness in Web Agents}, 
      author={Ido Levy and Ben Wiesel and Sami Marreed and Alon Oved and Avi Yaeli and Segev Shlomov},
      year={2025},
      eprint={2410.06703},
      archivePrefix={arXiv},
      primaryClass={cs.AI},
      url={https://arxiv.org/abs/2410.06703}, 
}

@online{Sammeta2025ReportCard,
  author       = {Sammeta, LaxmiKumar Reddy},
  title        = {The {AI} Agent Report Card You've Been Ignoring: Why 30\% of Your Agent's ``Successes'' Are Actually Failures},
  year         = {2025},
  month        = dec,
  day          = {5},
  url          = {https://laxmikumars.medium.com/the-ai-agent-report-card-youve-been-ignoring-why-30-of-your-agent-s-successes-are-actually-498fbebf44f9},
  note         = {Accessed: 2026-02-23}
}

@misc{ferrag2026alpha3benchunifiedbenchmarksafety,
      title={$\alpha^3$-Bench: A Unified Benchmark of Safety, Robustness, and Efficiency for LLM-Based UAV Agents over 6G Networks}, 
      author={Mohamed Amine Ferrag and Abderrahmane Lakas and Merouane Debbah},
      year={2026},
      eprint={2601.03281},
      archivePrefix={arXiv},
      primaryClass={eess.SY},
      url={https://arxiv.org/abs/2601.03281}, 
}

@misc{arike2025technicalreportevaluatinggoal,
      title={Technical Report: Evaluating Goal Drift in Language Model Agents}, 
      author={Rauno Arike and Elizabeth Donoway and Henning Bartsch and Marius Hobbhahn},
      year={2025},
      eprint={2505.02709},
      archivePrefix={arXiv},
      primaryClass={cs.AI},
      url={https://arxiv.org/abs/2505.02709}, 
}

@article{ManishEvalAgents2025,
	doi = {10.20944/preprints202508.1847.v1},
	url = {https://doi.org/10.20944/preprints202508.1847.v1},
	year = 2025,
	month = {August},
	publisher = {Preprints},
	author = {Manish Shukla},
	title = {Evaluating Agentic AI Systems: A Balanced Framework for Performance, Robustness, Safety and Beyond},
	journal = {Preprints}
}

@online{Yadav2025TenEssential,
  author       = {Yadav, Navya},
  title        = {10 Essential Steps for Evaluating the Reliability of {AI} Agents},
  year         = {2025},
  month        = nov,
  day          = {26},
  url          = {https://www.getmaxim.ai/articles/10-essential-steps-for-evaluating-the-reliability-of-ai-agents/},
  organization = {Maxim AI},
  note         = {Accessed: 2026-02-24}
}

@misc{qiu2025chainoftriggeragenticbackdoorparadoxically,
      title={Chain-of-Trigger: An Agentic Backdoor that Paradoxically Enhances Agentic Robustness}, 
      author={Jiyang Qiu and Xinbei Ma and Yunqing Xu and Zhuosheng Zhang and Hai Zhao},
      year={2025},
      eprint={2510.08238},
      archivePrefix={arXiv},
      primaryClass={cs.AI},
      url={https://arxiv.org/abs/2510.08238}, 
}

@online{Mason2025ISO42001,
  author       = {Mason, Danielle},
  title        = {{ISO/IEC} 42001 {AI} Security Implementation Guide},
  year         = {2025},
  month        = oct,
  day          = {1},
  url          = {https://www.bdemerson.com/article/iso-iec-42001-ai-security-implementation-guide},
  organization = {BD Emerson},
  note         = {Accessed: 2026-02-24}
}

@online{Rijo2026UCBerkeley,
  author       = {Rijo, Luis},
  title        = {{UC} Berkeley unveils framework as {AI} agents threaten to outrun oversight},
  year         = {2026},
  month        = feb,
  day          = {15},
  url          = {https://ppc.land/uc-berkeley-unveils-framework-as-ai-agents-threaten-to-outrun-oversight/},
  organization = {PPC Land},
  note         = {Accessed: 2026-02-24}
}

@misc{carlini2025autoadvexbenchbenchmarkingautonomousexploitation,
      title={AutoAdvExBench: Benchmarking autonomous exploitation of adversarial example defenses}, 
      author={Nicholas Carlini and Javier Rando and Edoardo Debenedetti and Milad Nasr and Florian Tramèr},
      year={2025},
      eprint={2503.01811},
      archivePrefix={arXiv},
      primaryClass={cs.CR},
      url={https://arxiv.org/abs/2503.01811}, 
}

@inproceedings{Rahman2021PEPPA-X:,title={PEPPA-X: Finding Program Test Inputs to Bound Silent Data Corruption Vulnerability in HPC Applications},author={Md. Hasanur Rahman and Aabid Shamji and Shengjian Guo and Guanpeng Li},booktitle={SC21: International Conference for High Performance Computing, Networking, Storage and Analysis},year={2021},pages={1-14},doi={10.1145/3458817.3476147}}

@ARTICLE{Gholami2024AI,
  author={Gholami, Amir and Yao, Zhewei and Kim, Sehoon and Hooper, Coleman and Mahoney, Michael W. and Keutzer, Kurt},
  journal={IEEE Micro}, 
  title={AI and Memory Wall}, 
  year={2024},
  volume={44},
  number={3},
  pages={33-39},
  doi={10.1109/MM.2024.3373763}}

@article{Chang2025SagaLLM,
author = {Chang, Edward Y. and Geng, Longling},
title = {SagaLLM: Context Management, Validation, and Transaction Guarantees for Multi-Agent LLM Planning},
year = {2025},
issue_date = {August 2025},
publisher = {VLDB Endowment},
volume = {18},
number = {12},
issn = {2150-8097},
url = {https://doi.org/10.14778/3750601.3750611},
doi = {10.14778/3750601.3750611},
journal = {Proc. VLDB Endow.},
month = aug,
pages = {4874–4886},
numpages = {13}
}

@article{Rajput2020Multi-agent,title={Multi-agent architecture for fault recovery in self-healing systems},author={P. Rajput and Geeta Sikka},journal={Journal of Ambient Intelligence and Humanized Computing},year={2020},volume={12},pages={2849 - 2866},doi={10.1007/s12652-020-02443-8}}

@INPROCEEDINGS{Vinay2025CoMAS-HPC,
  author={Tejas Vinay, P and Saurav, Sumit Kumar},
  booktitle={2025 Supercomputing India (SCI)}, 
  title={CoMAS-HPC: A Collaborative Multi-Agent System for HPC Administration}, 
  year={2025},
  volume={},
  number={},
  pages={1-8},
  doi={10.1109/SCI68648.2025.11333875}}

@INPROCEEDINGS{Sidorov2016Methods,
  author={Sidorov, I.A.},
  booktitle={2016 39th International Convention on Information and Communication Technology, Electronics and Microelectronics (MIPRO)}, 
  title={Methods and tools to increase fault tolerance of high-performance computing systems}, 
  year={2016},
  volume={},
  number={},
  pages={226-230},
  doi={10.1109/MIPRO.2016.7522142}}

@INPROCEEDINGS{mricnndiagnosis,
  author={Al-Refai, Omar and Alabed, Aya},
  booktitle={2025 16th International Conference on Information and Communication Systems (ICICS)}, 
  title={Explainable Brain Tumor Classification Using Transfer Learning of Deep Convolutional Neural Networks}, 
  year={2025},
  volume={},
  number={},
  pages={1-6},
  doi={10.1109/ICICS65354.2025.11073099},
  }

@INPROCEEDINGS{Park2005MAS,
  author={Jeongmin Park and Giljong Yoo and Eunseok Lee},
  booktitle={Third ACIS Int'l Conference on Software Engineering Research, Management and Applications (SERA'05)}, 
  title={Proactive self-healing system based on multi-agent technologies}, 
  year={2005},
  volume={},
  number={},
  pages={256-263},
  doi={10.1109/SERA.2005.55}}

@incollection{dongarra2015fault,
  title={Fault tolerance techniques for high-performance computing},
  author={Dongarra, Jack and Herault, Thomas and Robert, Yves},
  booktitle={Fault-tolerance techniques for high-performance computing},
  pages={3--85},
  year={2015},
  publisher={Springer}
}

@ARTICLE{abft_og,
  author={Kuang-Hua Huang and Abraham, Jacob A.},
  journal={IEEE Transactions on Computers}, 
  title={Algorithm-Based Fault Tolerance for Matrix Operations}, 
  year={1984},
  volume={C-33},
  number={6},
  pages={518-528},
  doi={10.1109/TC.1984.1676475}}

@inproceedings{Hespe2022ReStore,
  author    = {Hespe, Demian and H{\"u}bner, Lukas and Sanders, Peter and Stamatakis, Alexandros},
  editor    = {Cano, Jos{\'e} and Tr{\"a}ff, Jesper Larsson},
  title     = {ReStore: In-Memory REplicated STORagE for Rapid Recovery in Fault-Tolerant Algorithms},
  booktitle = {Euro-Par 2022: Parallel Processing},
  year      = {2022},
  publisher = {Springer International Publishing},
  address   = {Cham},
  pages     = {203--217},
  doi       = {10.1007/978-3-031-12597-3_13},
  url       = {https://doi.org/10.1007/978-3-031-12597-3_13}
}

@misc{ye2026clawevaltrustworthyevaluationautonomous,
      title={Claw-Eval: Toward Trustworthy Evaluation of Autonomous Agents}, 
      author={Bowen Ye and Rang Li and Qibin Yang and Yuanxin Liu and Linli Yao and Hanglong Lv and Zhihui Xie and Chenxin An and Lei Li and Lingpeng Kong and Qi Liu and Zhifang Sui and Tong Yang},
      year={2026},
      eprint={2604.06132},
      archivePrefix={arXiv},
      primaryClass={cs.AI},
      url={https://arxiv.org/abs/2604.06132}, 
}

@article{survey_communication,
author = {Vangalapat, Tharakesavulu and Shaikh, Samreen},
year = {2026},
month = {02},
pages = {1660},
title = {Trustworthy Agentic AI: A Survey and Taxonomy of Secure Coordination and Hallucination Mitigation in Multi-Agent Large Language Model Systems},
journal = {International Journal of Innovative Science and Research Technology},
doi = {10.38124/ijisrt/26feb1090}
}

@online{HackerNoon_AIAgents2024,
  author       = {{HackerNoon}},
  title        = {Inside the Push to Standardize Communication Between {AI} Agents},
  year         = {2024},
  url          = {https://hackernoon.com/inside-the-push-to-standardize-communication-between-ai-agents},
  organization = {HackerNoon},
  note         = {Accessed: 2026-04-15}
}

@misc{du2026enabling,
title={Enabling Agents to Communicate Entirely in Latent Space},
author={Zhuoyun Du and Runze Wang and Huiyu BAI and zouying cao and Xiaoyong Zhu and Bo Zheng and Wei Chen and Haochao Ying},
year={2026},
url={https://openreview.net/forum?id=rmYbgsehTd}
}

@inproceedings{zheng2025thought,
title={Thought Communication in Multiagent Collaboration},
author={Yujia Zheng and Zhuokai Zhao and Zijian Li and Yaqi Xie and Mingze Gao and Lizhu Zhang and Kun Zhang},
booktitle={The Thirty-ninth Annual Conference on Neural Information Processing Systems},
year={2025},
url={https://openreview.net/forum?id=tq9lyV9Cml}
}

@misc{jo2025byzantinerobustdecentralizedcoordinationllm,
      title={Byzantine-Robust Decentralized Coordination of LLM Agents}, 
      author={Yongrae Jo and Chanik Park},
      year={2025},
      eprint={2507.14928},
      archivePrefix={arXiv},
      primaryClass={cs.DC},
      url={https://arxiv.org/abs/2507.14928}, 
}

@inproceedings{BlockAgents,
author = {Chen, Bei and Li, Gaolei and Lin, Xi and Wang, Zheng and Li, Jianhua},
title = {BlockAgents: Towards Byzantine-Robust LLM-Based Multi-Agent Coordination via Blockchain},
year = {2024},
isbn = {9798400710117},
publisher = {Association for Computing Machinery},
address = {New York, NY, USA},
url = {https://doi.org/10.1145/3674399.3674445},
doi = {10.1145/3674399.3674445},
booktitle = {Proceedings of the ACM Turing Award Celebration Conference - China 2024},
pages = {187–192},
numpages = {6},
location = {Changsha, China},
series = {ACM-TURC '24}
}

@inproceedings{jacobs2021lumi,
  author    = {Arthur S. Jacobs and Ricardo J. Pfitscher and Rafael H. Ribeiro and Ronaldo A. Ferreira and Lisandro Z. Granville and Walter Willinger and Sanjay G. Rao},
  title     = {Hey, Lumi! Using Natural Language for Intent-Based Network Management},
  booktitle = {2021 USENIX Annual Technical Conference (USENIX ATC 21)},
  year      = {2021},
  pages     = {625--639},
  url       = {https://www.usenix.org/conference/atc21/presentation/jacobs}
}

@inproceedings{kim2023intender,
  author    = {Jiwon Kim and Dongyeong Kim and Cody Cutler and Zhiqiang Lin and Dongyan Xu and Xiangyu Zhang},
  title     = {INTENDER: Fuzzing Intent-Based Networking with Intent-State Transition Guidance},
  booktitle = {32nd USENIX Security Symposium (USENIX Security 23)},
  year      = {2023},
  pages     = {8861--8878},
  url       = {https://www.usenix.org/conference/usenixsecurity23/presentation/kim-jiwon}
}

@article{xie2021deepsc,
  author  = {Huiqiang Xie and Zhijin Qin and Geoffrey Ye Li and Biing-Hwang Juang},
  title   = {Deep Learning Enabled Semantic Communication Systems},
  journal = {IEEE Transactions on Signal Processing},
  volume  = {69},
  pages   = {2663--2675},
  year    = {2021},
  doi     = {10.1109/TSP.2021.3071210},
  url     = {https://arxiv.org/abs/2006.10685}
}

@article{weng2021speech,
  author  = {Zhenzi Weng and Zhijin Qin},
  title   = {Semantic Communication Systems for Speech Transmission},
  journal = {IEEE Journal on Selected Areas in Communications},
  volume  = {39},
  number  = {8},
  pages   = {2434--2444},
  year    = {2021},
  doi     = {10.1109/JSAC.2021.3087240},
  url     = {https://arxiv.org/pdf/2102.12605}
}

@inproceedings{manias2024intent,
  author    = {Dimitrios Michael Manias and Ali Chouman and Abdallah Shami},
  title     = {Towards Intent-Based Network Management: Large Language Models for Intent Extraction in 5G Core Networks},
  booktitle = {2024 20th International Conference on the Design of Reliable Communication Networks (DRCN)},
  year      = {2024},
  doi       = {10.1109/DRCN60692.2024.10539172},
  url       = {https://arxiv.org/abs/2403.02238}
}

@inproceedings{manias2024semanticrouting,
  author    = {Dimitrios Michael Manias and Ali Chouman and Abdallah Shami},
  title     = {Semantic Routing for Enhanced Performance of LLM-Assisted Intent-Based 5G Core Network Management and Orchestration},
  booktitle = {2024 IEEE Global Communications Conference (GLOBECOM)},
  year      = {2024},
  doi       = {10.1109/GLOBECOM52923.2024.10901065},
  url       = {https://arxiv.org/abs/2404.15869}
}

@inproceedings{qin2023securesemantic,
  author    = {Qi Qin and Yankai Rong and Guoshun Nan and Shaokang Wu and Xuefei Zhang and Qimei Cui and Xiaofeng Tao},
  title     = {Securing Semantic Communications with Physical-layer Semantic Encryption and Obfuscation},
  booktitle = {2023 IEEE International Conference on Communications (ICC)},
  year      = {2023},
  doi       = {10.1109/ICC45041.2023.10279807},
  url       = {https://arxiv.org/pdf/2304.10147}
}

@inproceedings{ruan2024toolemu,
  author    = {Yangjun Ruan and Honghua Dong and Andrew Wang and Silviu Pitis and Yongchao Zhou and Jimmy Ba and Yann Dubois and Chris J. Maddison and Tatsunori Hashimoto},
  title     = {Identifying the Risks of LM Agents with an LM-Emulated Sandbox},
  booktitle = {The Twelfth International Conference on Learning Representations (ICLR 2024)},
  year      = {2024},
  url       = {https://openreview.net/forum?id=GEcwtMk1uA}
}

@inproceedings{zhang2025asb,
  author    = {Hanrong Zhang and Jingyuan Huang and Kai Mei and Yifei Yao and Zhenting Wang and Chenlu Zhan and Hongwei Wang and Yongfeng Zhang},
  title     = {Agent Security Bench (ASB): Formalizing and Benchmarking Attacks and Defenses in LLM-based Agents},
  booktitle = {The Thirteenth International Conference on Learning Representations (ICLR 2025)},
  year      = {2025},
  url       = {https://openreview.net/forum?id=V4y0CpX4hK}
}

@inproceedings{ames2019control,
  title={Control barrier functions: Theory and applications},
  author={Ames, Aaron D and Coogan, Samuel and Egerstedt, Magnus and Notomista, Gennaro and Sreenath, Koushil and Tabuada, Paulo},
  booktitle={2019 18th European control conference (ECC)},
  pages={3420--3431},
  year={2019},
  organization={Ieee}
}

@inproceedings{mehmood2022black,
  title={The black-box simplex architecture for runtime assurance of autonomous CPS},
  author={Mehmood, Usama and Sheikhi, Sanaz and Bak, Stanley and Smolka, Scott A and Stoller, Scott D},
  booktitle={NASA formal methods symposium},
  pages={231--250},
  year={2022},
  organization={Springer}
}

@inproceedings{chen2022runtime,
  title={Runtime safety assurance for learning-enabled control of autonomous driving vehicles},
  author={Chen, Shengduo and Sun, Yaowei and Li, Dachuan and Wang, Qiang and Hao, Qi and Sifakis, Joseph},
  booktitle={2022 International Conference on Robotics and Automation (ICRA)},
  pages={8978--8984},
  year={2022},
  organization={IEEE}
}

@article{ma2024controlloc,
  title={Controlloc: Physical-world hijacking attack on visual perception in autonomous driving},
  author={Ma, Chen and Wang, Ningfei and Zhao, Zhengyu and Wang, Qian and Chen, Qi Alfred and Shen, Chao},
  journal={arXiv preprint arXiv:2406.05810},
  year={2024}
}

@inproceedings{lou2024first,
  title={A First $\{$Physical-World$\}$ Trajectory Prediction Attack via $\{$LiDAR-induced$\}$ Deceptions in Autonomous Driving},
  author={Lou, Yang and Zhu, Yi and Song, Qun and Tan, Rui and Qiao, Chunming and Lee, Wei-Bin and Wang, Jianping},
  booktitle={33rd USENIX Security Symposium (USENIX Security 24)},
  pages={6291--6308},
  year={2024}
}

@article{ma2024slowperception,
  title={Slowperception: Physical-world latency attack against visual perception in autonomous driving},
  author={Ma, Chen and Wang, Ningfei and Zhao, Zhengyu and Chen, Qi Alfred and Shen, Chao},
  journal={arXiv preprint arXiv:2406.05800},
  year={2024}
}

@article{Cappello2009Toward,title={Toward Exascale Resilience},author={F. Cappello and A. Geist and W. Gropp and L. Kalé and B. Kramer and M. Snir},journal={The International Journal of High Performance Computing Applications},year={2009},volume={23},pages={374 - 388},doi={10.1177/1094342009347767}}

@article{Losada2020Fault,title={Fault tolerance of MPI applications in exascale systems: The ULFM solution},author={Nuria Losada and P. González and María J. Martín and G. Bosilca and A. Bouteiller and K. Teranishi},journal={Future Gener. Comput. Syst.},year={2020},volume={106},pages={467-481},doi={10.1016/j.future.2020.01.026}}

@inproceedings{samadi2023safe,
  title={SAFE: Saliency-aware counterfactual explanations for DNN-based automated driving systems},
  author={Samadi, Amir and Shirian, Amir and Koufos, Konstantinos and Debattista, Kurt and Dianati, Mehrdad},
  booktitle={2023 IEEE 26th International Conference on Intelligent Transportation Systems (ITSC)},
  pages={5655--5662},
  year={2023},
  organization={IEEE}
}

@article{samadi2024safe,
  title={SAFE-RL: Saliency-aware counterfactual explainer for deep reinforcement learning policies},
  author={Samadi, Amir and Koufos, Konstantinos and Debattista, Kurt and Dianati, Mehrdad},
  journal={IEEE Robotics and Automation Letters},
  volume={9},
  number={11},
  pages={9994--10001},
  year={2024},
  publisher={IEEE}
}

@article{eldosouky2019drones,
  title={Drones in distress: A game-theoretic countermeasure for protecting UAVs against GPS spoofing},
  author={Eldosouky, AbdelRahman and Ferdowsi, Aidin and Saad, Walid},
  journal={IEEE Internet of Things Journal},
  volume={7},
  number={4},
  pages={2840--2854},
  year={2019},
  publisher={IEEE}
}

@article{davidovich2022towards,
  title={Towards the detection of GPS spoofing attacks against drones by analyzing camera’s video stream},
  author={Davidovich, Barak and Nassi, Ben and Elovici, Yuval},
  journal={Sensors},
  volume={22},
  number={7},
  pages={2608},
  year={2022},
  publisher={MDPI}
}

@article{hasan2020securing,
  title={Securing vehicle-to-everything (V2X) communication platforms},
  author={Hasan, Monowar and Mohan, Sibin and Shimizu, Takayuki and Lu, Hongsheng},
  journal={IEEE Transactions on Intelligent Vehicles},
  volume={5},
  number={4},
  pages={693--713},
  year={2020},
  publisher={IEEE}
}

@article{ling2026physical,
  title={Physical Prompt Injection Attacks on Large Vision-Language Models},
  author={Ling, Chen and Hu, Kai and Liu, Hangcheng and Han, Xingshuo and Zhang, Tianwei and Ou, Changhai},
  journal={arXiv preprint arXiv:2601.17383},
  year={2026}
}

@article{zhang2024visual,
  title={Visual adversarial attack on vision-language models for autonomous driving},
  author={Zhang, Tianyuan and Wang, Lu and Zhang, Xinwei and Zhang, Yitong and Jia, Boyi and Liang, Siyuan and Hu, Shengshan and Fu, Qiang and Liu, Aishan and Liu, Xianglong},
  journal={arXiv preprint arXiv:2411.18275},
  year={2024}
}

@article{no2021155,
  title={155 [Uniform provisions concerning the approval of vehicles with regards to cyber security and cyber security management system]},
  author={No, UN Regulation},
  journal={UNECE, UN Regulation No},
  year={2021}
}

@inproceedings{siddiqui2023cybersecurity,
  title={Cybersecurity engineering: bridging the security gaps in advanced automotive systems and ISO/SAE 21434},
  author={Siddiqui, Fahad and Khan, Rafiullah and Tasdemir, Sena Yengec and Hui, Henry and Sonigara, Balmukund and Sezer, Sakir and McLaughlin, Kieran},
  booktitle={2023 IEEE 97th Vehicular Technology Conference (VTC2023-Spring)},
  pages={1--6},
  year={2023},
  organization={IEEE}
}

@article{chen2025xgridagent,
  author  = {Yihan Wen and Xin Chen},
  title   = {{X-GridAgent}: An {LLM}-Powered Agentic {AI} System for Assisting Power Grid Analysis},
  journal = {arXiv preprint arXiv:2512.20789},
  year    = {2025},
  url     = {https://arxiv.org/abs/2512.20789}
}

@inproceedings{jin2025gridmind,
  author    = {Hongwei Jin and Kibaek Kim and Jonghwan Kwon},
  title     = {{GridMind}: {LLMs}-Powered Agents for Power System Analysis and Operations},
  booktitle = {Proceedings of the SC'25 Workshops of the International Conference for High Performance Computing, Networking, Storage and Analysis},
  pages     = {560--568},
  year      = {2025}
}

@article{badmus2025powerchain,
  author  = {Emmanuel O. Badmus and Peng Sang and Dimitrios Stamoulis and Amritanshu Pandey},
  title   = {{PowerChain}: Automating Distribution Grid Analysis with Agentic {AI} Workflows},
  journal = {arXiv preprint arXiv:2508.17094},
  year    = {2025},
  url     = {https://arxiv.org/abs/2508.17094}
}

@article{zhang2025gridagent,
  author  = {Yan Zhang and Ahmad Mohammad Saber and Amr Youssef and Deepa Kundur},
  title   = {Semantic Reasoning Meets Numerical Precision: An {LLM}-Powered Multi-Agent System for Power Grid Control},
  journal = {arXiv preprint arXiv:2508.05702},
  year    = {2025},
  url     = {https://arxiv.org/abs/2508.05702}
}

@inproceedings{saha2025dragent,
  author    = {Barun Kumar Saha and {Aarthi V.} and O. D. Naidu},
  title     = {{DrAgent}: An Agentic Approach to Fault Analysis in Power Grids Using Large Language Models},
  booktitle = {2025 International Conference on Artificial Intelligence in Information and Communication (ICAIIC)},
  pages     = {938--945},
  year      = {2025}
}

@inproceedings{zheng2025agenticplanning,
  author    = {Xiaodong Zheng and Siheng Zhao and Tianzhuo Shi and Tao Wang and Ruilin Chen and Shixuan Yu and Shuangsi Xue and Hui Cao},
  title     = {Agentic Planning for Power System Simulation Task Orchestrations Based on Multi-Agent Systems},
  booktitle = {2025 8th International Conference on Robotics, Control and Automation Engineering (RCAE)},
  pages     = {576--581},
  year      = {2025},
  doi       = {10.1109/RCAE66389.2025.11355202}
}

@article{jia2025enhancing,
  author  = {Mengshuo Jia and Zeyu Cui and Gabriela Hug},
  title   = {Enhancing {LLMs} for Power System Simulations: A Feedback-Driven Multi-Agent Framework},
  journal = {IEEE Transactions on Smart Grid},
  volume  = {16},
  pages   = {5556--5572},
  number  = {6},
  year    = {2025}
}

@article{cheng2025gaia,
  author  = {Yuheng Cheng and Huan Zhao and Xiyuan Zhou and Junhua Zhao and Yuji Cao and Chao Yang and Xinlei Cai},
  title   = {A Large Language Model for Advanced Power Dispatch},
  journal = {Scientific Reports},
  volume  = {15},
  year    = {2025}
}

@inproceedings{slavchev2026finetuned,
  author       = {Nikola Slavchev and Eric M. Keller and Jonathan M. Snodgrass and Thomas J. Overbye},
  title        = {Testing Fine-Tuned Large Language Models for Power System Analysis},
  booktitle    = {2026 Texas Power and Energy Conference (TPEC)},
  year         = {2026},
  month        = feb,
  address      = {College Station, TX, USA},
  organization = {IEEE}
}

@article{zhang2025poweragent,
  author  = {Qian Zhang and Le Xie},
  title   = {{PowerAgent}: A Road Map Toward Agentic Intelligence in Power Systems: Foundation Model, Model Context Protocol, and Workflow},
  journal = {IEEE Power and Energy Magazine},
  volume  = {23},
  number  = {5},
  pages   = {93--101},
  year    = {2025},
  month   = sep,
  doi     = {10.1109/MPE.2025.3579718}
}

@article{xie2026foundationmodels,
  author  = {Le Xie and Qian Zhang and Minlan Yu and Paul L. Joskow and Chanan Singh},
  title   = {Crucial Role of Foundation Models in Enhancing the Interaction of {AI} and Power Systems: Achieving Integrated Frameworks},
  journal = {IEEE Energy Sustainability Magazine},
  pages   = {22--28},
  year    = {2026},
  month   = feb,
  doi     = {10.1109/ESM.2025.3630998}
}

@article{kiasari2026agenticsg,
  author    = {Mahmoud Kiasari and Hamed H. Aly},
  title     = {Agentic Artificial Intelligence for Smart Grids: A Comprehensive Review of Autonomous, Safe, and Explainable Control Frameworks},
  journal   = {Energies},
  volume    = {19},
  number    = {3},
  pages     = {617},
  year      = {2026},
  doi       = {10.3390/en19030617},
  publisher = {MDPI}
}

@article{ghosh2025agenticee,
  author  = {Soham Ghosh and Gaurav Mittal},
  title   = {Agentic {AI} Systems in Electrical Engineering: Current State-of-the-Art and Challenges},
  journal = {arXiv preprint arXiv:2511.14478},
  year    = {2025},
  url     = {https://arxiv.org/abs/2511.14478}
}

@article{elmakroum2026hems,
  author    = {Reda {El Makroum} and Sebastian Zwickl-Bernhard and Lukas Kranzl},
  title     = {Agentic {AI} Home Energy Management System: A Large Language Model Framework for Residential Load Scheduling},
  journal   = {Results in Engineering},
  volume    = {29},
  pages     = {109857},
  year      = {2026},
  doi       = {10.1016/j.rineng.2026.109857},
  publisher = {Elsevier}
}

@misc{2025deepseek,
author = {Ramachandran, Anand},
year = {2025},
month = {01},
pages = {},
title = {DeepSeek: Revolutionizing AI with Open-Source Reasoning Models -Advancing Innovation, Accessibility, and Competition with OpenAI and Gemini 2.0}
}

@online{MetaIntelligenceReasoning2026,
  author       = {{Meta Intelligence}},
  title        = {Reasoning Model Practical Guide: Enterprise Comparison and Deployment Strategies for DeepSeek R1, OpenAI o3, and Gemini 3},
  year         = {2026},
  url          = {https://www.meta-intelligence.tech/en/insight-reasoning-models},
  organization = {Meta Intelligence},
  note         = {Accessed: 2026-06-30}
}

@online{GoogleCloud_GeminiPlatform_ManagedAgents,
  author       = {{Google Cloud}},
  title        = {Building Managed Agents on the {Gemini Enterprise Agent Platform}},
  year         = {2026},
  url          = {https://docs.cloud.google.com/gemini-enterprise-agent-platform/build/managed-agents},
  organization = {Google Cloud Documentation},
  note         = {Accessed: 2026-06-30}
}

@online{GoogleAI_GeminiAPI_CustomAgents,
  author       = {{Google AI for Developers}},
  title        = {Building Managed Agents with the {Gemini API}},
  year         = {2026},
  url          = {https://ai.google.dev/gemini-api/docs/custom-agents},
  organization = {Google AI Documentation},
  note         = {Accessed: 2026-06-30}
}

@online{GoogleCloud_GeminiPlatform_Interaction,
  author       = {{Google Cloud}},
  title        = {Interacting with Managed Agents on the {Gemini Enterprise Agent Platform}},
  year         = {2026},
  url          = {https://docs.cloud.google.com/gemini-enterprise-agent-platform/build/managed-agents/interact-with-agents},
  organization = {Google Cloud Documentation},
  note         = {Accessed: 2026-06-30}
}

@online{GoogleCloudBlog_IO26_AgentNews,
  author       = {Google Cloud Developers and Practitioners Team},
  title        = {{I/O '26} News for Agent Developers on {Google Cloud}},
  year         = {2026},
  month        = may,
  url          = {https://cloud.google.com/blog/topics/developers-practitioners/io26-news-for-agent-developers-on-google-cloud},
  organization = {Google Cloud Blog},
  note         = {Accessed: 2026-06-30}
}

@INPROCEEDINGS{Composable2026,
  author={Al-Refai, Omar and Shahbaz, Ibrahim and Hammad, Eman},
  booktitle={2026 56th Annual IEEE International Conference on Dependable Systems and Networks Workshops (DSN-W)}, 
  title={Composable Trust in Agentic AI: Bridging Architectural Capability and System-Level Assurance}, 
  year={2026},
  volume={},
  number={},
  pages={17-20},
  doi={10.1109/DSN-W70714.2026.00023}}

% ================================================================
% Author Biographies
% ================================================================
%\newpage
\begin{IEEEbiography}
[{\includegraphics[
    width=1in,
    height=1.25in,
    clip,
    keepaspectratio
]{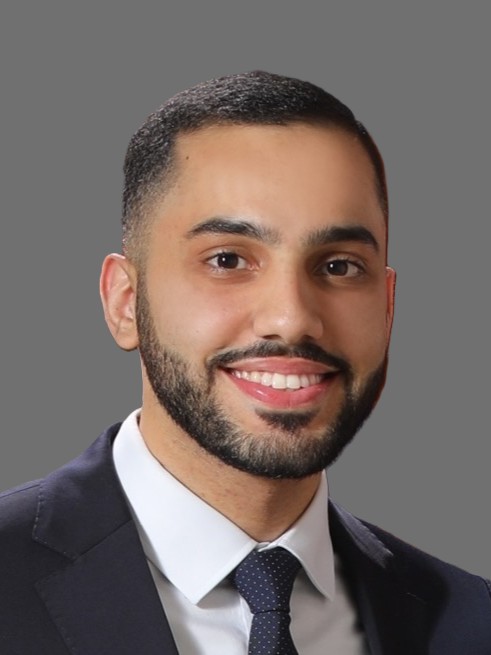}}]
{Omar Al-Refai}
(Student Member, IEEE) received the B.Sc. degree in computer engineering from Princess Sumaya University for Technology, Amman, Jordan. He is currently pursuing the Ph.D. degree in computer engineering at Texas A\&M University, College Station, TX, USA. His research at the Innovations in Systems Trust and Resilience (iSTAR) Laboratory focuses on enhancing trust and resilience in autonomous, collaborative, and human-in-the-loop systems and on connecting digital realms through the integration of humans, machines, and agents. His broader research interests include artificial intelligence, machine learning, and security.
\end{IEEEbiography}

\begin{IEEEbiography}[{\includegraphics[width=1in,height=1.25in,clip,keepaspectratio]{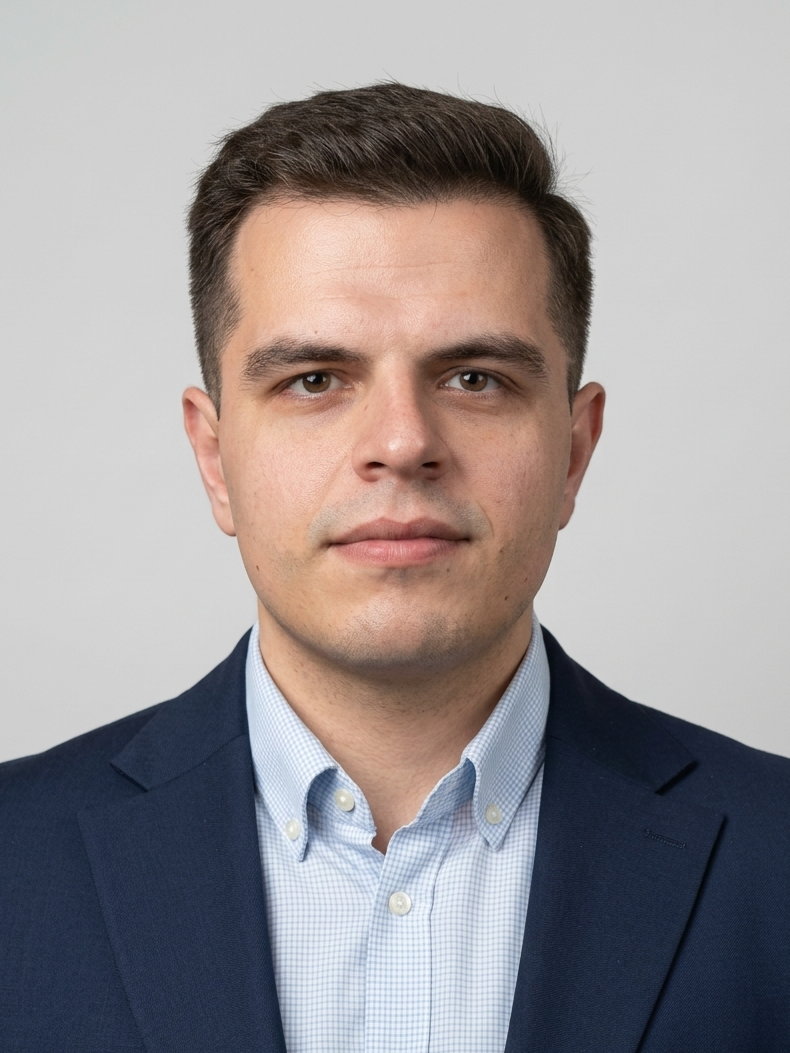}}]
{Ibrahim Shahbaz} (Student Member, IEEE) received the B.Sc. degree in electrical engineering from the University of Jordan, Amman, Jordan, and the M.Sc. degree in Data Science from Princess Sumaya University for Technology, Amman, Jordan. He is currently a Ph.D. student in electrical engineering at Texas A\&M University, College Station, TX, USA. His research interests at the iSTAR Lab include intelligent cyber-physical systems, resilient smart grids, and scientific machine learning.
\end{IEEEbiography}

\begin{IEEEbiography}[{\includegraphics[width=1in,height=1.25in,clip,keepaspectratio]{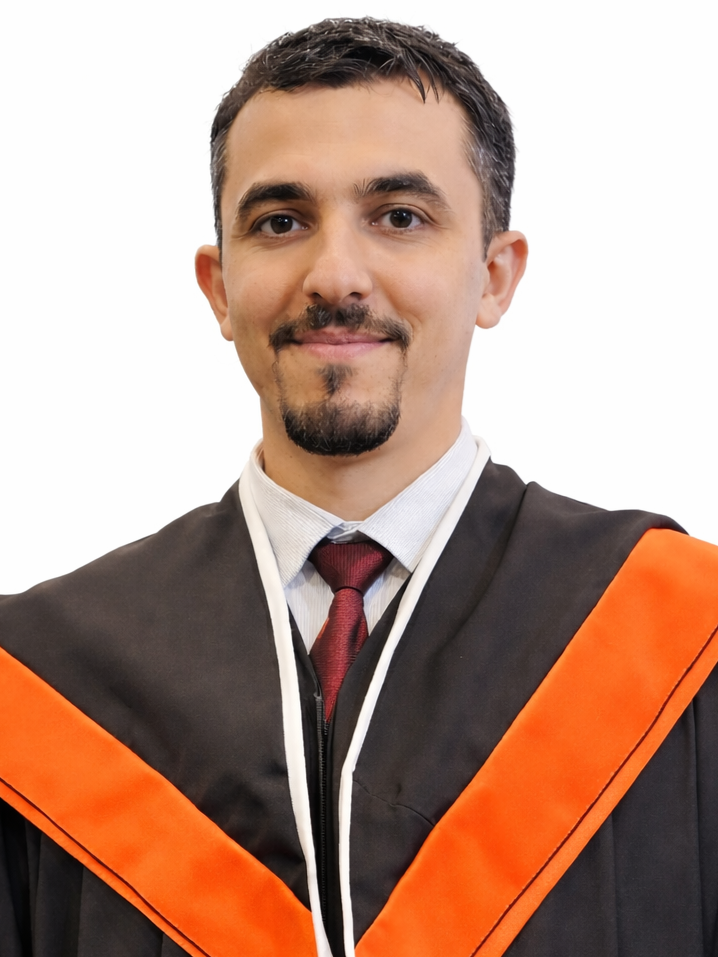}}]{Adam Ali Husseinat}(Member, IEEE) received the B.Sc. degree in electrical engineering from Jordan University of Science and Technology, Jordan, in 2009, and the M.Sc. degree in systems and computer engineering from Carleton University, Canada, in 2025. He is currently pursuing the Ph.D. degree in electrical and computer engineering at Texas A\&M University, College Station, TX, USA. His research interests include wireless communications, machine learning for communications, semantic communication, and 5G/6G intelligent networks.
\end{IEEEbiography}

\begin{IEEEbiography}[{\includegraphics[width=1in,height=1.25in,clip,keepaspectratio]{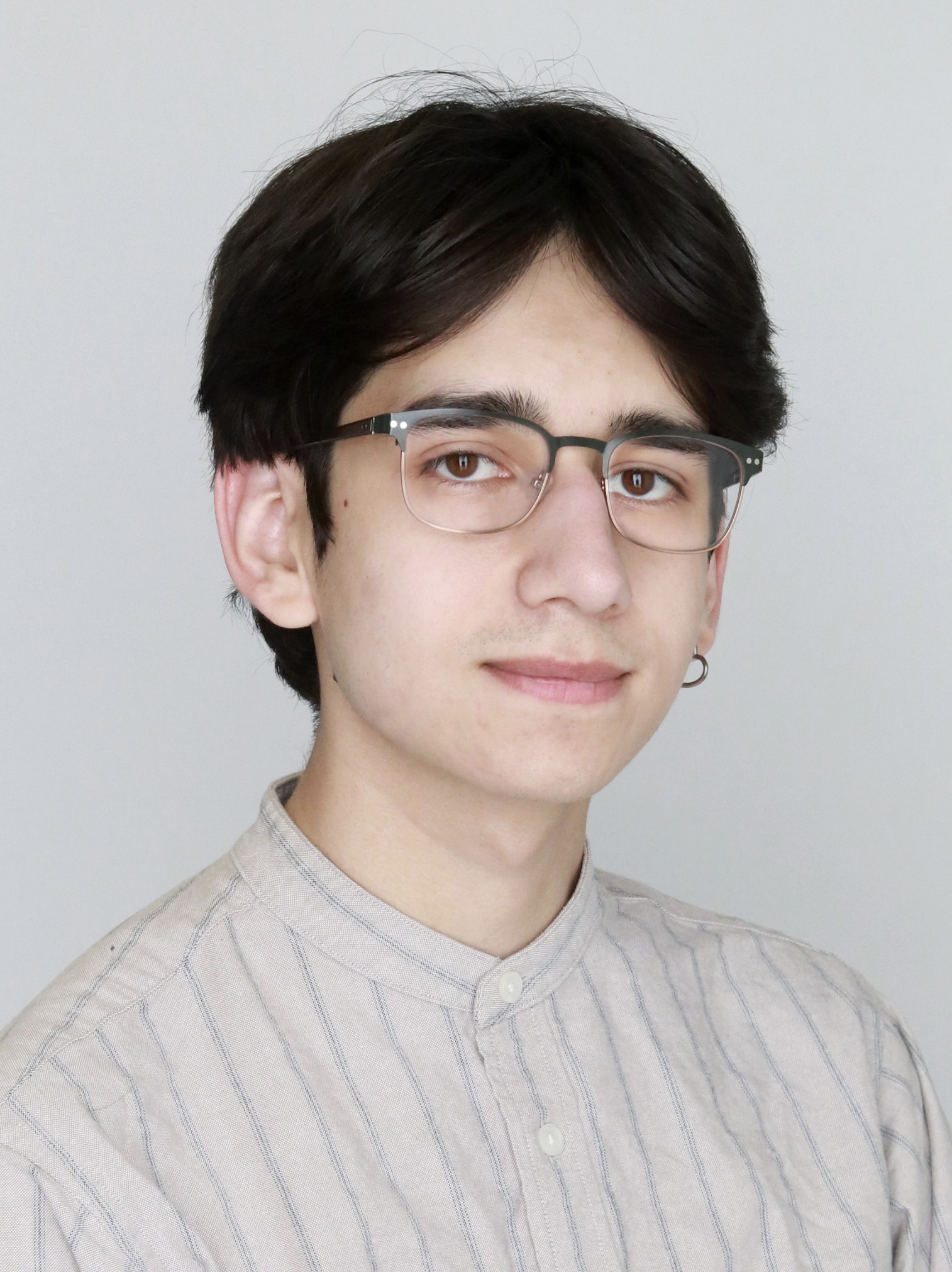}}]{Michael Mandulak}
received his Ph.D. in Computer Science from Rensselaer Polytechnic Institute (RPI), Troy, NY, USA. in 2025. He is currently a Postdoctoral Research Associate at the Texas A\&M University Institute of Data Science (TAMIDS) in the Security, Privacy and Resilience for Trusted AI (SPARTA) Laboratory. His research interests include high-performance computing, large-scale graph analytics, distributed graph processing, matching algorithms, and security in graph-based AI systems.
\end{IEEEbiography}

\begin{IEEEbiography}[{\includegraphics[width=1in,height=1.25in,clip,keepaspectratio]{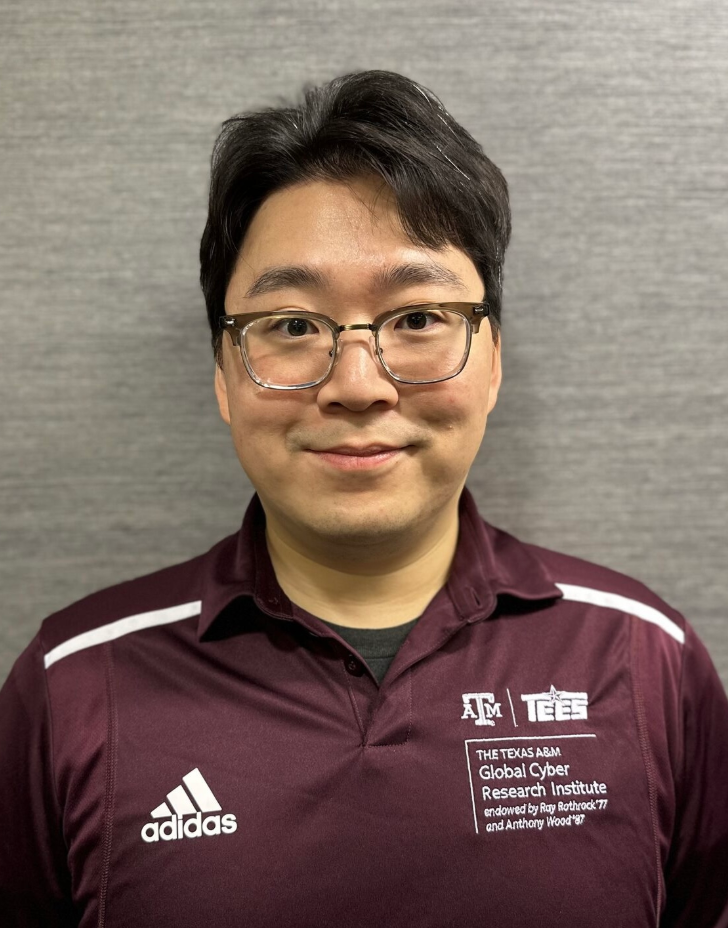}}]{Jaewon Kim}(Member, IEEE)  received his Ph.D. degree in Computer Engineering from Texas A\&M University in 2023. He is currently an Assistant Research Scientist at the Texas A\&M Global Cyber Research Institute (GCRI). Prior to joining GCRI, he was a Postdoctoral Associate in the Laboratory for Information \& Decision Systems (LIDS) at the Massachusetts Institute of Technology (MIT). He spearheads an interdisciplinary research program at the critical intersection of cyber-physical systems (CPS) and security of future autonomous infrastructures. His work investigates resilient real-time network and zero-trust architectures (ZTA), incorporating robust authentication and micro-segmentation, to secure unmanned aerial and ground vehicles (UAVs/UGVs). 
\end{IEEEbiography}

\begin{IEEEbiography}
[{\includegraphics[
    width=1in,
    height=1.25in,
    clip,
    keepaspectratio
]{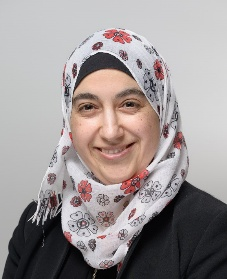}}]
{Eman Hammad}
(Senior Member, IEEE) is an Assistant Professor and the Director of the Innovations in Systems Trust and Resilience (iSTAR) Laboratory at Texas A\&M University, College Station, TX, USA. She is also the Director of the Texas A\&M Data Institute Thematic Laboratory for Security, Privacy, and Resilience for Trusted AI (SPARTA). She received the Ph.D. degree from the Department of Electrical and Computer Engineering at the University of Toronto, Toronto, ON, Canada. Her research is primarily in the domain of security and resilience of complex network, cyber-physical systems and critical infrastructures' with applications in power systems, 6G/NG wireless networks, connected and autonomous vehicles (UAV/UGV), and public safety. Focus areas include  trustworthy autonomous systems, physical AI, human-AI interactions, risk-informed operations, 6G/NG communication networks, semantic and graph-based methods, digital well-being, and CyberAI education.

%Her research interests include large-scale adaptive, reliable, and trustworthy heterogeneous networks, connected intelligence, systems integration and interoperability, metrics-informed design, and security and resilience by design.
\end{IEEEbiography}
\newpage

% \section{Biography Section}
% If you have an EPS/PDF photo (graphicx package needed), extra braces are
%  needed around the contents of the optional argument to biography to prevent
%  the LaTeX parser from getting confused when it sees the complicated
%  $\backslash${\tt{includegraphics}} command within an optional argument. (You can create
%  your own custom macro containing the $\backslash${\tt{includegraphics}} command to make things
%  simpler here.)
 
% \vspace{11pt}

% \bf{If you include a photo:}\vspace{-33pt}
% \begin{IEEEbiography}[{\includegraphics[width=1in,height=1.25in,clip,keepaspectratio]{fig1}}]{Michael Shell}
% Use $\backslash${\tt{begin\{IEEEbiography\}}} and then for the 1st argument use $\backslash${\tt{includegraphics}} to declare and link the author photo.
% Use the author name as the 3rd argument followed by the biography text.
% \end{IEEEbiography}

% \vspace{11pt}

% \bf{If you will not include a photo:}\vspace{-33pt}
% \begin{IEEEbiographynophoto}{John Doe}
% Use $\backslash${\tt{begin\{IEEEbiographynophoto\}}} and the author name as the argument followed by the biography text.
% \end{IEEEbiographynophoto}

\vfill

\end{document}